\documentclass[a4paper,fleqn]{cas-dc}

\usepackage{amssymb}
\usepackage[numbers]{natbib}
\usepackage{hhline}
\usepackage{threeparttable}
\usepackage{tabularx}
\usepackage{multirow}
\usepackage{xcolor}
\usepackage{booktabs}
\usepackage{multicol}
\usepackage[ruled,vlined]{algorithm2e}
\usepackage[figuresright]{rotating} 
\usepackage{subfigure}
\usepackage{graphicx}
\usepackage{colortbl}
\usepackage{color}
\usepackage[numbered]{bookmark}
\usepackage{hyperref}
\hypersetup{
        bookmarksnumbered=true,
        bookmarksopen=true,
        bookmarksopenlevel=5,
        pdfstartview=Fit,
        pdfpagemode=UseOutlines
    }
\usepackage{array}
\usepackage[figuresright]{rotating}
\usepackage{threeparttable}

\usepackage[switch]{lineno}

\def\tsc#1{\csdef{#1}{\textsc{\lowercase{#1}}\xspace}}
\tsc{WGM}
\tsc{QE}
\tsc{EP}
\tsc{PMS}
\tsc{BEC}
\tsc{DE}

\begin{document}
\let\WriteBookmarks\relax
\def\floatpagepagefraction{1}
\def\textpagefraction{.001}
\shorttitle{Reinforcement Learning-assisted Evolutionary Algorithm: A Survey and Research Opportunities}
\shortauthors{Yanjie Song et~al.}

\title [mode = title]{Reinforcement Learning-assisted Evolutionary Algorithm: A Survey and Research Opportunities}                      

%

\author[1]{Yanjie Song }[orcid=0000-0002-4313-8312]
\ead{songyj_2017@163.com}
\address[1]{National Defense University, Haidian District, Beijing, China}

\author[2]{Yutong Wu }
\ead{wuyutong119@gmail.com}
\address[2]{Department of Analytics, Operations and Systems, University of Kent, Canterbury, Kent, UK}

\author[3]{Yangyang Guo }
\ead{g2002dmu@163.com}
\address[3]{School of Systems Science, Beijing Jiaotong University, Beijing, China}

\author[4]{Ran Yan}
\ead{ran.yan@ntu.edu.sg}
\address[4]{School of Civil and Environmental Engineering, Nanyang Technological University, Singapore}

\author[5]{Ponnuthurai Nagaratnam Suganthan}
\ead{p.n.suganthan@qu.edu.qa}
\cormark[1]
\address[5]{KINDI Center for Computing Research, College of Engineering, Qatar University, Doha, Qatar}

\author[6,7]{Yue Zhang}
\ead{zhangyue1127@buaa.edu.cn}
\address[6]{School of Reliability and Systems Engineering, Beihang University, China}
\address[7]{Department of Industrial Systems Engineering and Management, National University of Singapore, Singapore}

\author[8,9,10]{Witold Pedrycz}
\ead{wpedrycz@ualberta.ca}
\address[8]{Department of Electrical and Computer Engineering, University of Alberta, Edmonton, Canada}
\address[9]{Systems Research Institute, Polish Academy of Sciences, Poland}
\address[10]{Faculty of Engineering and Natural Sciences, Department of Computer Engineering, Sariyer/Istanbul, Turkiye}


\author[11]{Swagatam Das}
\ead{swagatam.das@isical.ac.in}
\address[11]{Electronics and Communication Sciences Unit Indian Statistical Institute, Kolkata, India}

\author[12]{Rammohan Mallipeddi}
\ead{mallipeddi.ram@gmail.com}
\cormark[2]
\address[12]{School of Electronics Engineering, Kyungpook National University, Taegu, South Korea}

\author[12]{Oladayo Solomon Ajani}
\ead{oladayosolomon@gmail.com}

\author[6]{Qiang Feng}
\ead{fengqiang@buaa.edu.cn}


%
%
%
%
%
%
%
%
%

\cortext[cor1]{Corresponding author}
\cortext[cor2]{Corresponding author}
%

\begin{abstract}
Evolutionary algorithms (EA), a class of stochastic search methods based on the principles of natural evolution, have received widespread acclaim for their exceptional performance in various real-world optimization problems. While researchers worldwide have proposed a wide variety of EAs, certain limitations remain, such as slow convergence speed and poor generalization capabilities. Consequently, numerous scholars actively explore improvements to algorithmic structures, operators, search patterns, etc., to enhance their optimization performance. Reinforcement learning (RL) integrated as a component in the EA framework has demonstrated superior performance in recent years. This paper presents a comprehensive survey on integrating reinforcement learning into the evolutionary algorithm, referred to as reinforcement learning-assisted evolutionary algorithm (RL-EA). We begin with the conceptual outlines of reinforcement learning and the evolutionary algorithm. We then provide a taxonomy of RL-EA. Subsequently,  we discuss the RL-EA integration method, the RL-assisted strategy adopted by RL-EA, and its applications according to the existing literature. The RL-assisted procedure is divided according to the implemented functions including solution generation, learnable objective function, algorithm/operator/sub-population selection, parameter adaptation, and other strategies. Additionally, different attribute settings of RL in RL-EA are discussed. In the applications of RL-EA section, we also demonstrate the excellent performance of RL-EA on several benchmarks and a range of public datasets to facilitate a quick comparative study. Finally, we analyze potential directions for future research. This survey serves as a rich resource for researchers interested in RL-EA as it overviews the current state-of-the-art and highlights the associated challenges. By leveraging this survey, readers can swiftly gain insights into RL-EA to develop efficient algorithms, thereby fostering further advancements in this emerging field.

\end{abstract}



\begin{keywords}
evolutionary algorithm \sep reinforcement learning \sep optimization \sep reinforcement learning-assisted strategy \sep generating solution \sep learnable objective function \sep algorithm/operator/sub-population selection
\sep parameter adaptation
\end{keywords}

\maketitle

\section{Introduction}

The concept of "optimization" has become widely recognized in various fields in recent years. Presently, a multitude of real-world problems spanning scientific research, economics, and engineering are essentially modeled as intricate optimization problems and tackled through algorithmic approaches \cite{singh2012overview}. The present optimization problems have been addressed by researchers through the development of numerous classical algorithms, as well as their enhanced variants. Broadly speaking, solution algorithms can be classified into two categories: exact solution algorithms \cite{monaci2013exact} and approximate solution algorithms \cite{babaei2013general}. Exact solution algorithms can theoretically guarantee the optimal solution at the cost of an exponential explosion in the solution time as the problem size increases. In contrast, approximate solution algorithms are not constrained by problem size and are well-suited for addressing numerous practical large-scale problems, particularly those characterized by multi-modality, dynamics, discontinuities, and nonlinearity. Among these algorithms, evolutionary algorithms (EAs) have gained significant popularity and are widely used in various problem scenarios due to their ability to simulate the natural evolution process and perform population-based random searches.

While EA has achieved remarkable success in various problem domains, it suffers from a significant drawback of low sampling efficiency during iterative search \cite{de2017evolutionary}. This limitation becomes more stringent as the size of the solution space increases. In addition, due to factors such as increased problem complexity, advancements in theoretical methods and computational resources, and the pursuit of efficiency, enhanced versions of EA have emerged with strong robustness, exploratory capabilities, and rapid convergence. Among the attempts at algorithmic improvement, a series of machine learning (ML) methods have recently gained popularity due to their exceptional performance in extracting explicit knowledge by fully leveraging the available data \cite{mahesh2020machine}. The data generated through EA iterations can be consumed by ML techniques to obtain useful knowledge to enhance the algorithm's performance \cite{talbi2021machine}. Unlike conventional ML methods (including supervised, semi-supervised, and unsupervised learning), which have demonstrated desired outcomes in numerous optimization tasks, reinforcement learning (RL) methods consider that agents take actions by interacting with the environment, thus eliciting a strong reasoning ability \cite{jordan2015machine}. 

Nowadays, RL has demonstrated remarkable performance in decision-making problems across various complex and stochastic environments, particularly when combined with deep neural networks \cite{mnih2013playing}. This integration significantly enhances the method's inference and learning capabilities for uncertain sequential decision-making problems \cite{franccois2018introduction}. Therefore, numerous studies have integrated RL into the EA framework to establish reinforcement learning-assisted evolutionary algorithms (RL-EAs), which have been proven successful in addressing a wide range of complex optimization problems.

\begin{figure}[htp]
\centering
\includegraphics[width=0.45\textwidth]{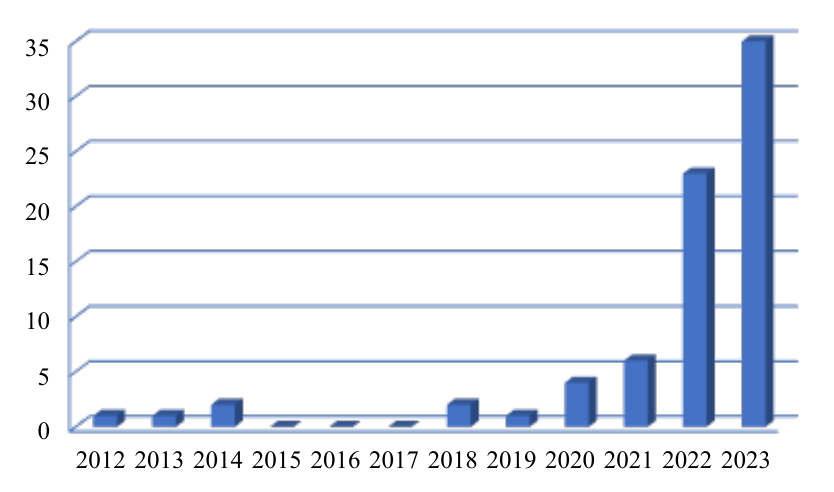}
\caption{Overview of the number of RL-EA related studies by year}
\label{Overview of the number of RL-EA related studies by year}
\end{figure}

The statistics presented in Fig.\ref{Overview of the number of RL-EA related studies by year} demonstrate increasing interest and research efforts dedicated to RL-EA for optimization problems from 2012 up to August 4, 2023. It is evident that the field has witnessed a growing momentum over time, with particular emphasis on the last three years. This surge can be attributed to researchers' enhanced comprehension of these algorithms. In general, RL can be effectively integrated into the EA framework for several reasons. Firstly, it enables the conversion of any optimization problem into a sequential decision problem that can be modeled and solved by RL. Secondly, RL exhibits strong learning capability and low time complexity during the testing stage. Additionally, there exists valuable algorithm design experience pertaining to specific problems which can greatly facilitate further research endeavors.

In this survey, we focus on various aspects of RL-EA design and implementation, including algorithmic frameworks, implementation strategies, and major applications. \textbf{To the best of our knowledge, there is currently no comprehensive survey available on the subject of RL-EA.} The most relevant previous review to this paper is \cite{talbi2021machine}, which primarily focuses on in-depth analysis and discussion regarding the integration of machine learning methods with meta-heuristic algorithms. Although it involved some strategies for combining RL and EA, it only provided a brief introduction without offering targeted suggestions for algorithm design. Another related survey focuses on the interaction between RL and EA \cite{drugan2019reinforcement}. There also exist some surveys that focus on evolutionary reinforcement learning \cite{bai2023evolutionary}, the application of reinforcement learning methods to optimization problems \cite{mazyavkina2021reinforcement,yang2020survey,song2023ensemble}.  Different from these surveys, this survey discusses successful applications of RL integration within the EA framework specifically. This survey enables readers to quickly grasp the relevant design methodologies of RL-EA and perform further and comprehensive research on specific issues. The key contributions of this survey cover the following four aspects. 1) We systematically analyze the integration of RL and EA frameworks in RL-EA, providing targeted guidance for algorithmic design. 2) Detailed discussions are conducted on how RL can assist EA search through learning objective functions, solution generation, operator and algorithm selection, etc. 3) The primary application domains of RL-EA in optimization are categorized. Furthermore, a list of public datasets attached with the previous work on RL-EA is sorted out. 4) A range of future directions for addressing challenges are discussed in depth. 

The remainder of this paper is structured as follows: Section \ref{Backgrounds} provides an overview of the background of evolutionary algorithms assisted by reinforcement learning. Section \ref{The Taxonomy of RL-EA} elaborates on the classification of RL-EA. Section \ref{RL and EA Integration Methods} indicates the integration methods employed for combining RL with EA. Section \ref{RL Assistance Strategies Used in RL-EA} describes the RL-assisted strategies applied to RL-EA. Section \ref{Attribute Settings of RL in RL-EA} gives an overview of the attribute settings of RL in RL-EA. Section \ref{Application of RL-EA} explores various applications of RL-EA. Section \ref{Future Research Directions} discusses potential future research directions. Section \ref{Conclusion} summarizes the study.

\section{Backgrounds}
\label{Backgrounds}
In this section, we provide a brief overview of evolutionary algorithms and reinforcement learning.

\subsection{Evolutionary Algorithms}

\begin{figure}[htp]
\centering
\includegraphics[width=0.35\textwidth]{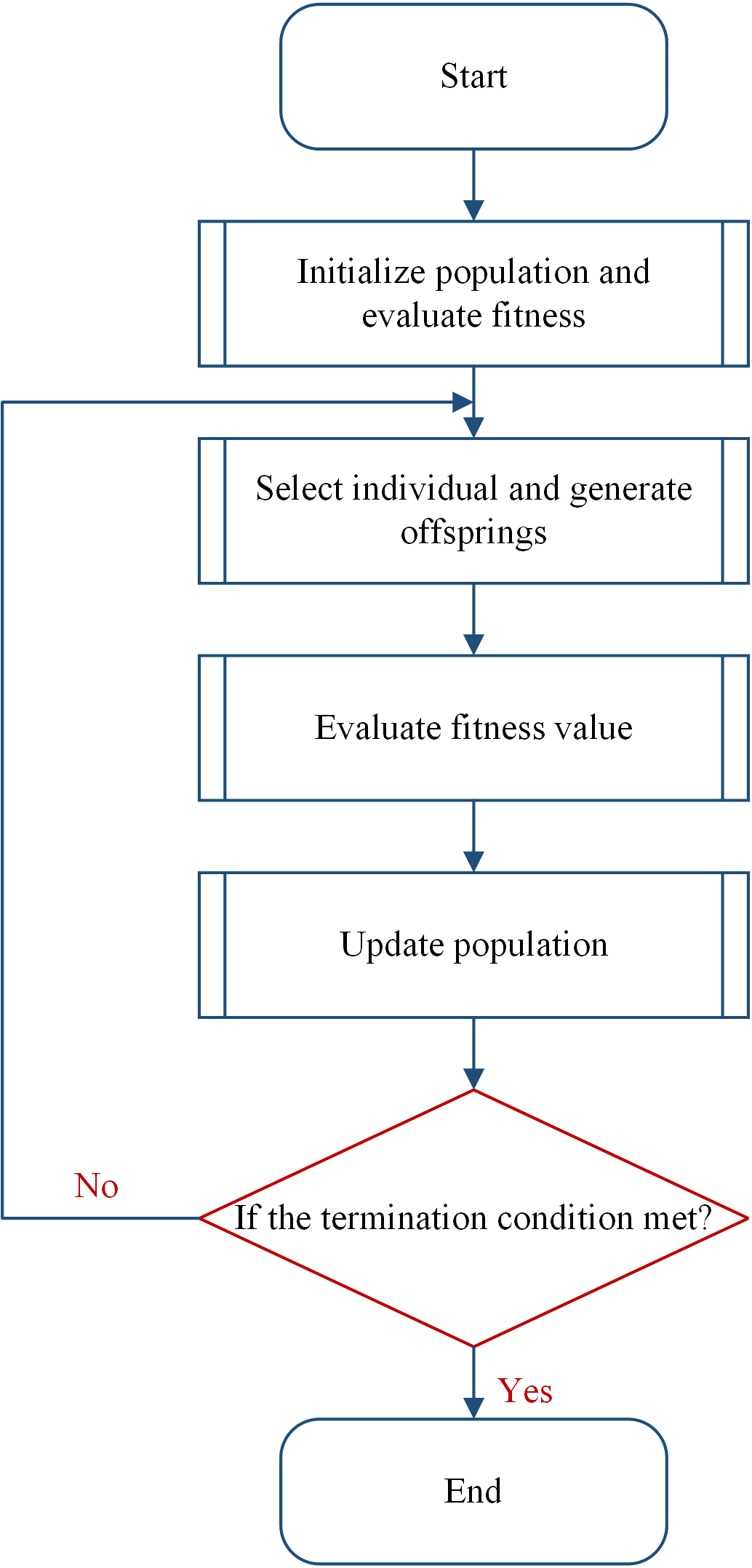}
\caption{Flowchart of a generic evolutionary algorithm}
\label{The general flowchart of evolutionary algorithm}
\end{figure}

The evolutionary algorithms (EA) belong to a category of stochastic search algorithms that are designed based on the principles governing evolution in nature. Though EAs may not guarantee optimality when solving complex problems, they can effectively discover an approximate optimal solution through population-based exploration, which is deemed acceptable for numerous application scenarios. For an optimization problem, EA persistently explores and refines solutions through a series of specific mechanisms such as selection, evolution, and evaluation in the form of iterative population updates. Here, we present a brief flowchart of an evolutionary algorithm (refer to Fig.\ref{The general flowchart of evolutionary algorithm}). Initially, the algorithm is initialized to generate a population. Subsequently, a subset of individuals from this population are randomly selected and then they produce offspring through specific methods such as crossover or mutation. Afterward, the optimal individuals among these parents and offspring are chosen to form a new population. Based on the aforementioned description, the process of population evolution continues until the termination condition is met. Ultimately, the most high-performing individuals are selected as the final solution.

Through continuous development, researchers have developed various algorithms inspired by different types of natural behavior and enhanced versions of classical algorithms, such as the genetic algorithm (GA) \cite{mirjalili2019genetic}, differential evolution (DE) algorithm \cite{price2013differential}, and particle swarm optimization (PSO) algorithm \cite{kennedy1995particle}. However, as stated in the No Free Lunch (NFL) theorem, it is exceedingly difficult to find an algorithm that performs well at a low time cost and is adaptable to various complex problem scenarios \cite{wolpert1997no}. Therefore, it is crucial to customize the algorithmic configuration specifically for a given problem. Over the past few decades, a multitude of strategies have emerged with the aim of enhancing evolutionary algorithms, among which GA has been extensively employed in RL-EA.

Here, we provide a concise introduction to the GA, which is a classical EA after Darwinian theory. In this approach, each potential solution corresponds to a chromosome and the search process involves initialization, fitness evaluation, selection, crossover, and mutation. Within this work, crossover and mutation are the two primary evolutionary operations in traditional GA. With the continuous advancement of algorithms, additional operations such as reproduction and recombination have also emerged. The key distinction between crossover and mutation lies in their respective impacts on chromosome structure and likelihood of occurrence. The rational utilization of these operators directly influences the quality of the solution. Therefore, many works have focused on the design of evolutionary operators and the control of search strategies. Among these, the memetic algorithm (MA) \cite{neri2012memetic} stands out as a classical variant that introduces local search based on population search to enhance its exploitation capabilities. In addition, the non-dominated sorting genetic algorithm II (NSGA-II) \cite{deb2002fast}, which incorporates a fast non-dominated sorting mechanism, and the reference-point-based many-objective evolutionary algorithm following the NSGA-II framework (NSGA-III) \cite{deb2013evolutionary} are two prominent algorithms for solving multi-objective problems (MOPs), respectively.

\subsection{Reinforcement Learning}

\begin{figure}[htp]
\centering
\includegraphics[width=0.4\textwidth]{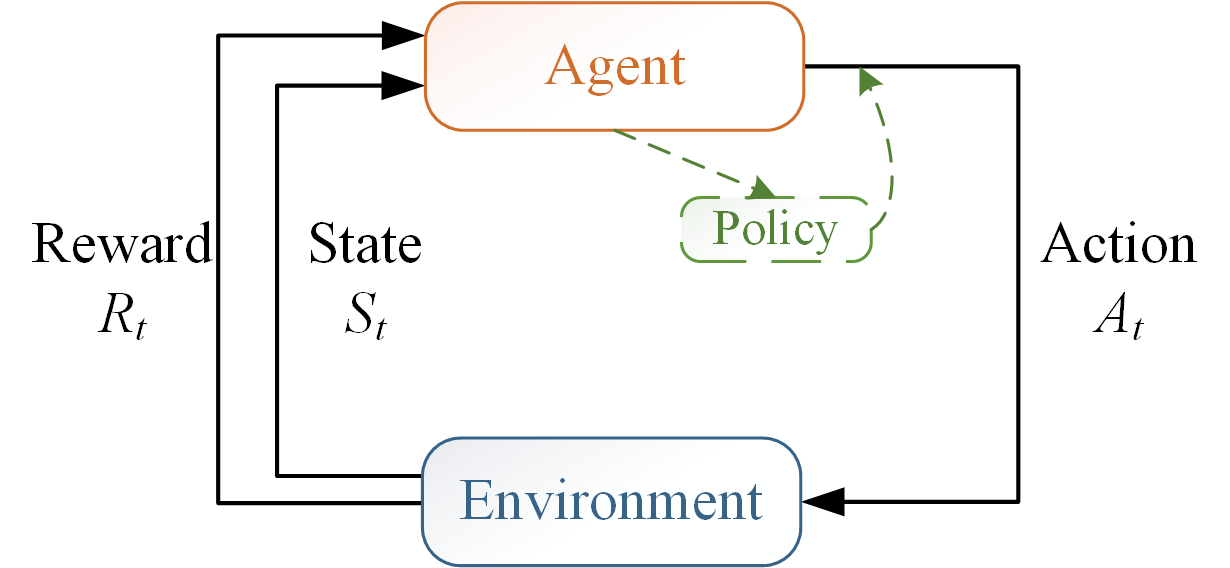}
\caption{Information interaction between agent and environment in RL}
\label{Information interaction between agent and environment in RL}
\end{figure}

Reinforcement learning methods are a powerful category of ML methods for proficient policy decision-making in diverse and uncertain environments, commonly referred to as approximate dynamic programming \cite{powell2007approximate}. The distinguishing characteristic of RL is learning from interactions with the environment, setting it apart from both supervised and unsupervised learning approaches. In RL, the agent aims to maximize cumulative rewards by actively engaging with the environment and making optimal decisions. The complete framework of a RL system typically encompasses not only the agent, but also the state space $S$, action space $A$, state transfer function $P$, reward function $R$, and discount factor $\gamma$. In other words, a quintuple $\langle S, A, P, R, \gamma\rangle$ constitutes a Markov Decision Process (MDP), serving as the foundation for agent decision-making. The agent selects an action based on the current state, as depicted in Fig.\ref{Information interaction between agent and environment in RL}. Subsequently, the action's performance is assessed through interaction with the environment to obtain a reward. Then, considering both the state change and policy table, the agent determines the next action for decision-making. In this manner, policy $\pi$ is constructed by combining state and action. Furthermore, the agent continuously updates the policy to maximize the desired rewards, as given by:
\begin{equation}
Q^{\pi}(s,a)=\mathbb{E}_{\pi}\left[\sum_{t=0}^{\infty}\gamma^{t}R(s_{t},a_{t})|s_{0}=s,a_{0}=a\right].
\end{equation}

The RL methods can be classified into model-based RL and model-free RL, depending on whether they utilize a model or not \cite{sutton2018reinforcement}. Model-based RL constructs a complete environment model by predicting state transfer and reward function. This type of RL offers the advantage of accurate decision-making. However, it often encounters numerous challenges in obtaining the actual MDP for many problems. Unlike model-based RL, model-free RL does not involve learning the environment model. Due to the relatively lower learning complexity and the policy representation capabilities of artificial neural networks (ANNs), model-free RL has emerged as the dominant focus of research and application in recent years. Therefore, we focus on the introduction of model-free RL in our survey.

In model-free RL, there exist two categories of methods: policy-based methods and value-based methods. Policy-based methods optimize policies by parameterizing them based on the Policy Gradients Theorem \cite{sutton2018reinforcement}. Some common policy-based RLs include Policy Gradient (PG) \cite{sutton1999policy}, Asynchronous Advantage Actor-Critic (A3C) \cite{mnih2016asynchronous} Proximal Policy Optimization (PPO) \cite{schulman2017proximal}, and Deep Deterministic Policy Gradient (DDPG) \cite{lillicrap2015continuous}. The value-based methods focus on utilizing the Q-function to predict the state-action combination and select the action that yields the highest Q-value. Two value-based state-of-the-art (SOTA) methods are Q-learning \cite{watkins1992q} and Deep Q Network (DQN) \cite{mnih2015human} methods. Q-learning utilizes a Q-table to store the Q-values of $\langle S, A \rangle$ pair, which is straightforward, efficient, and particularly suitable for scenarios with limited state space. It stands as one of the most commonly employed RL methods in RL-EA. In addition, the more intricate Deep Q-Network (DQN) \cite{schaul2015prioritized} employs ANNs instead of Q-tables and incorporates the relay buffer mechanism in subsequent studies. The parameters are updated during DQN training using a training method that minimizes the Temporal Difference (TD) \cite{seymour2004temporal} loss. Furthermore, to enhance the robustness and sampling efficiency of DQN, several RL methods have been proposed, including dueling DQN \cite{wang2016dueling} and double DQN (DDQN) \cite{van2016deep}.

\section{The Taxonomy of RL-EA}
\label{The Taxonomy of RL-EA}

\begin{figure*}[htp]
\centering
\includegraphics[width=1\textwidth]{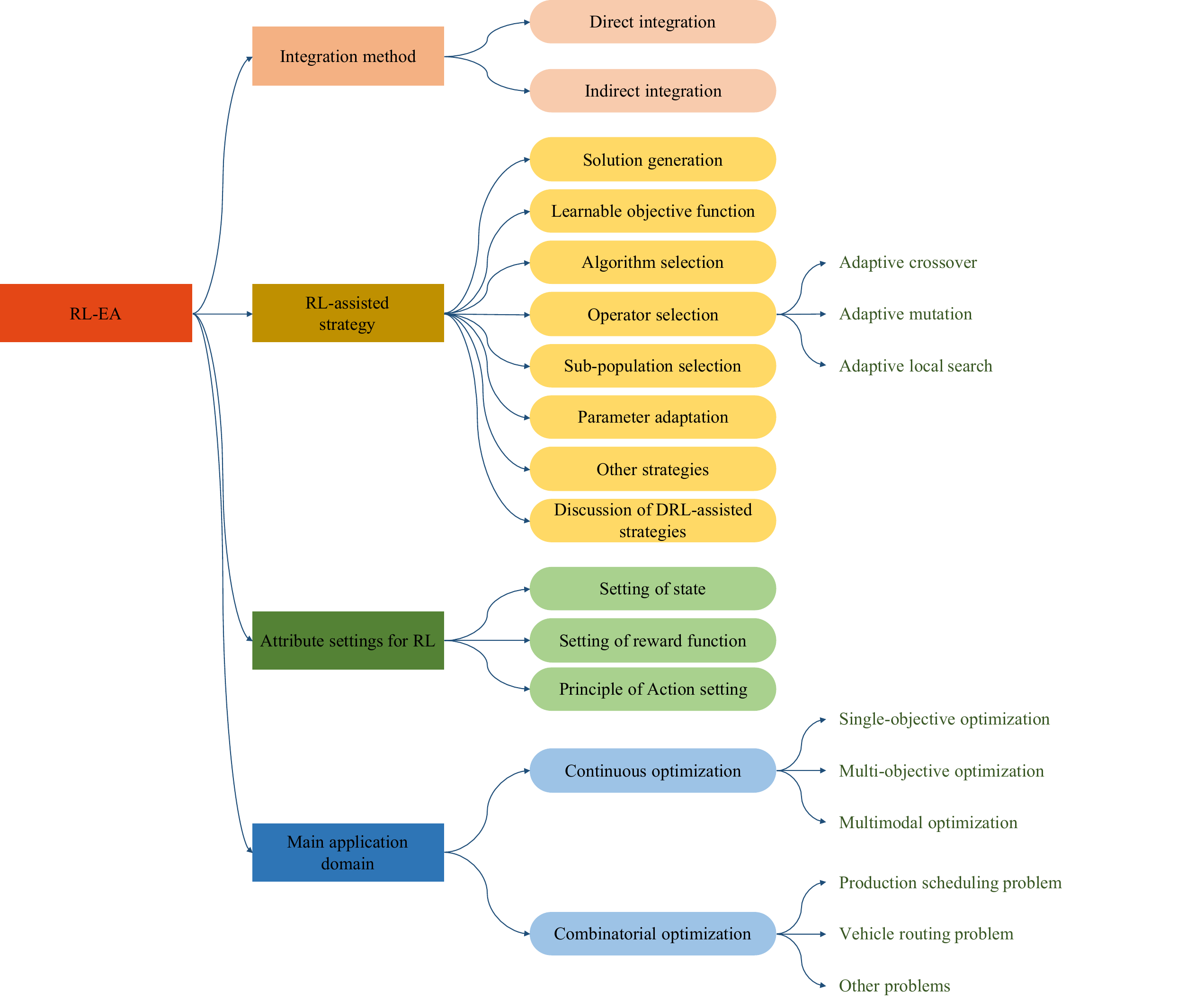}
\caption{The taxonomy of RL-EA in the survey}
\label{The taxonomy of RL-EA in the survey}
\end{figure*}

This section presents a taxonomy of RL-EA for solving optimization problems, which serves as the foundation for subsequent in-depth discussions in this survey. As illustrated in Fig.\ref{The taxonomy of RL-EA in the survey}, three main aspects of RL-EA are highlighted in this survey: the way RL integrates with EA, the assistance strategies of RL, attribute settings of RL in RL-EA, and the application of RL-EA. In Section \ref{RL and EA Integration Methods}, we discuss the integration approach of RL and EA. Subsequently, we provide a detailed presentation of the enhancement strategies employed in related studies to improve EA performance using RL in Section \ref{RL Assistance Strategies Used in RL-EA}. Section \ref{Attribute Settings of RL in RL-EA} introduces attribute settings of RL in RL-EA. Finally, an overview of RL-EA for specific optimization tasks in the main application domains is presented in Section \ref{Application of RL-EA}.

\section{Integration Methods for RL and EA}
\label{RL and EA Integration Methods}

Determining the form of information interaction between RL and EA is a prerequisite for RL to assist EA in population or individual searches. The integration of RL and EA typically involves both direct integration and indirect integration, in which algorithm design and information interaction are different. In the following section, we will provide detailed explanations of these two integration methods. 

\subsection{Direct Integration}

\begin{figure}[htp]
\centering
\includegraphics[width=0.4\textwidth]{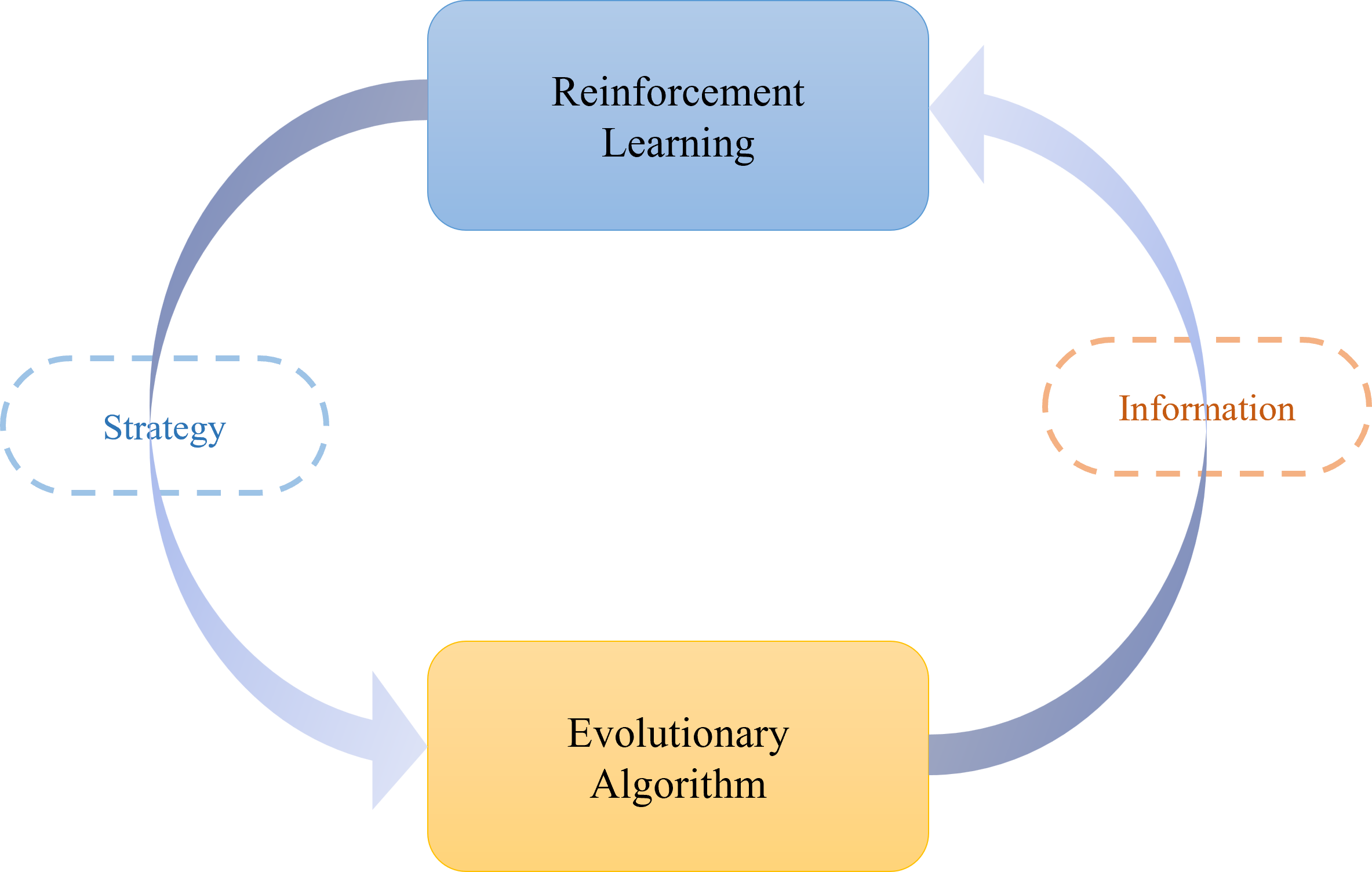}
\caption{Direct integration of RL and EA}
\label{Direct integration of RL and EA}
\end{figure}

The direct integration of RL and EA provides a simple approach to combining these two methods. As illustrated in Fig.\ref{Direct integration of RL and EA}, the action made by RL is directly applied to the search process within the algorithm, while feedback from search performance information is used by the agent to update the model. Although the functionality is relatively simple, the integration approach is easy to understand. Specifically, common strategies for RL using direct combination approaches include direct solution generation and automated tuning of control parameters \cite{li2023reinforcementb,hu2021reinforcement}.

The advantages of direct integration lie in the simplicity of algorithmic structure and the ease of implementation, allowing for state, action, and reward to be easily determined based on the design goals of the algorithm. Similarly, the EA does not require additional adjustments to use the policies generated by the RL decisions. However, direct integration is difficult to replace the original search strategy in EA, and can only play a certain role in auxiliary regulation of algorithmic search. Therefore, other improvement strategies are designed in many RL-EAs that use direct integration.

\subsection{Indirect Integration}

\begin{figure}[htp]
\centering
\includegraphics[width=0.4\textwidth]{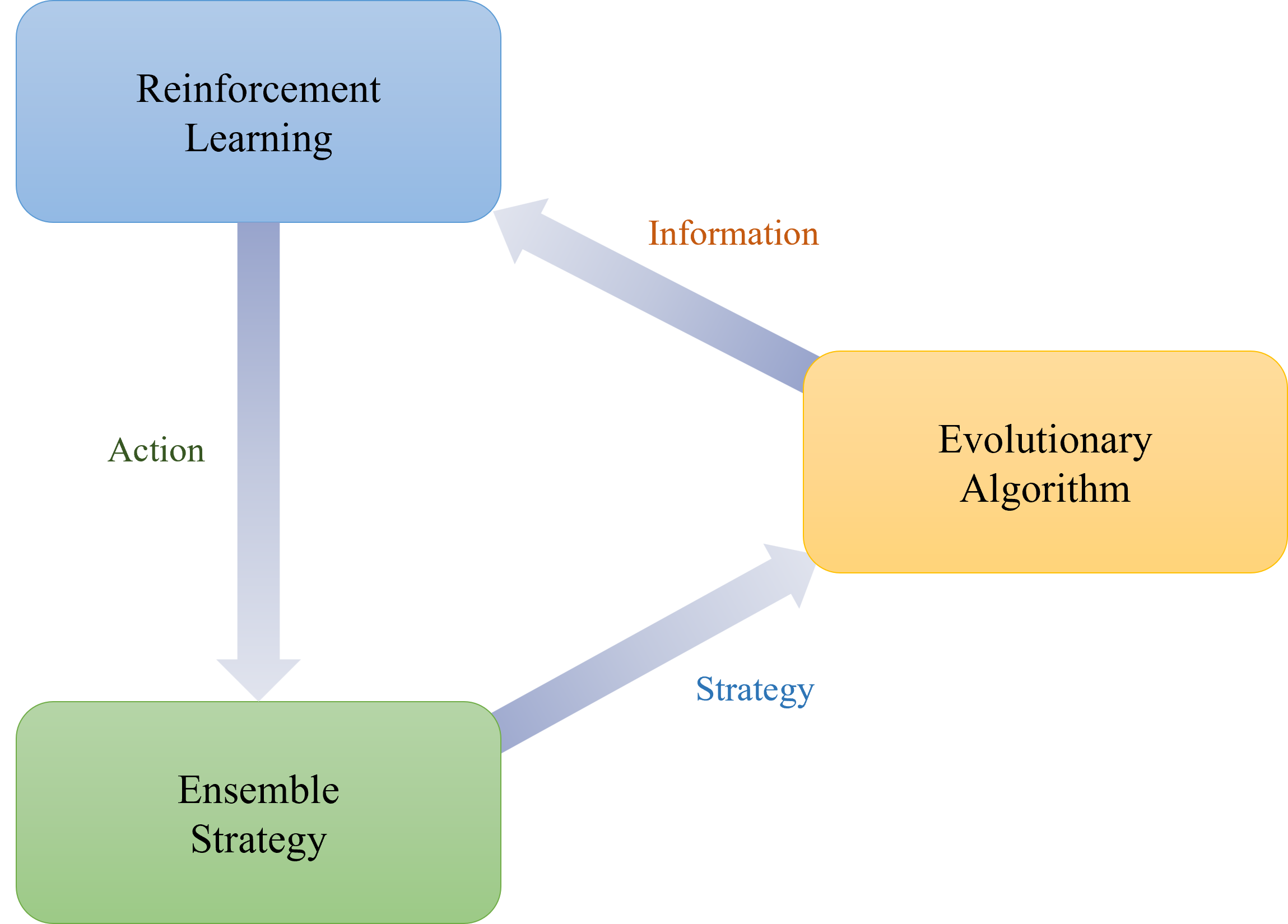}
\caption{Indirect integration of RL and EA}
\label{Indirect integration of RL and EA}
\end{figure}

The indirect integration approach is a commonly employed method that combines RL and EA. In this approach, RL does not directly influence the population search process but instead operates on the components within the ensemble strategies (ES) \cite{wu2019ensemble}. As depicted in Fig.\ref{Indirect integration of RL and EA}, RL makes decisions on the strategies in ES, which are then utilized during the EA search. After the search, RL updates these strategies based on their performance. RL-EA uses indirect integration and the algorithmic performance is influenced by the design of RL and ES. Notably, ES represents a widely adopted approach for enhancing EA performance, which can be smarter as driven by RL. The flexibility inherent in ES allows for the incorporation of multiple algorithms or operators as well as other operations. 

The advantage of employing this integration approach lies in its ability to enhance the algorithm's robustness significantly. This is attributed to the simultaneous utilization of diverse strategies within one framework, enabling it to effectively handle various complex scenarios. Inevitably, it should be noted that complexly structured algorithms often entail substantial computational resource consumption. Moreover, it may also encounter challenges such as mismatched search strategies and problem features, resulting in inefficient searches. To minimize the occurrence of the above situations, algorithms should possess the flexibility to adaptively determine the adopted strategies based on both previous and current search information. 

\textcolor[rgb]{0,0,0}{During the RL model training phase, algorithms are trained either individually or collectively based on the chosen strategy settings. Stand-alone training follows the same procedure as using only the RL method and does not require incorporating results from the EA search into the model input. Conversely, integrated training relies on RL-EA to optimize the RL model using data generated through iterative algorithmic search. The data or scenarios used for training the RL model differ from those employed in actual problem-solving and necessitate additional methods for their generation.}

\textcolor[rgb]{0,0,0}{In practical implementation, RL-EA exhibits no discernible differences from direct EA execution, and the RL model within the algorithm does not require further training at this stage. Moreover, both EA and RL effectively achieve a balance between data consumption and generation.}

\section{RL Assistance Strategies Used in RL-EA}
\label{RL Assistance Strategies Used in RL-EA}

This section provides an introduction to improvement strategies for RL-assisted algorithmic search in RL-EA. Previous related work has attempted to enhance the search performance of EA by utilizing RL techniques, including generating solutions, learning objective functions, selecting algorithms/populations/operators, adapting parameters, and other aspects. \textcolor[rgb]{0,0,0}{This section also separately discusses the use of Deep RL (DRL)-assisted strategies.} By implementing specific strategies, RL can improve the search performance of these algorithms.

\subsection{Generating Solution}

The primary objective of this solution generation method is to generate a complete (or partial) solution available for use in an EA search. This is similar to certain enhanced forms of EA that employ domain knowledge for generating solutions in a heuristic manner. Unlike heuristics, RL can swiftly adapt to diverse problem scenarios by interacting with the environment. The interaction enables RL to update and undertake actions that are more adaptable to the new environment. There are two ways of employing RL for generating solutions to aid EA: one is directly generating solutions for the target problem, while the other involves obtaining solutions that serve as subproblems of the target problem. 

The utilization of RL in directly generating the solution to the objective problem can enhance the search efficiency of the algorithm, which has garnered a great deal of attention in academia. Zhang et al. proposed a Q-learning method for controlling solution generation in the particle swarm algorithm \cite{zhang2022multi}. The algorithm incorporated a reward function based on fitness improvement and employed a learning rate adaptive adjustment mechanism, enabling dynamic tuning of hyper-parameters according to the number of iterations. Based on the solution, \cite{sun2022reinforcement} explored the neighborhood structure of solutions using tabu search.

Sometimes, certain sub-problems within an optimization problem may require a significant amount of computational resources to obtain results, while the utilization of RL in generating solutions for such sub-problems can greatly reduce the overall solution time of the problem at hand. Based on this, Liu et al. proposed a hybrid algorithm that combined multi-objective evolutionary algorithms (MOEA) and RL to solve the traveling salesman problem (TSP) \cite{liu2022hybridization}. The RL approach employed a pointer network model to decode each individual within the population and generate the route of the salesman. This solution generation method can significantly reduce the iterative search time of the algorithm.

The benefit of implementing this strategy lies in the ease with which algorithms demonstrating exceptional performance in previous studies can be utilized. The aforementioned algorithms are ingeniously designed to tackle a wide range of optimization problems, including multi-objective optimization \cite{lin2022pareto}, railroad line planning \cite{khadilkar2018scalable}, resource management \cite{mao2016resource}, and production scheduling \cite{waschneck2018optimization,su2023evolution}. When designing a new RL-EA, we can easily specify the proportion of RL-generated solutions that account for individuals or populations.

\subsection{Learnable Objective Function}

Though the optimization objective function serves as the foundation for evaluating the performance of the EA, the traditional objective function remains unaltered throughout the EA search process. It only provides solely evaluation results without aiding in determining which direction to pursue. Particularly, a directionless search not only hinders finding an optimal solution for intricate optimization problems but also poses the risk of persistently searching in the wrong direction. Hence, a proficient objective function should possess dynamic adjustability to effectively guide the algorithm during its search process. The essence of a learnable objective function lies in incorporating the explored (or excavated) information and problem characteristics into the objective function, enabling the algorithm to search in an optimal direction. Inverse reinforcement learning (IRL), a specific type of RL, is commonly employed to acquire the reward function \cite{budhraja2017neuroevolution}. Unlike traditional RL, IRL does not necessitate a predefined reward function; instead, it is learned or estimated from a given environment. The reward function obtained by IRL exploration can be integrated with EA to guide the population search process.

IRL can enhance the search efficiency of EA by selecting from a diverse range of functions. For instance, in literature \cite{zhao2023inverse}, IRL had been incorporated into the moth-flame optimization (MFO) framework, which employs distinct MDP models and search strategies across sub-populations. The proposed algorithm utilizes four reward functions. With this search model, MFO can effectively tackle large-scale real-parameter optimization problems. 

The integration of IRL in EA offers distinct advantages for addressing problems that are challenging to formulate using optimization models. In literature \cite{liu2020driver}, the function obtained through IRL learning evaluated the performance and iteratively updated it by using real parameters. The proposed algorithm is improved based on the particle swarm optimization framework, which identifies the optimal cost function through the evaluation iteration mechanism.

\subsection{Algorithm Selection}
The process of algorithm selection is commonly employed in hyper-heuristics that consist of low-level heuristics (LLHs). The main task is to select from a range of LLHs that are beneficial for exploration and exploitation. These LLHs can be either simple heuristic rules or EA. As different algorithms cater to different types of optimization problems, it is easy to develop effective methods for multiple problems through their integration. However, the integration of these LLHs needs a necessary filter for the strategy. Adaptive selection of LLHs based on problem characteristics is the core of hyper-heuristic design. The utilization of RL for algorithm selection has gained popularity within this research domain.

Q-learning has been widely used in selecting LLHs, as evidenced by numerous studies \cite{choong2018automatic,zhao2022hyperheuristic,wu2023aq,zhang2023q,zhu2023hyper,shang2023green}. Choong et al. identified suitable LLHs for direct use in the current search phase \cite{choong2018automatic}, while Zhao et al. integrated six LLHs as options for RL action selection using a similar approach \cite{zhao2022hyperheuristic}. Such RL-assisted search ideas have also been employed in other works \cite{wu2023aq,zhang2023q,zhu2023hyper}. Furthermore, it was discovered that repairing solutions and conducting local searches after global searches can accelerate iterative algorithmic searches \cite{shang2023green}. Additionally, Q-learning can be used to select optimizers \cite{cheng2022multi}.

There have also been several endeavors to employ RL methods \cite{qin2021novel,zhang2022deep,tu2023deep}. In \cite{qin2021novel}, A3C was utilized for adaptive selection from the LLHs, which encompassed the artificial bee colony (ABC) algorithm, ant colony optimization (ACO) algorithm, cuckoo search (CS) \cite{yang2010engineering}, GA, PSO, and simulated annealing (SA) \cite{bertsimas1993simulated}. Moreover, \cite{zhang2022deep} and \cite{tu2023deep} employed DDQN and dueling double deep Q network respectively. In \cite{tu2023deep}, a feature fusion mechanism was used for feature design enabling RL to facilitate more precise decision-making.

The advantage of this integration strategy is that it is possible to constantly involve high-quality algorithms in the RL-EA framework. A new algorithm can be implemented with minimal or no modification of the existing ones. Additionally, the RL integrated into the algorithmic framework can adapt to more complex information interaction processes after incorporating new algorithms due to its powerful problem feature exploitation capability. Certainly, adding new LLHs into the framework may increase overall search time due to the high complexity. Therefore, it is essential to consider both search efficiency and effectiveness when making decisions on which LLHs to include in the algorithmic framework.

\subsection{Operator Selection}

Operator selection is also a common algorithmic improvement strategy that RL integrates into the EA framework. Usually, evolutionary operators are meticulously designed based on problem characteristics, which limits their applicability to specific scenarios. To enhance algorithm generalization, many studies employ multiple evolutionary operators for the proposed algorithms. Consequently, designing an adaptive operator selection mechanism becomes crucial. The ultimate goal of algorithm design is to achieve a concise adaptive mechanism that effectively reflects the performance of the operators. Motivated by these observations, various works have employed RL to adaptively select operators. Specifically, RL primarily performs operator selection for three types of evolutionary operations: crossover, mutation, and local search in EA. In the following, this paper will introduce each type of evolutionary operation according to the type of RL adaptive selection.

\begin{table*}[htbp]
	\centering
	\caption{RL-assisted operator selection strategy in existing literature}
	\label{operator selection}
	\begin{tabularx}{\textwidth}{m{1cm}m{3.5cm}m{2cm}m{9cm}}
		\toprule[1.5pt]
		Year & Authors & Type of RL & Operator Set \\
		\midrule[0.75pt]
		2014 & Buzdalova et al. \cite{buzdalova2014selecting} & Q-learning & Inversion mutation; tail inversion mutation; obstructive mutation \\
		2019 & Li et al. \cite{li2019differential} & Q-learning & DE/rand/1/bin; DE/best/1/bin; DE/rand to best/bin \\
		2022 & Fister et al. \cite{fister2022reinforcement} & Q-learning & DE/current-to-pbest/1/bin; DE/current-to-pbest-w/1/bin \\
		2023 & Li et al. \cite{li2023scheduling} & DDQN & DE/rand/1/bin; DE/best/1/bin; DE/rand/2/bin; DE/best/2/bin; DE/current to best/1/bin \\
		2023 & Zhang et al. \cite{zhang2023reinforcement} & Q-learning & Swap; forward insert; backward insert\\
		\bottomrule[1.5pt]                            
	\end{tabularx}
\end{table*}

\subsubsection{Adaptive Crossover}

\begin{figure*}[htp]
\centering
\includegraphics[width=0.75\textwidth]{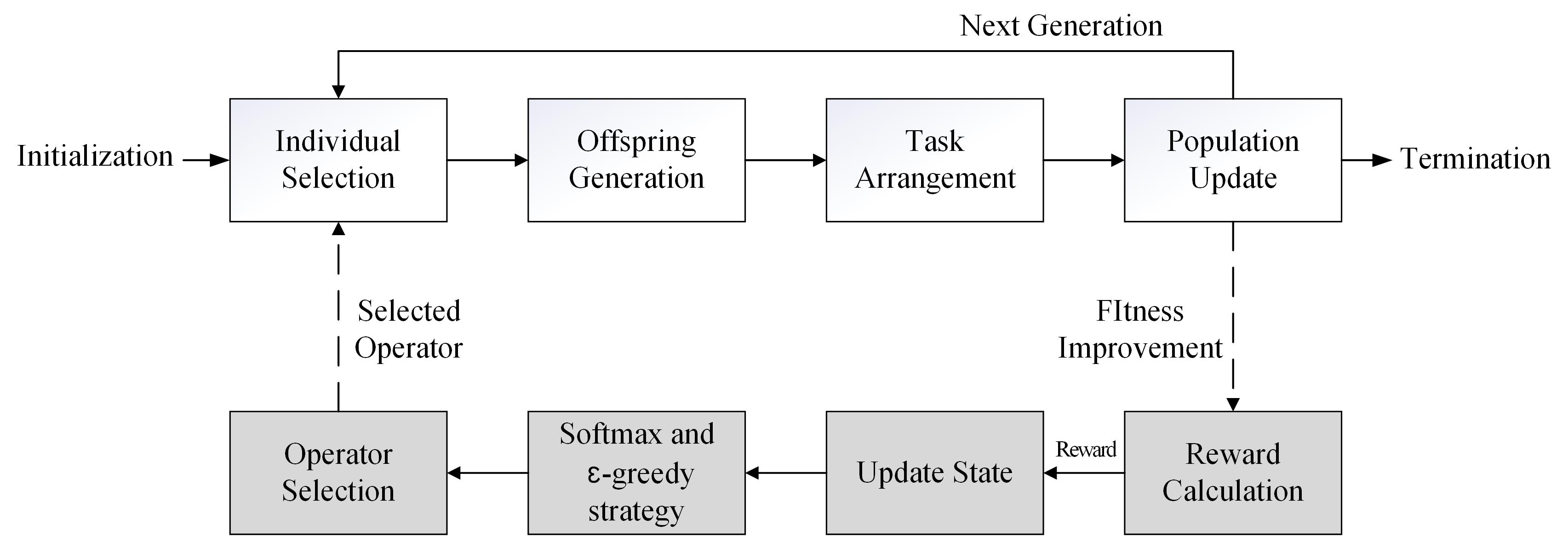}
\caption{Information flow between RL and GA framework \cite{song2023rl}}
\label{Information flow between RL and GA framework}
\end{figure*}

The crossover operation is the main operation of evolutionary search in various state-of-the-art algorithms, such as GA, NSGA-II, and multiobjective evolutionary algorithm based on decomposition (MOEA/D) \cite{zhang2007moea}. Among them, some previous studies have attempted to address real-world problems using RL-EA \cite{zhang2023reinforcement,song2023rl,tian2022deep}. For example, Song et al. proposed an improved GA based on RL \cite{song2023rl}. Fig.\ref{Information flow between RL and GA framework} illustrates the information flow between the RL and GA framework. In this proposed algorithm, Q-learning was employed to dynamically select crossover operators. There were seven crossover operators, including short segment crossover operator, medium segment crossover operator, long segment crossover operator, segment flip, foremost position crossover, and two heuristic crossover operators. The selection of operators based on the $\varepsilon$-greedy mechanism was conducted prior to the commencement of each generation in population search. Subsequently, simulation experiments were employed to evaluate the performance of the proposed algorithm in solving large-scale complex scenarios.

In addition to single-objective optimization EAs, Q-learning has also been applied in multi-objective optimization EAs. Zhang et al. proposed an RL-based NSGA-II (RL-NSGA-II) \cite{zhang2023reinforcement}, which incorporated five types of operators: two-point crossover, single-point crossover, order crossover, position-based crossover, and cycle crossover. The algorithm presented in this study, similar to \cite{song2023rl}, also defined the agent's state based on individual improvement. In both studies, the integration of Q-learning into the EA framework is relatively straightforward. However, for MOEA/D, a greater degree of algorithmic improvement is required for operator selection using RL. In \cite{tian2022deep}, a DQN-based operator adaptive selection strategy was used in the MOEA/D framework. Four crossover operators were employed as candidate evolutionary operations for individuals within each mating pool. Upon individual selection, a random operator was determined using the roulette method. Subsequently, a credit assignment approach was utilized to evaluate fitness improvement and determine the reward. 

\subsubsection{Adaptive Mutation}

The mutation is also a crucial evolutionary operation for multiple EA searches to acquire high-quality solutions. Unlike crossover, the probability of mutation occurrence and the magnitude of change to an individual are relatively low. Although there is no guarantee that the solution will necessarily improve in the optimization direction, this operator is indispensable for efficient global search in EA. In particular, mutation plays a crucial role in DE algorithms. RL can flexibly utilize these diverse operators based on problem characteristics, so ensemble mutation strategies are widely favored in related research. In \cite{li2019differential,fister2022reinforcement,li2023scheduling}, two to three mutation operators were selected as actions for the DE search by RL. This approach maintains the low computational complexity and excellent global search performance of DE.

The study conducted by Buzdalova et al. \cite{buzdalova2014selecting} focused on the selection of mutation operators for a generalized EA framework using Q-learning. They also provided a comprehensive validation of the effectiveness of integrating \textit{softmax}, $\varepsilon$-greedy strategies in RL. Furthermore, apart from the commonly used operators in SOTA algorithms, heuristic operators closely related to the problem to be solved can be considered as options for agent action selection \cite{zhang2023reinforcement}. To facilitate clarification, Table \ref{operator selection} presents a summary of relevant research on RL adaptive mutation operator selection.

\subsubsection{Adaptive Local Search}
\textcolor[rgb]{0,0,0}{The exploitation ability of the algorithm can be enhanced by incorporating additional local search (LS) into population search for exploring the local solution space in complex optimization problems \cite{karimi2023learning,ren2023novel}. }Previous studies have extensively utilized ABC to improve algorithmic performance by using RL to select LS operators \cite{wang2022adaptive,li2022improved,zhao2022reinforcementa,zhou2023adaptive}. The assembly scheduling problem was addressed by Wang et al., who proposed a three-stage ABC with Q-learning (QABC) approach \cite{wang2022adaptive}. In the QABC, Q-learning consisted of twelve distinct states based on search performance and six different actions defined by a combination of four neighborhood structures. In addition to the general knowledge of optimization problems, problem-specific knowledge can also be utilized in the design of local search operators \cite{li2022improved}. Of course, using both types of knowledge in the design of local search operators has been proven to be more effective for specific problems \cite{zhao2022reinforcementa}. In \cite{zhao2022reinforcementa}, a Q-table that integrates operator number, Q-score, operator usage count, and operator success count was defined. Accordingly, this work proposes a simplified method for updating Q-values. In addition, the algorithmic framework for action selection by RL can also incorporate destroy and repair operators, similar to those used in adaptive large neighborhood search (ALNS) \cite{zhou2023adaptive}.

RL can also be integrated into EA based on the GA framework. Zheng et al. attempted to solve the traveling salesman problem (TSP) \cite{zheng2023reinforced} using an RL-assisted GA. This proposed algorithm employed Q-learning to update the salesman's city visit order. In addition to being integrated into single-objective GA, RL can also be applied in the multi-objective GA framework. In \cite{qi2022qmoea}, Q-learning was integrated into NSGA-II to perform an adaptive selection of five neighborhood path optimization operators. Unlike \cite{qi2022qmoea}, \cite{li2023muti} used general variable neighborhood search to further enhance the individual's neighborhood structure after utilizing Q-learning for selecting the neighborhood search operators. 

Furthermore, several studies have incorporated RL into other EAs \cite{du2022knowledge,yan2023novel,zhao2022reinforcementb}. Du et al. employed a hybrid approach in their research, where half of the population utilized a distribution estimation algorithm (EDA) for global search, while the other half used DQN for local search \cite{du2022knowledge}. State features of the agent were defined based on metrics related to problem size, optimization objective, and scheduling compositions. Yan et al. incorporated Q-learning into the tuna swarm optimization algorithm (TSOA) to enhance search operator selection through four local search strategies: spiral foraging, parabolic foraging, optimal adjustment, and enhanced symbiotic organism search (ESOS) \cite{yan2023novel}. In addition to single-population EA, bi-population algorithms can also utilize RL to control local search. For example, Zhao et al. proposed a bi-population cooperative EA (BPCEA) \cite{zhao2022reinforcementb}, which was inspired by bird migration for population search and adopted distinct search strategies in two sub-populations. Correspondingly, double Q-learning selected local structure to update operators for the sub-populations respectively.

The evaluation of the efficacy of ensemble RL-based LS strategies in different EAs can offer valuable algorithmic design insights for other researchers. Gao et al. compared the optimization performance of hybrid algorithms (HAs) that combined five different algorithms, GA, ABC, Jaya, PSO, and harmony search (HS), with three distinct Q-learning-based LS strategies, respectively \cite{gao2023ensemble}. Six LS operators were used to update an individual's neighborhood structure based on agent decision-making. Extensive numerical experiments demonstrated that the combination of PSO and Q-learning yielded optimal optimization performance. 

In addition, the algorithm can adopt perturbation operators as a strategic measure to avoid local optima, which can be selected from the candidate operators using RL. Zhao et al. conducted an analysis of the problem and designed two classes of perturbation operators with a total of five heuristics \cite{zhao2023knowledge}. Subsequently, a kind of Q-learning was employed to select the appropriate perturbation operator. The state of the agent in Q-learning was defined by whether or not there were changes in population performance. Following the perturbation operation, the reward was updated using +1/0/-1 based on fitness changes. The proposed algorithm updated the known optimal solution in 71\% of benchmark instances. 

Without exception, all RL-EAs that use RL for operator selection employ value-based RL. Specifically, Q-learning is utilized in the vast majority of studies, with a few studies implementing double Q-learning, DQN, and DDQN. This phenomenon can be ascribed to the constraint that operators can only be selected from a limited set of discrete candidate options. The utilization of policy-based RL not only necessitates the inclusion of an additional decision-making process but also poses challenges in ensuring the efficacy of agent learning. Therefore, employing policy-based RL is not a reasonable approach. Hence, prioritizing the adoption of the value-based RL method for operator selection in conjunction with RL-assisted EA is recommended. 

\subsection{Sub-population Selection}
Traditional EA relies on a single population for exploring the solution space. Such a composition of the population structure may result in a tendency for the structure of individuals within the population to become increasingly similar as the search process continues. To enhance the diversity of individuals within the population, researchers have endeavored to employ multiple sub-populations collaboratively to collectively solve optimization problems. The synergy among these sub-populations can manifest through competition, cooperation, or a combination of both competitive and cooperative relationships. Therefore, the selection of the sub-population and the composition of individuals within it play a crucial role in determining the optimization results of a multiple-population algorithm consisting of multiple sub-populations.

Cooperative multiple population-based EA can be dynamically adjusted using RL. Jia et al. proposed a multiple population-based MA \cite{jia2023q}. The algorithm contained seven sub-populations with respective crossover strategies. To select the composition of individuals within each sub-population, a specialized Q-learning approach tailored to the problem characteristics was used. More specifically, the Q-learning employed strategies of increasing, decreasing, and maintaining a constant number of individuals within each sub-population to optimize solution performance. This algorithm has been effectively validated through simulation experiments and real-life examples. 

\begin{figure}[htp]
\centering
\includegraphics[width=0.4\textwidth]{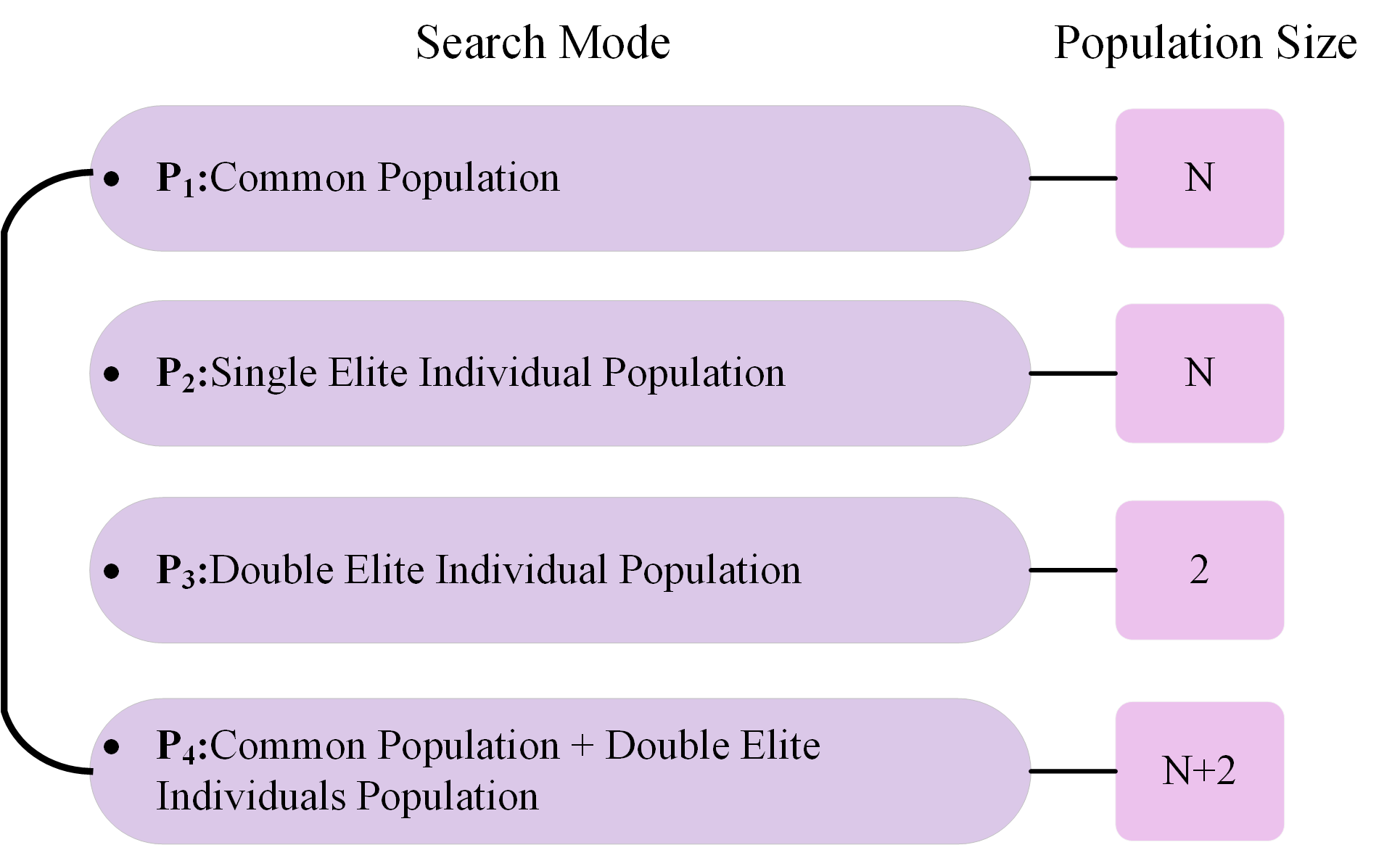}
\caption{An example of competitive sub-populations \cite{guo2023reinforcement}}
\label{An example of competitive sub-populations}
\end{figure}

The sub-populations utilized in the search can be determined by a competition-based multiple population EA through RL. Guo et al. embedded RL and an ensemble population strategy (EPS) into a genetic programming (GP) framework, named RL-assisted GP (RL-GP) \cite{guo2023reinforcement}. As illustrated in Fig.\ref{An example of competitive sub-populations}, the EPS comprises four competing sub-populations, with one being selected for each round of iterative search. These populations include the common population, the single/double elite population, and a novel population consisting of both common and elite individuals.

\subsection{Parameter Adaptation}

A well-designed EA should primarily focus on global exploration during the initial stages of the search and gradually shift toward local exploitation as the search progresses. Since the choice of search mode employed by an EA heavily relies on its control parameters, the control parameters play a vital role in the search performance of EAs \cite{eiben2012evolutionary}. Traditional EA, however, utilizes only one parameter in the iterative search process. Such a singular parameter setting can easily restrict the algorithm's applicability to specific problem scenarios. Consequently, a range of tuning parameter methods have been proposed. Although these methods enhance the algorithm's search performance to some extent, they suffer from drawbacks such as complex tuning processes and the involvement of too many parameters \cite{eiben2012evolutionary}. In contrast to the aforementioned methods, RL possesses the capability to dynamically adjust strategies in accordance with specific problems and algorithmic search performance \cite{shahrabi2017reinforcement}. Consequently, some researchers have employed RL for adaptive control parameter adjustment in EAs. Table \ref{parameter adaptation} shows various algorithms, such as DE, GA, and NSGA-II, that have attempted to utilize RL to adaptively adjust parameters. In the following, the parameter control within the two types of algorithms, DE and GA, is mainly described in detail:

\begin{table*}[htbp]
\centering
\caption{RL-assisted parameter adaptation strategy in existing literature}
\label{parameter adaptation}
\footnotesize
\begin{tabularx}{\textwidth}{m{1cm}m{3.5cm}m{3cm}m{2cm}m{5.5cm}}
\toprule[1.5pt]
Year & Authors & Type of EA & Type of RL & Controlled Parameters \\
\midrule[0.75pt]
2013 & Rakshit et al. \cite{rakshit2013realization} & MA & Q-learning & Scaling factor \\
2014& Karafotias   et al.  \cite{karafotias2014generic} &  Simple self-coded Evolution   Strategy, cellular GA,  GA with Multi-Parent Crossover (GA MPC) & Q-learning & ES: population size, generation gap, mutation step size, the tournament size for survivor selection; Cellular GA:  crossover rate,   mutation rate, mutation variance; GA MPC: population size, the maximum size of the parent tournament, randomization probability, percentage of the population put in the archive\\
2018 & Sadhu et al. \cite{sadhu2018synergism}       & Firefly algorithm & Q-learning & Fixed light absorption coefficient, randomization parameter \\
2020& Kaur and   Kumar \cite{kaur2020reinforcement} &  NSGA-II & Q-learning & Crossover rate, mutation rate \\
2021& Huynh et   al. \cite{huynh2021q} &  Brief DE & Q-learning & Scale factor, crossover rate \\
2021& Hu et   al. \cite{hu2021reinforcement} & DE & Q-learning & Scale factor \\
2021 &Sun et   al. \cite{sun2021learning} &  DE & PG (network: LSTM) & Scale factor, crossover rate \\
2022& Tessari   et al. \cite{tessari2022reinforcement} &  Covariance matrix adaptation evolution strategies (CMA-ES) \cite{hansen2003reducing}, DE & PPO & CMA-ES: Step-size; DE: scale factor, crossover rate \\
2022 & Zhang et   al. \cite{zhang2022variational} & DE & Variational PG & Scale factor, crossover rate \\
2022& Cheng et   al. \cite{cheng2022scheduling} &  GA & Q-learning & Crossover rate, mutation rate \\
2022 & Li et   al. \cite{li2022learning} & MA & Q-learning & Size of the neighborhood \\
2022 & Li et   al. \cite{li2022reinforcement} & MOEA/D & Q-learning & Neighborhood number \\
2022 & Shiyuan Yin   \cite{shiyuan2022reinforcement} &  PSO & DDPG & Three linearly increased/reduced parameter \\
2023& Peng et   al. \cite{peng2023reinforcement} &  DE & Q-learning & Two trigger parameters of local search \\
2023 & Liu et   al. \cite{liu2023learning} & DE & PG (network: LSTM) & Scale factor, crossover rate \\
2023& Li et   al. \cite{li2023scheduling} &  Multi-objective DE & DDPG & Scale factor, crossover rate \\
2023& Song et   al. \cite{song2023laga} &  GA & DDQN (network: GRU) & Crossover rate, mutation rate\\
2023& Li et al. \cite{li2023reinforcement} &  PSO & Q-learning & Particle update rate \\
2020, 2023 & Tatsis and Parsopoulos \cite{tatsis2020reinforced,tatsis2023reinforcement} & DE & REINFORCE algorithm & Scale factor, crossover rate\\
2023 & Gao et al. \cite{gao2023improved} & PSO & Q-learning & Inertia weights and acceleration parameters \\
\bottomrule[1.5pt]                            
\end{tabularx}
\end{table*}

$\bullet$ \textbf{DE}: The scale factor is a crucial control parameter in DE, which can be determined by Hu et al. \cite{hu2021reinforcement} through increasing or decreasing the adjustment value on top of the base value. Additionally, many studies have explored adaptive control of crossover probabilities. For example, Huynh et al. \cite{huynh2021q} chose from seven sets of hyper-parameter configuration combinations. Peng et al. \cite{peng2023reinforcement} dynamically adjusted the two triggering parameters through local search combined with DE. In contrast to the aforementioned selection method based on predetermined parameters, \cite{li2023scheduling, sun2021learning, zhang2022variational, liu2023learning} proposed more precise parameter tuning. In these studies, the policy-based RL directly outputs the values of the control parameters through the network model.

\begin{figure}[htp]
\centering
\includegraphics[width=0.5\textwidth]{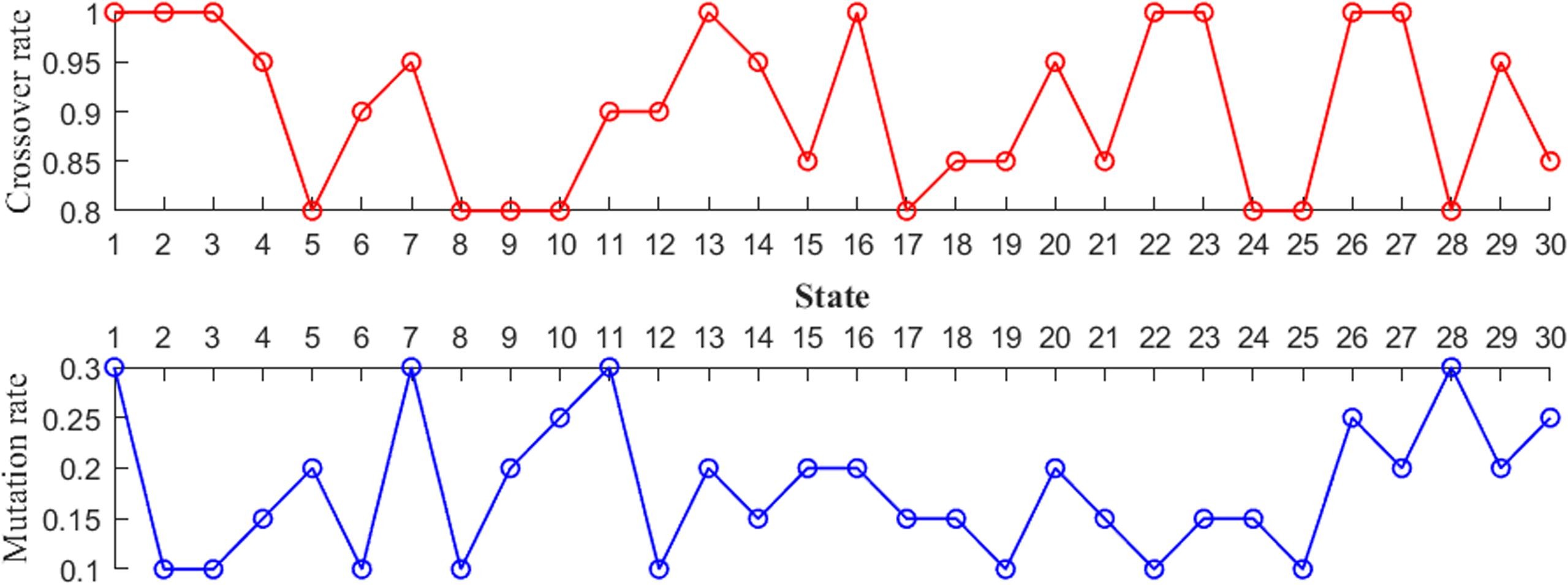}
\caption{Parameter values for crossover and mutation probabilities \cite{cheng2022scheduling}}
\label{Parameter values for crossover and mutation probabilities}
\end{figure}

$\bullet$ \textbf{GA}: The control of crossover probability and mutation probability can help achieve parameter adaptation for GPs. Cheng et al. \cite{cheng2022scheduling} employed the Q-learning method to select from 18 sets of parameter combinations determined by orthogonal experiments for population search. The parameter values, depicted in Fig.\ref{Parameter values for crossover and mutation probabilities}, exhibit a decreasing trend in crossover probability and an increasing trend in mutation probability overall. The crossover rate and mutation rate can affect the diversity of the population. When the mutation rate is high, the search strategy shifts towards exploration while reducing the crossover probability to decrease computation time. Conversely, when the mutation rate is low, a higher crossover rate enhances algorithmic exploitation capabilities \cite{cheng2022scheduling}. Apart from this, the control parameters can also be regulated by PG. Song et al. \cite{song2023laga} considered the population evolution process as a time series and employed the gated recurrent unit (GRU) model to estimate the crossover probability and mutation probability. Furthermore, RL is equally applicable in improved algorithms that integrate LS \cite{li2022learning}.

\subsection{Other Strategies}

\begin{table*}[htbp]
	\centering
	\caption{RL-assisted other strategies in existing literature}
	\label{other strategies}
	\begin{tabularx}{\textwidth}{m{1cm}m{3.5cm}m{2cm}m{9cm}}
		\toprule[1.5pt]
		Years & Authors & Type of RL & Role of RL \\
		\midrule[0.75pt]
		2012 & Buzdalova and Buzdalov \cite{buzdalova2012increasing} & Q-learning & Choosing auxiliary fitness functions \\
		2020 & Huang et al. \cite{huang2020fitness} & Q-learning & Determining the optimal probability distribution of the algorithm’s search strategy set \\
		2021 & Xia et al. \cite{xia2021reinforcement} & Q-learning & Selection of subspace \\
		2021 & Radaideh and Shirvan \cite{radaideh2021rule} & PPO & Matching some of the problem rules/constraints \\
		2022 & Wang et al. \cite{wang2022reinforcement} & Q-learning & Determining the level of each individual \\
		2023 & Gao et al. \cite{gao2023efficient} & Actor-Critic & Mining sparse variables to reduce the problem dimensionality \\
		2023 & Zhou et al. \cite{zhou2023improved} & DQN & Choosing candidate individual \\
		2023 & Liu et al. \cite{liu2023NeuroCrossover} & PPO & Selection of genetic locus\\
            2023 & Qiu et al. \cite{qiu2023q} & Q-learning & Choosing multi-exemplar selection strategy \\
		\bottomrule[1.5pt]                            
	\end{tabularx}
\end{table*}

The preceding parts outline a variety of strategies for RL-assisted EA search, which have yielded exceptional performance in diverse optimization problems when implemented through proposed RL-EAs. In addition to the aforementioned strategies, there exist other equally effective RL-assisted strategies for achieving global or local searches in EA, which are presented in Table \ref{other strategies}. These research findings primarily focus on designing the corresponding algorithms based on specific problem scenarios and have achieved superior optimization performance. Some ingenious designs for problems, fitness evaluation, and individual processing are well worth studying and learning from. 

The analysis and processing of the problem solution space can also effectively enhance the search efficiency of algorithms. \cite{xia2021reinforcement} employed Q-learning to select the subspace within the problem domain, thereby enabling a more targeted DE search process. Similarly, \cite{radaideh2021rule} leveraged RL to reduce the solution space by matching problem rules/constraints. Furthermore, \cite{gao2023efficient} optimized algorithmic searches from a variable perspective by using RL to exploit sparse variables to reduce the problem dimension. In addition, \cite{buzdalova2012increasing} proposed an RL-based mechanism for the selection of fitness assessment methods. This mechanism employed multiple auxiliary fitness functions, which are utilized by RL decisions to evaluate population performance and enhance algorithm efficiency. 

\textcolor[rgb]{0,0,0}{\subsection{Discussion of DRL-assisted Strategies}}

\textcolor[rgb]{0,0,0}{In recent years, deep neural networks have emerged as a prominent class of methods in RL, garnering significant attention from researchers. These methods have also demonstrated strong performance in aiding EA search, enabling various auxiliary tasks such as operator adaptive selection, solution generation, and parameter optimization. The key advantages of DRL over simpler RL methods lie in its better reasoning capabilities, ability to adapt to the environment, and ability to realize more accurate algorithmic control. For example, using a simple RL approach to dynamically adjust the control parameters of the EA is limited to selecting from a predefined combination of parameters \cite{li2023evolutionary}. In contrast, DRL enables setting the network model within reasonable intervals based on predicted output values \cite{song2023laga}. Such flexible parameter settings facilitate efficient and diverse searches of the EA.}

\textcolor[rgb]{0,0,0}{Consequently, training network models for DRL necessitates substantial amounts of data, significant time and computational resources investment, and may also encounter instances where the model outcomes fail to meet the anticipated results. The requirement for extensive data consumption is an inherent characteristic of DRL training, which compels researchers to prepare copious additional datasets or problem scenarios. Despite having sufficient data, the trained model may not always yield satisfactory results, posing a challenge for researchers to accept that their efforts do not necessarily align with the expected performance of the model. DRL adopts a deep neural network model, which makes the method a "black box" and it is difficult to guarantee the optimal training effect of the model. Furthermore, it is worth noting that relying solely on deep models for predictions in DRL can lead to a lack of environmental perception and reflection. Consequently, the resulting network model exhibits limited generalization ability, making it challenging to address novel problems effectively. It is also prudent not to excessively invest efforts in training DRL models since the algorithm's core lies in population search, with DRL playing a supplementary role. Therefore, researchers should prioritize evaluating the overall performance of the algorithm rather than solely pursuing optimal DRL performance.}

\textcolor[rgb]{0,0,0}{It is challenging to assert that the integration of DRL within an EA framework will inevitably surpass the performance of simple RL. The selection between DRL and structurally straightforward RL should be based on task-specific requirements. For tasks with low complexity, a basic RL approach suffices. However, if DRL is employed, there exists an additional trade-off between training cost and model effectiveness. Moreover, certain complex tasks or those with extensive state spaces may pose difficulties for simple RL but not for DRL. In some cases, both simple RL and DRL might exhibit similar effectiveness; thus, researchers need to consider factors such as training cost, utilization efficacy, and method generalizability when determining the appropriate approach.}

\section{Attribute Settings of RL in RL-EA}
\label{Attribute Settings of RL in RL-EA}
 In addition to the different strategies used by the RL to support the EA search, other attribute settings (i.e., state, reward) in the RL have also been designed to be problem-specific in related work. These attributes have a dual impact on both the agent's decision-making process and the training performance of the RL. This section provides an overview of the adopted settings for the state and reward in RL. \textcolor[rgb]{0,0,0}{In addition, the principles of action setting are discussed in this section.}

\subsection{Setting of State}
The state serves as the input for the agent to make decisions, which will be directly associated with the selected action. For RL to assist EA search effectively, it is crucial that the state information accurately reflects the latest progress of the algorithm's search. One direct approach to achieve this is by using the chromosome structure as the state representation, as demonstrated in \cite{karafotias2014generic}. This type of state representation directly reflects population optimization. However, it also faces the problem that a large amount of information being included in the Q-table will impact agent sampling efficiency. To avoid this problem, many studies such as \cite{hu2021reinforcement,song2023rl} set the overall performance of the population as the state representation. However, when it comes to representing state information in finite dimensions (i.e. continuous state), Q-learning is almost powerless. To address this limitation, Ref. \cite{cheng2022multi} proposed a discretized approach for representing continuous states by dividing them into subintervals. While this representation enabled Q-learning to handle continuous states to some extent, it still suffered from the drawback of only applying to fewer state dimensions. In contrast to Q-learning, the utilization of ANN models in RL methods eliminates the decision-making challenges caused by the expansion of state dimensions. Incorporating abundant statistical information into ANN inputs, as demonstrated in \cite{sun2021learning, qin2021novel}, enhanced prediction accuracy. Moreover, the strong predictive capability exhibited by ANN empowers this type of RL to effectively tackle various complex tasks. 

Usually, the states used as the basis for the agent's choice of actions are inconsistent with the actions. While a state-action-consistent RL modeling approach has been applied in individual studies, such as Ref. \cite{wang2022reinforcement}. The authors categorized particles in PSO into different levels and dynamically adjusted them using RL. In this framework, the states and actions of the agent were the number of levels within the population. According to this setting, the Q-table was a square array with an equal number of states and actions, which can quickly feedback the results of the action decisions to the agent.

\subsection{Setting of Reward Function}
The reward is a feedback signal obtained by the agent after interacting with the environment, which influences subsequent decision-making. In previous studies related to RL-EA, the reward function primarily includes two forms: direct calculation based on fitness value changes and acquisition through specific mapping relationships. For instance, Refs. \cite{sun2021learning,guo2023reinforcement} obtain it by using the difference calculation of the optimal fitness values within two generations of populations. This calculation directly employed the metrics used for evaluating the performance of the RL-EA search. Considering the limitations of this evaluation method, Ref. \cite{xia2021reinforcement} additionally introduces a reward adjustment operator based on the frequency of occurrence in the subspace, which was able to balance exploration and exploitation. Ref. \cite{karafotias2014generic}, on the other hand, uses the ratio method to compute the rewards. In addition to the aforementioned reward computation, the use of mapping specific reward values based on the performance has also been employed in the literature \cite{hu2021reinforcement,qiu2023q,yan2023novel}. This approach facilitates a smoother change in reward values, thereby enhancing the stability of RL across diverse scenarios. Furthermore, for specialized RL algorithms like IRL, the best reward function is autonomously explored by the agent performance during the algorithm training process \cite{liu2020driver,zhao2023inverse}.

\textcolor[rgb]{0,0,0}{The evaluation of rewards is a crucial factor for researchers to consider when determining the reward function. The choice of using either the population or individual as the basis for calculation in the reward function has been explored by several studies. Refs. \cite{tian2022deep,liu2023NeuroCrossover,song2024generalized}, consider the population as a whole, calculating the reward value based on the overall population performance improvement. This method ensures that reward value calculations remain independent of population size, allowing for consistent evaluations at any scale. Conversely, Refs.\cite{song2023rl,gao2023improved} assess an individual's optimization performance immediately after completing evolutionary operations. In such cases, an increase in population size leads to a proportional increase in reward evaluations.}

In summary, the setting of states and rewards is not as easy as actions based on what the RL is trying to achieve, but have to be tried and tested based on the problem and the specific performance of the RL. Identifying these attributes of RL can be a time-consuming task for researchers, which is challenging to avoid. A simpler approach involves leveraging prior studies to emulate and enhance specific aspects, such as incorporating problem-specific features and performing a series of processes on the outcomes. To make it easier for readers to design novel RL-EAs, we will introduce the applications of RL-EAs in different fields in the next section.

\textcolor[rgb]{0,0,0}{\subsection{Principle of Action setting}}
\textcolor[rgb]{0,0,0}{The RL-assisted strategies discussed in section \ref{RL Assistance Strategies Used in RL-EA} also involve the agent's action selection process. The design of actions should adhere to the principles of simple structure, fully integrating the EA search methods and features and improving the performance of the algorithm. Simplicity is a fundamental requirement for action design, ensuring that it does not significantly increase the overall space and time complexity of the algorithm. For example, Refs. \cite{tian2022deep,song2023laga,liu2023NeuroCrossover} enable the agent to select an action prior to each search generation. This operation has the minimum impact on the overall search process of the algorithm. Since RL consumes negligible computational resources during testing phases, designers only need to ensure that agents can perform actions without excessive concerns.}

\textcolor[rgb]{0,0,0}{Adequately integrating EA search methods with features is a crucial factor to be considered in designing the action. The incorporation of RL into EA for searching should align with the originally designed algorithmic framework's search strategies and patterns. Taking GA as an example, it encompasses various steps such as initialization, selection, crossover, mutation, and generation of offspring population. Several studies have enhanced this algorithm by incorporating RL methods based on GA's search characteristics. Refs. \cite{zhang2023reinforcement,tian2022deep} try to propose more diverse search strategies. Refs. \cite{li2023scheduling,eiben2012evolutionary} enhance the search performance of GA by optimizing the balance between exploration and exploitation in algorithmic search. On the other hand, Refs. \cite{zhao2022reinforcementa,zhou2023adaptive} employ neighborhood search to address the limited search capability in the local solution space.}

\textcolor[rgb]{0,0,0}{The primary objective of designing the action is to enhance algorithmic performance, irrespective of the RL-assisted strategies employed. The aim remains to facilitate efficient exploration of the solution space by EA. For this purpose, the researcher should primarily focus on quantifying the extent to which the algorithm with the improved strategy enhances performance compared to the original algorithm. In particular, when faced with multiple alternative actions, it is advisable to prioritize the strategy that exhibits the most significant enhancement in algorithm performance. Occasionally, this pursuit of improved performance necessitates a complex RL-assisted strategy, demanding researchers to strike a balance between performance and complexity.}

\section{Application of RL-EA}
\label{Application of RL-EA}

\begin{figure*}[htp]
\centering
\includegraphics[width=0.75\textwidth]{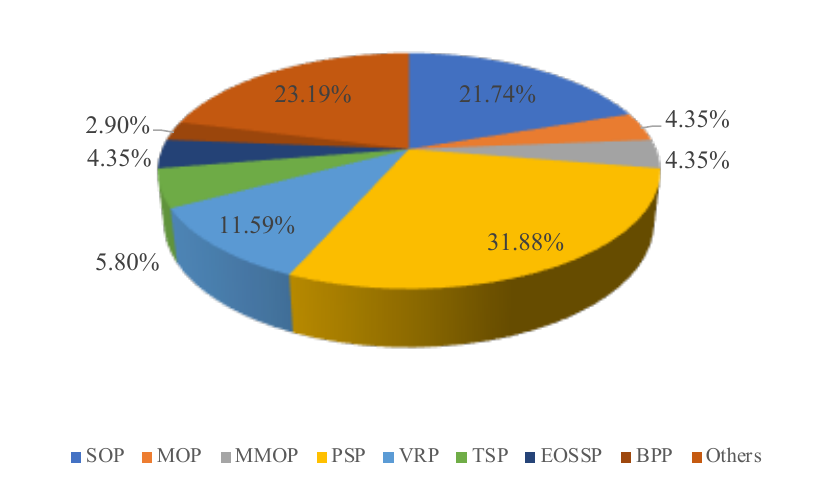}
\caption{Application of RL-EA in different domains (SOP: single objective optimization problem, MOP:multi-objective optimization problem, MMOP: multimodal problem, PSP:
production scheduling problem, VRP: vehicle routing problem, TSP: traveler salesman problem, EOSSP: earth observation satellite scheduling problem, BPP: bin packing problem)}
\label{Application of RL-EA in different domains}
\end{figure*}

The RL-EA algorithm has undergone extensive testing across various fields. These successful algorithmic solution ideas can be applied to other studies aiming to address optimization problems using RL-EA. Fig. \ref{Application of RL-EA in different domains} illustrates the application of RL-EA in various domains. In this section, we present the advancements made in applying RL-EA by categorizing the problem-solving process based on variable value types.

\subsection{Continuous Optimization Problems}

\begin{table*}[htbp]
	\centering
	\caption{Application in continuous optimization problems}
	\label{continuous optimization problem}
	\begin{tabularx}{\textwidth}{m{1.5cm}m{3cm}m{3cm}m{5cm}m{3.5cm}}
		\toprule[1.5pt]
		Year & Authors & Problem & Type of EA & Type of RL \\
		\midrule[0.75pt]
            2013 & Rakshit et al. \cite{rakshit2013realization} & SOP & MA  & Q-learning \\
		2014 & Karafotias et al.  \cite{karafotias2014generic} & SOP & Evolution Strategy, Cellular GA & Q-learning \\
            2018 & Sadhu et al. \cite{sadhu2018synergism}      & SOP & Firefly algorithm & Q-learning \\
		2019 & Li et al. \cite{li2019differential} & MMOP, MOP & DE & Q-learning \\
		2020 & Huang et al. \cite{huang2020fitness} & MOP & MODE & Q-learning \\
		2021 & Xia et al.  \cite{xia2021reinforcement} & MMOP & DE & Q-learning \\
		2021 & Sun et al. \cite{sun2021learning} & SOP & DE & PG \\
		2022 & Zhang et al. \cite{zhang2022variational} & SOP & DE & Variational PG \\
		2022 & Fister et al. \cite{fister2022reinforcement} & SOP & DE & Q-learning \\
		2022 & Tian et al. \cite{tian2022deep} & MOP & MOEA/D & DQN \\
		2022 & Tessari et al. \cite{tessari2022reinforcement} & SOP & CMA-ES, DE & PPO \\
		2022 & Yin \cite{shiyuan2022reinforcement} & SOP & PSO & DDPG \\
		2022 & Wang et al. \cite{wang2022reinforcement} & SOP (large scale) & PSO & Q-learning \\
		2023 & Gao et al. \cite{gao2023efficient} & Sparse MOP & MOEA & Actor-critic method \\
		2023 & Liu et al. \cite{liu2023learning} & SOP & DE & PG \\
		2023 & Li et al. \cite{li2023reinforcement} & SOP & PSO & Q-learning \\
		2023 & Zhao et al.  \cite{zhao2023inverse} & SOP & Moth-flame optimization algorithm & Inverse RL\\
            2020, 2023 & Tatsis and Parsopoulos \cite{tatsis2020reinforced,tatsis2023reinforcement} & SOP, MMOP & DE & REINFORCE algorithm\\
            2023 & Qiu et al. \cite{qiu2023q} & MMOP, SOP & PSO & Q-learning \\
            2023 & Zhao et al. \cite{zhao2023multi} & SOP (large scale) & ABC & Q-learning \\
		\bottomrule[1.5pt]
	\end{tabularx}
\end{table*}

Continuous optimization problems, also known as numerical optimization problems, are problems that attempt to find an optimal solution (maximum or minimum) or a Pareto solution within the solution space. These problems involve an optimization objective function and a domain of definition for the independent variables. As research on this subject progresses, various problem forms have emerged including multimodal optimization problems, multitasking optimization problems, and constrained optimization problems. In the subsequent discussion, we will provide a concise introduction to single-objective optimization problems (SOPs), multi-objective optimization problems (MOPs), and multimodal problems (MMOPs), which are commonly encountered when employing RL-EA in solving continuous optimization problems. 

$\bullet$ \textbf{Single-objective optimization problem}: The single-objective optimization problem is the fundamental class of continuous optimization problems, characterized by a sole objective function. Depending on the problem set, the optimization objective can be either to find the maximum or the minimum value of this function. Typically, an SOP (we consider minimization) can be formulated as:
\begin{equation}
\min_{\mathbf{x}\in\Omega}f(\mathbf{x})
\end{equation}
where $ \Omega\ $ is the feasible solution set, $\Omega=\{\mathbf{x}\in\mathbb{R}^D|g_i(\mathbf{x})\geq\\0,1\leq i\leq m,h_j(\mathbf{x})=0,1\leq j\leq n\} \ \text{or} \ \Omega=\bigcup[a_i,b_i]$.\\

$\bullet$ \textbf{Multi-objective optimization problem}: A multi-objective optimization problem is a continuous optimization problem with multiple sub-objective functions constructed on the basis of SOP. These sub-objectives have conflicting relationships with each other, and thus the algorithm's optimization results cannot be simply represented by a solution and an objective function value. Such problems can only use Pareto solution sets to represent the search results. \textcolor[rgb]{0,0,0}{Unlike the solution of a single-objective optimization problem, a set of solutions can be obtained for a multi-objective optimization problem. Consequently, the set of non-dominated solutions (Pareto solution set) can be obtained from this pool as potential solution candidates. Specifically, non-dominated refers to a solution that is not dominated by any other solutions, meaning there is no superior solution to it. A collection of such solutions forms the set of non-dominated solutions.} Here, we give the mathematical formulation of the MOP problem of minimizing the optimization objective as an example.

\begin{equation}
\begin{aligned}
    &\text{Minimize}: \mathbf{f}(\mathbf{x})=(f_1(\mathbf{x}),\ldots,f_M(\mathbf{x})) \\
    &\text{Subject to}: l_d\leq x_d\leq u_d, d=1,\ldots,D.
\end{aligned}
\end{equation}

$\bullet$ \textbf{Multimodal optimization problem}: The class of multimodal optimization problems (MMOPs) frequently encountered in real-world scenarios can be categorized as either single-objective MMOPs or multi-objective MMOPs. There are multiple optimal solutions in the MMOPs simultaneously. This class of problems can be formulas expressed by the previously introduced SOP or MOP \cite{wang2014mommop}.

It is evident that each class of continuous optimization problems exhibits distinct characteristics, imposing stringent demands on the algorithms' generalizability. Therefore, numerous researchers have concentrated their efforts on enhancing algorithms to achieve optimal performance across multiple problem scenarios. Among the many algorithms available for solving optimization problems, recent research has shown interest in employing RL strategies to augment the algorithm's information acquisition capabilities. The results presented in Table \ref{continuous optimization problem} indicate that the application of RL-EA is the most prevalent in SOP for continuous optimization problems, while the number of studies related to MOP and MMOP is approximately equal. In terms of the RL model, the Q-learning method emerges as the most commonly employed method due to its simple structure and strong ability to explore uncharted environments. \textcolor[rgb]{0,0,0}{Furthermore, the optimization performance of the improved algorithm for adaptive operator selection using DQN based on the MOEA/D framework (MOEA/D-DQN) \cite{tian2022deep} in the MOP benchmark is presented in Table \ref{IGD/HV values obtained by seven multi-objective EAs on benchmark suites/neural network training}.} From the results, it can be observed that MOEA/D-DQN demonstrates superior overall performance across 34 benchmark functions. This algorithm effectively balances global and local search through the adaptive selection operator.


\begin{sidewaystable*}[htbp]
	\scriptsize
	\centering
	\caption{IGD/HV values obtained by seven multi-objective EAs on benchmark suites/neural network training \cite{tian2022deep}}
	\label{IGD/HV values obtained by seven multi-objective EAs on benchmark suites/neural network training}
	\begin{threeparttable}
	\begin{tabular}{cccccccc}
		\toprule[1.5pt]
		Problem & MOEA/D & MOEA/D-DRA & IM-MOEA & SMEA & MOEA/D-FRRMAB & MOEA/D-DYTS & MOEA/D-DQN \\
		\midrule[0.75pt]
		ZDT1 & 2.6800e-2 (1.58e-2) - & 4.6123e-2 (1.82e-2) - & 6.2559e-2 (7.72e-3) - & 7.2358e-1 (1.01e-1) - & 3.8123e-2 (2.14e-2) - & 3.8144e-2 (2.08e-2) - & \textbf{6.9980e-3 (1.76e-3)} \\
		ZDT2 & 5.9006e-2 (7.55e-2) - & 1.1258e-1 (4.76e-2) - & 6.2866e-2 (1.08e-2) - & 1.4689e+0 (1.80e-1) - & 1.2262e-1 (5.89e-2) - & 1.1662e-1 (4.88e-2) - & \textbf{2.4188e-2 (2.15e-2)} \\
		ZDT3 & 3.1487e-2 (2.29e-2) - & 3.1938e-1 (5.43e-2) - & 7.6586e-2 (1.45e-2) - & 7.8261e-1 (9.28e-2) - & 1.9540e-1 (4.85e-2) - & 2.0511e-1 (4.72e-2) - & \textbf{1.7122e-2 (5.58e-3)} \\
		ZDT4 & 1.7288e-1 (1.12e-1) + & 1.1806e+1 (6.89e+0) - & \textbf{1.0835e-2 (1.47e-3) +} & 2.8495e+1 (4.86e+0) - & 7.8472e+0 (4.44e+0) - & 3.9875e+0 (3.34e+0) - & 2.5453e-1 (9.70e-2) \\
		ZDT6 & 1.1214e-2 (2.47e-3) - & 5.3427e-2 (1.06e-1) - & 1.5028e+0 (9.01e-2) - & 9.8493e-2 (1.98e-1) - & 4.2009e-3 (3.96e-3) = & \textbf{3.5353e-3 (9.44e-4) =} & 3.8825e-3 (2.74e-3) \\
		\midrule[0.75pt]
		DTLZ1 & 2.0982e-2 (3.27e-4) - & 1.2916e+0 (1.53e+0) - & 3.4132e+0 (7.67e-1) - & 2.0141e+0 (1.26e+0) - & 3.4192e-1 (6.12e-1) - & 6.3591e-1 (9.08e-1) - & \textbf{1.8507e-2 (8.58e-4)} \\
		DTLZ2 & 5.4467e-2 (1.28e-6) - & 7.5716e-2 (6.94e-4) - & 9.4179e-2 (5.15e-3) - & 8.0869e-2 (3.18e-3) - & 7.5370e-2 (6.65e-4) - & 7.5319e-2 (5.90e-4) - & \textbf{4.7007e-2 (1.53e-4)} \\
		DTLZ3 & \textbf{9.7856e-1 (1.26e+0) =} & 2.5743e+1 (2.40e+1) - & 6.0011e+1 (1.01e+1) - & 1.4375e+1 (1.81e+1) - & 1.8153e+1 (1.89e+1) - & 1.7439e+1 (2.28e+1) - & 1.2903e+0 (3.04e+0) \\
		DTLZ4 & 3.8444e-1 (3.21e-1) - & 2.2549e-1 (1.76e-1) - & 7.9127e-2 (3.12e-3) - & 1.1164e-1 (6.26e-3) - & 1.2233e-1 (7.09e-2) - & 1.6551e-1 (9.04e-2) - & \textbf{4.7086e-2 (2.36e-4)} \\
		DTLZ5 & 3.3764e-2 (6.54e-5) - & \textbf{1.4166e-2 (1.45e-4) +} & 2.4473e-2 (4.64e-3) + & 2.2885e-2 (2.25e-3) + & 1.4252e-2 (1.58e-4) + & 1.4401e-2 (1.19e-4) + & 2.6538e-2 (5.78e-4) \\
		DTLZ6 & 3.3881e-2 (2.56e-5) - & 1.4338e-2 (6.68e-5) + & 4.7806e+0 (1.13e-1) - & \textbf{8.4479e-3 (1.45e-3) +} & 1.4528e-2 (5.14e-5) + & 1.4537e-2 (3.79e-5) + & 2.9008e-2 (4.72e-5) \\
		DTLZ7 & 1.9743e-1 (1.64e-1) - & 2.2761e-1 (6.01e-2) - & 3.4424e-1 (4.58e-2) - & 4.7891e-1 (1.78e-1) - & 1.9837e-1 (3.99e-2) - & 2.1575e-1 (7.33e-2) - & \textbf{1.1006e-1 (3.56e-4)} \\
		\midrule[0.75pt]
		WFG1 & \textbf{3.4553e-1 (3.64e-2) +} & 1.3536e+0 (8.07e-2) - & 1.3058e+0 (6.48e-2) - & 1.6100e+0 (5.58e-2) - & 1.5443e+0 (7.38e-2) - & 1.4408e+0 (1.09e-1) - & 7.1675e-1 (8.33e-2) \\
		WFG2 & 2.6195e-1 (1.71e-2) - & 3.4125e-1 (2.29e-2) - & 2.6466e-1 (1.52e-2) - & 2.7646e-1 (1.44e-2) - & 3.6394e-1 (3.00e-2) - & 3.5163e-1 (2.96e-2) - & \textbf{2.0742e-1 (1.96e-2)} \\
		WFG3 & 2.0213e-1 (5.61e-2) - & 1.8917e-1 (3.15e-2) - & 2.1991e-1 (1.76e-2) - & 2.6856e-1 (3.86e-2) - & 2.1493e-1 (3.59e-2) - & 1.9109e-1 (4.03e-2) - & \textbf{1.2825e-1 (4.77e-3)} \\
		WFG4 & 2.6556e-1 (7.53e-3) - & 3.8005e-1 (1.12e-2) - & 3.3302e-1 (7.61e-3) - & 3.7257e-1 (2.00e-2) - & 3.9968e-1 (1.25e-2) - & 3.9868e-1 (1.57e-2) - & \textbf{2.3703e-1 (5.38e-3)} \\
		WFG5 & 2.5422e-1 (3.39e-3) - & 3.3639e-1 (4.02e-3) - & 3.4015e-1 (1.63e-2) - & 2.7850e-1 (8.20e-3) - & 3.3880e-1 (4.34e-3) - & 3.3789e-1 (5.03e-3) - & \textbf{2.2353e-1 (3.01e-3)} \\
		WFG6 & \textbf{2.9751e-1  (1.92e-2) +} & 4.3393e-1 (2.83e-2) = & 3.5758e-1 (1.53e-2) + & 3.5166e-1(3.43e-2) + & 4.3816e-1 (3.03e-2) - & 4.3740e-1 (4.00e-2) = & 4.2542e-1 (4.02e-2) \\
		WFG7 & 3.5619e-1 (5.14e-2) + & 3.6897e-1 (1.03e-2) = & 3.5277e-1 (1.18e-2) + & \textbf{3.2754e-1 (1.56e-2) +} & 3.6693e-1 (9.62e-3) = & 3.6391e-1 (7.96e-3) = & 3.6882e-1 (1.13e-2) \\
		WFG8 & \textbf{3.3073e-1 (1.16e-2) +} & 4.7517e-1 (4.77e-2) = & 4.2368e-1 (1.53e-2) + & 4.0988e-1 (1.51e-2) + & 4.4389e-1 (2.41e-2) + & 4.5188e-1 (4.59e-2) + & 4.8352e-1 (5.64e-2) \\
		WFG9 & \textbf{3.2760e-1 (6.22e-2) +} & 3.4008e-1  (1.98e-2) = & 3.3717e-1 (1.39e-2) = & 3.2822e-1  (5.93e-2) = & 3.8344e-1 (4.80e-2) = & 3.4880e-1 (3.77e-2) + & 3.5458e-1 (3.56e-2) \\
		\midrule[0.75pt]
		UF1 & 1.4922e-1 (5.41e-2) - & 1.9429e-3 (7.06e-4) - & 5.4924e-2 (1.06e-2) - & 3.1349e-2 (3.10e-3) - & 2.3837e-3 (1.94e-3) = & 2.7501e-2 (7.77e-2) = & \textbf{1.3666e-3 (1.05e-4)} \\
		UF2 & 4.5377e-2 (3.63e-2) - & 3.8904e-3 (1.98e-3) - & 2.0652e-2 (4.63e-3) - & 3.0338e-2 (2.61e-3) - & 3.0706e-3 (9.33e-4) - & 2.0981e-3 (7.41e-4) = & \textbf{2.0228e-3 (4.02e-4)} \\
		UF3 & 3.0328e-1 (3.32e-2) - & 2.7810e-2 (4.57e-2) = & 7.0425e-2 (7.56e-3) - & 1.2291e-1 (3.10e-2) - & \textbf{1.8076e-2 (1.59e-2) -} & 2.5324e-2 (2.19e-2) = & 2.1607e-2(1.97e-2) \\
		UF4 & 5.6160e-2 (5.72e-3) - & 6.1483e-2 (4.47e-3) - & 5.9624e-2 (2.70e-3) - & 5.1042e-2 (2.55e-3) - & 5.1265e-2 (2.86e-3) - & 4.9417e-2 (2.48e-3) - & \textbf{3.3923e-2 (1.36e-3)} \\
		UF5 & \textbf{4.8038e-1 (1.27e-1) =} & 5.3079e-1 (1.07e-1) = & 5.7316e-1 (1.02e-1) = & 1.0703e+0 (9.55e-2) - & 5.2543e-1 (1.38e-1) = & 6.3341e-1 (1.26e-1) - & 5.2563e-1 (1.62e-1) \\
		UF6 & 3.9102e-1 (1.26e-1) = & 3.9160e-1 (7.58e-2) = & \textbf{1.4211e-1 (2.47e-2) +} & 3.7554e-1 (2.18e-2) = & 4.4067e-1 (6.41e-2) - & 4.5274e-1 (4.32e-2) - & 3.7661e-1 (7.47e-2) \\
		UF7 & 4.7546e-1 (1.77e-1) - & 2.3092e-1 (2.48e-1) - & 3.3065e-2 (4.82e-2) - & \textbf{1.3341e-2 (9.09e-4) +} & 2.2646e-1 (2.61e-1) - & 3.1521e-1 (2.59e-1) - & 3.1031e-2 (1.06e-1) \\
		UF8 & 2.8741e-1 (2.30e-1) - & 1.1058e-1 (5.11e-2) = & 1.5769e-1 (5.96e-2) - & 1.2721e-1 (9.70e-3) - & 9.6619e-2 (2.88e-2) = & 1.0052e-1 (1.29e-2) - & \textbf{9.3945e-2 (7.22e-2)} \\
		UF9 & 2.6355e-1 (3.35e-2) - & 1.7194e-1 (3.57e-2) - & 1.5937e-1 (6.44e-2) - & 1.1807e-1 (1.11e-2) - & 1.6370e-1 (5.13e-2) - & 1.8932e-1 (6.28e-3) - & \textbf{7.7286e-2 (6.56e-2)} \\
		UF10 & 7.3227e-1 (1.69e-1) - & 4.7879e-1 (7.97e-2) + & \textbf{2.3738e-1(5.84e-4) +} & 2.5258e+0 (1.61e-1) - & 8.5939e-1 (1.29e-1) - & 6.6767e-1 (8.06e-2) - & 5.3962e-1 (1.80e-1) \\
		\midrule[0.75pt]
		\multirow{2}{*}{NN1 (Statlog\_Australian)} & \multirow{2}{*}{6.2055e-1 (1.20e-2) -} & \multirow{2}{*}{7.6301e-1 (1.86e-2) -} & \multirow{2}{*}{6.9280e-1 (2.36e-2) -} & \multirow{2}{*}{7.9543e-1 (1.37e-2) =} & \multirow{2}{*}{7.2571e-1 (2.56e-2) -} & \multirow{2}{*}{7.3659e-1 (2.52e-2) -} & \multirow{2}{*}{\textbf{8.1633e-1 (1.16e-2)}} \\
		&  &  &  &  &  &  &  \\
		\multirow{2}{*}{NN2 (Climate)} & \multirow{2}{*}{6.3505e-1   (1.59e-2) -} & \multirow{2}{*}{8.0321e-1 (2.71e-2) -} & \multirow{2}{*}{7.0199e-1 (2.56e-2) -} & \multirow{2}{*}{8.2303e-1 (1.61e-2) -} & \multirow{2}{*}{7.6123e-1 (2.47e-2) -} & \multirow{2}{*}{7.7099e-1 (2.83e-2) -} & \multirow{2}{*}{\textbf{8.4754e-1 (7.12e-3)}} \\
		&  &  &  &  &  &  &  \\
		\multirow{2}{*}{NN3 (Bench\_Sonar)} & \multirow{2}{*}{5.0728e-1   (1.79e-2) -} & \multirow{2}{*}{6.7753e-1   (2.30e-2) -} & \multirow{2}{*}{5.4119e-1   (2.67e-2) -} & \multirow{2}{*}{\textbf{7.6268e-1   (1.50e-2) +}} & \multirow{2}{*}{6.7969e-1   (2.37e-2) -} & \multirow{2}{*}{6.7220e-1   (2.44e-2) -} & \multirow{2}{*}{7.2155e-1   (1.52e-2)} \\
		&  &  &  &  &  &  &  \\
		\midrule[0.75pt]
		+/-/= & 6/25/3 & 3/23/8 & 7/25/2 & 7/24/3 & 3/24/7 & 4/24/6 & —\\
		\bottomrule[1.5pt]
	\end{tabular}
		\begin{tablenotes}
		\item IGD: Inverted Generational Distance
		\item HV: Hypervolume 
	\end{tablenotes}
	\end{threeparttable}
\end{sidewaystable*}

\subsection{Combinatorial Optimization Problems}
Combinatorial optimization problems (COP) are a class of optimization problems with discrete variables. The optimization objective of the problem is to find the optimal solution from a set of feasible solutions. However, as many COPs are known to be NP-hard, it means that finding the optimal solution is either computationally intensive or almost impossible. In light of this challenge and the need for algorithmic efficiency in practical applications, employing EA has become a prevailing approach to searching for approximate optimal solutions. Various RL-EAs have been proposed and applied in domains such as intelligent manufacturing and transportation to address COPs effectively. Here, we primarily focus on providing an elaborate introduction to RL-EAs specifically designed for solving production scheduling problems and vehicle routing problems. Subsequently, research on other COPs will be briefly introduced.

\subsubsection{Production Scheduling Problem}

\begin{table*}[htbp]
	\centering
	\caption{Application in production scheduling problems}
	\label{Production scheduling problem}
	\begin{tabularx}{\textwidth}{m{1.5cm}m{3cm}m{5cm}m{3cm}m{3cm}}
		\toprule[1.5pt]
		Year & Authors & Problem & Type of EA & Type of RL \\
		\midrule[0.75pt]
		2022 & Cheng et al. \cite{cheng2022multi} & Energy-aware mixed shop scheduling problem & Hyper-heuristic with Bi-criteria selection & Q-learning \\
		2022 & Cheng et al. \cite{cheng2022scheduling} & Flexible manufacturing cell with no-idle flow-lines and job-shop scheduling problem & GA & Q-learning \\
		2022 & Li et al. \cite{li2022learning} & Multi-objective energy-efficient flexible job shop scheduling problem with type-2 processing time & MA & Q-learning \\
		2022 & Li et al. \cite{li2022reinforcement} & Bi-objective fuzzy flexible job shop scheduling problem & MOEA/D & Q-learning \\
		2022 & Wang et al. \cite{wang2022adaptive} & Distributed three-stage assembly scheduling with maintenance & ABC & Q-learning \\
		2022 & Li et al. \cite{li2022improved} & Permutation flow-shop scheduling problem & ABC & Q-learning \\
		2022 & Zhao et al.  \cite{zhao2022reinforcementb} & Energy-efficient distributed no-wait flow-shop scheduling problem with sequence-dependent setup time & Collaborative meta-heuristic & Double Q-learning \\
		2022 & Zhao et al.  \cite{zhao2022reinforcementa} & Distributed heterogeneous no-wait flow-shop scheduling problem with sequence-dependent setup times & ABC & Q-learning \\
		2022 & Zhao et al.  \cite{zhao2022hyperheuristic} & Multiobjective energy-efficient distributed blocking flow shop scheduling problem & Hyper-heuristic & Q-learning \\
		2022 & Du et al. \cite{du2022knowledge} & Flexible job shop scheduling problem & EDA & DQN \\
		2023 & Jia et al. \cite{jia2023q} & Distributed three-stage assembly hybrid flow shop scheduling with flexible preventive maintenance & MA & Q-learning \\
		2023 & Zhao et al.  \cite{zhao2023knowledge} & Distributed blocking flow shop scheduling problem & Cooperative scatter search & Q-learning \\
		2023 & Li et al. \cite{li2023muti} & Muti-objective energy-efficient hybrid flow shop scheduling problem & NSGA-II & Q-learning \\
		2023 & Li et al. \cite{li2023reinforcementb} & Flexible job-shop scheduling problem with lot streaming & ABC & Q-learning \\
		2023 & Zhang et al. \cite{zhang2023reinforcement} & Multimanned assembly line balancing under uncertain demand & NSGA-II & Q-learning\\
            2023 & Wu et al. \cite{wu2023aq} & Distributed flexible job-shop scheduling problem & Hyper-heuristic & Q-learning \\
            2023 & Zhang et al. \cite{zhang2023q} & Distributed flexible job-shop scheduling problem with crane transportation & Hyper-heuristic & Q-learning \\
            2023 & Zhu et al. \cite{zhu2023hyper} & Flow-shop scheduling problem with fuzzy processing times & Hyper-heuristic & Q-learning\\
            2023 & Gao et al. \cite{gao2023improved}& Flexible job-shop scheduling problem & PSO & Q-learning \\
            2023 & Zhao et al. \cite{zhao2023reinforcement}& No-wait flow shop scheduling problem  & Brain storm optimisation algorithm & Q-learning \\
	2023 & Zhao et al.  \cite{zhao2023cooperative}& Distributed permutation flowshop scheduling problem with sequence-dependent setup times & Cooperative scatter search & Q-learning \\
	2023 & Yu et al.  \cite{yu2023improved}& Distributed assembly permutation flowshop scheduling problem & ABC, PSO, GA, Jaya& Q-learning\\
		\bottomrule[1.5pt]
	\end{tabularx}
\end{table*}

\begin{figure}[htp]
\centering
\includegraphics[width=0.4\textwidth]{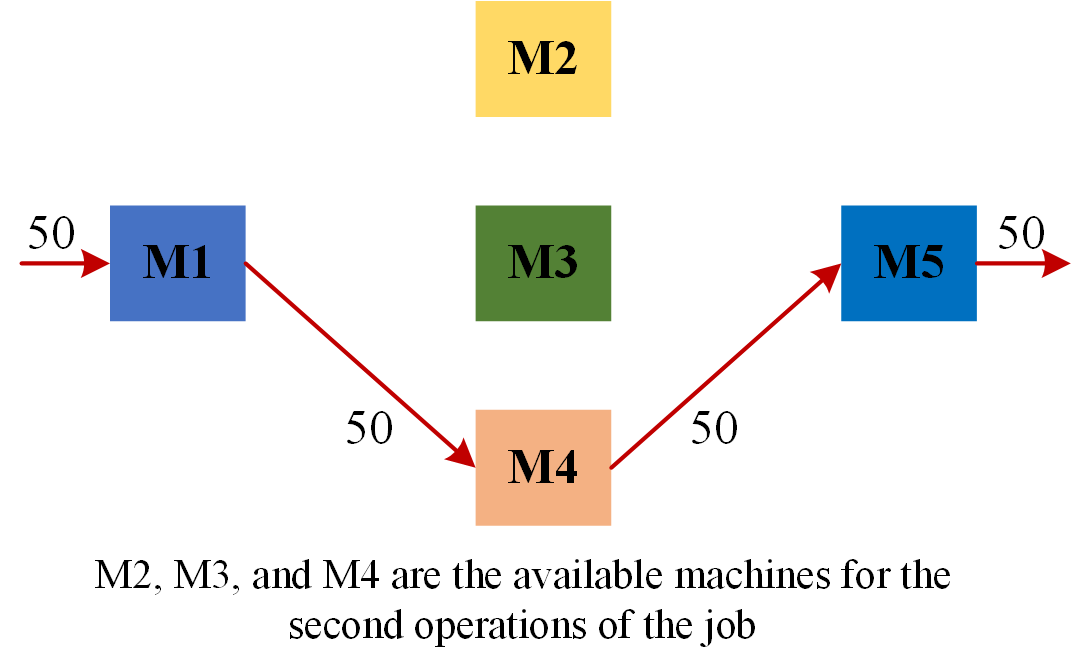}
\caption{A sample job route in the FJSP}
\label{A sample job route in the FJSP}
\end{figure}

The production scheduling problem (PSP) is a classical class of combinatorial optimization problems. In this problem, there are a series of machines and products to be processed. The objective of problem optimization is to complete all the product processing tasks as early as possible. PSPs have been categorized as parallel machine scheduling problems, flow production scheduling problems, and flexible production scheduling problems, etc. based on machine capabilities and product processing. Based on this, additional factors are considered in the study of this problem. In general, certain constraints arise from the limited number of product processing steps and equipment capacity requirements that contribute to the complexity of these problems. Consequently, the time-dependent nature of the problem results in an exceptionally large solution space. Fig.\ref{A sample job route in the FJSP} gives a sample job route in the flexible job-shop scheduling problem (FJSP). It is worth noting that this particular product necessitates three distinct processes to be fully completed. In order to obtain high-quality production solutions for various intricate PSPs, numerous RL-EAs have been proposed.

As shown in Table \ref{Production scheduling problem}, there has been a notable increase in research focus on the use of RL-EA for solving PSP during the recent two years. This indicates a growing interest among researchers in utilizing this novel type of EA for production scheduling purposes. Given the inherent variations in actual factory production environments, achieving complete consistency in research questions is challenging. In most studies, RL primarily handles the adaptive selection of LS operators due to the high complexity of PSP and the need for solution algorithms with strong exploitation capabilities. Therefore, LS plays a pivotal role in enhancing the quality of PSP solutions. Furthermore, the diverse range of EAs used in the study of RL-EA for solving PSP makes it difficult to ensure that the used algorithms can be adapted to other problem characteristics. Therefore, there is an urgent need to analyze which type of EA is best suited for combination with RL.

The increasing research on the application of RL-EA in PSP is highly related to the remarkable performance of the algorithm. Table \ref{Comparison of the average HV for algorithms} demonstrates the exceptional performance in terms of average HV values of a cooperative meta-heuristic algorithm based on Q-learning (CMAQ) \cite{zhao2022reinforcementb} in two cases of energy-efficient distributed no-wait flow-shop scheduling with sequence-dependent setup time. The performance of CMAQ stands out significantly across different sizes of ssd100 and ssd125 cases, consistently achieving the highest HV values except for size (100, 5). 

\begin{table*}[htbp]
	\footnotesize
	\centering
	\caption{Comparison of the average HV for algorithms in \textit{ssd}100 and \textit{ssd}125 \cite{zhao2022reinforcementb} }
	\label{Comparison of the average HV for algorithms}
	\begin{threeparttable}
	\begin{tabular}{cccccc|ccccc}
		\toprule[1.5pt]
		\multirow{2}{*}{\textit{\textbf{(n,m)}}} & \multicolumn{5}{c}{\textit{\textbf{ssd100}}} & \multicolumn{5}{c}{\textit{\textbf{ssd125}}} \\
		\cmidrule(l){2-6} \cmidrule(l){7-11}
		& \textit{\textbf{CMAQ}} & \textit{\textbf{KCA}} & \textit{\textbf{MOWSA}} & \textit{\textbf{IJaya}} & \textit{\textbf{MMOEA/D}} & \textit{\textbf{CMAQ}} & \textit{\textbf{KCA}} & \textit{\textbf{MOWSA}} & \textit{\textbf{IJaya}} & \textit{\textbf{MMOEA/D}} \\
		\midrule[0.75pt]
		(20,5) & \textbf{8.40E-01} & 6.44E-01 & 7.03E-01 & 5.27E-01 & 7.54E-01 & \textbf{8.31E-01} & 6.61E-01 & 6.93E-01 & 5.50E-01 & 7.33E-01 \\
		(20,10) & \textbf{8.67E-01} & 6.46E-01 & 6.81E-01 & 4.75E-01 & 7.41E-01 & \textbf{8.49E-01} & 6.50E-01 & 6.68E-01 & 5.03E-01 & 7.51E-01 \\
		(20,20) & \textbf{8.90E-01} & 6.66E-01 & 6.40E-01 & 5.07E-01 & 7.01E-01 & \textbf{9.07E-01} & 6.96E-01 & 6.20E-01 & 4.70E-01 & 6.63E-01 \\
		(50,5) & \textbf{8.37E-01} & 6.71E-01 & 7.32E-01 & 4.55E-01 & 7.51E-01 & \textbf{8.51E-01} & 6.30E-01 & 7.09E-01 & 4.67E-01 & 7.15E-01 \\
		(50,10) & \textbf{8.93E-01} & 6.13E-01 & 6.82E-01 & 4.41E-01 & 7.24E-01 & \textbf{8.73E-01} & 5.98E-01 & 6.84E-01 & 4.19E-01 & 6.82E-01 \\
		(50,20) & \textbf{8.81E-01} & 5.42E-01 & 6.35E-01 & 3.72E-01 & 6.37E-01 & \textbf{8.90E-01} & 5.30E-01 & 6.27E-01 & 3.55E-01 & 6.32E-01 \\
		(100,5) & 8.19E-01 & \textbf{8.27E-01} & 7.12E-01 & 4.42E-01 & 7.19E-01 & 8.02E-01 & \textbf{8.03E-01} & 7.09E-01 & 4.35E-01 & 6.95E-01 \\
		(100,10) & \textbf{8.62E-01} & 5.94E-01 & 7.11E-0 & 4.15E-01 & 7.01E-01 & \textbf{8,83E-01} & 6.02E-01 & 7.01E-01 & 4.29E-0 & 6.96E-01 \\
		(100,20) & \textbf{8.74E-01} & 5.24E-01 & 6.27E-01 & 3.16E-01 & 6.44E-01 & \textbf{8.71E-01} & 5.28E-01 & 6.34E-01 & 3.14E-01 & 6.63E-01 \\
		(200,10) & \textbf{8.31E-01} & 7.28E-01 & 5.97E-01 & 3.41E-01 & 6.70E-01 & \textbf{8.20E-01} & 7.45E-01 & 6.15E-01 & 3.13E-01 & 6.38E-01 \\
		(200,20) & \textbf{8.26E-01} & 6.63E-01 & 5.48E-01 & 2.70E-01 & 6.00E-01 & \textbf{8.13E-01} & 6.82E-01 & 5.20E-01 & 2.73E-01 & 5.98E-01 \\
		(500,20) & \textbf{7.54E-01} & 5.70E-01 & 3.00E-01 & 6.79E-01 & 7.41E-01 & \textbf{7.14E-01} & 6.48E-01 & 2.59E-01 & 6.35E-01 & 6.71E-01 \\
		Average & \textbf{8.48E-01} & 6.41E-01 & 6.31E-01 & 4.37E-01 & 6.99E-01 & \textbf{8.42E-01} & 6.48E-01 & 6.20E-01 & 4.30E-01 & 6.78E-01\\
		\bottomrule[1.5pt]
	\end{tabular}
			\begin{tablenotes}
		\item \textit{n}: The number of jobs to be processed. \textit{m}: The number of machines in each factory.
	\end{tablenotes}
	\end{threeparttable}
\end{table*}

\subsubsection{Vehicle Routing Problem}

\begin{table*}[htbp]
	\centering
	\caption{Application in vehicle routing problems}
	\label{VRP}
	\begin{tabularx}{\textwidth}{m{1.5cm}m{3cm}m{5cm}m{3cm}m{3cm}}
		\toprule[1.5pt]
		Year & Authors & Problem & Type of EA & Type of RL \\
		\midrule[0.75pt]
	2021 & Qin et al. \cite{qin2021novel} & Heterogeneous vehicle routing problem & Hyperheuristic & A3C \\
	2022 & Zhang et al. \cite{zhang2022deep} & Container terminal truck routing problem with uncertain service times & Hyperheuristic & Double deep Q network \\
	2022 & Qi et al. \cite{qi2022qmoea} & Time-dependent green vehicle routing problems with time windows & NSGA-II & Q-learning \\
	2022 & Rodr{\'\i}guez et al. \cite{rodriguez2022new}  & CEVRP & Hyperheuristic & Multi-armed bandit RL \\
	2022 & Zhang et al. \cite{zhang2022multi} & Path planning of unmanned air vehicles & PSO & Q-learning \\
	2023 & Zhou et al. \cite{zhou2023adaptive} & Milk-run vehicle scheduling problem & ABC & DQN \\
	2023 & Shang et al. \cite{shang2023green} & Green location routing problem & Hyperheuristic & Q-learning \\
	2023 & Yan et al. \cite{yan2023novel} & Autonomous underwater vehicle path planning & Tuna swarm optimization & Q-learning\\
	\bottomrule[1.5pt]
	\end{tabularx}
\end{table*}

\begin{figure}[htp]
\centering
\includegraphics[width=0.45\textwidth]{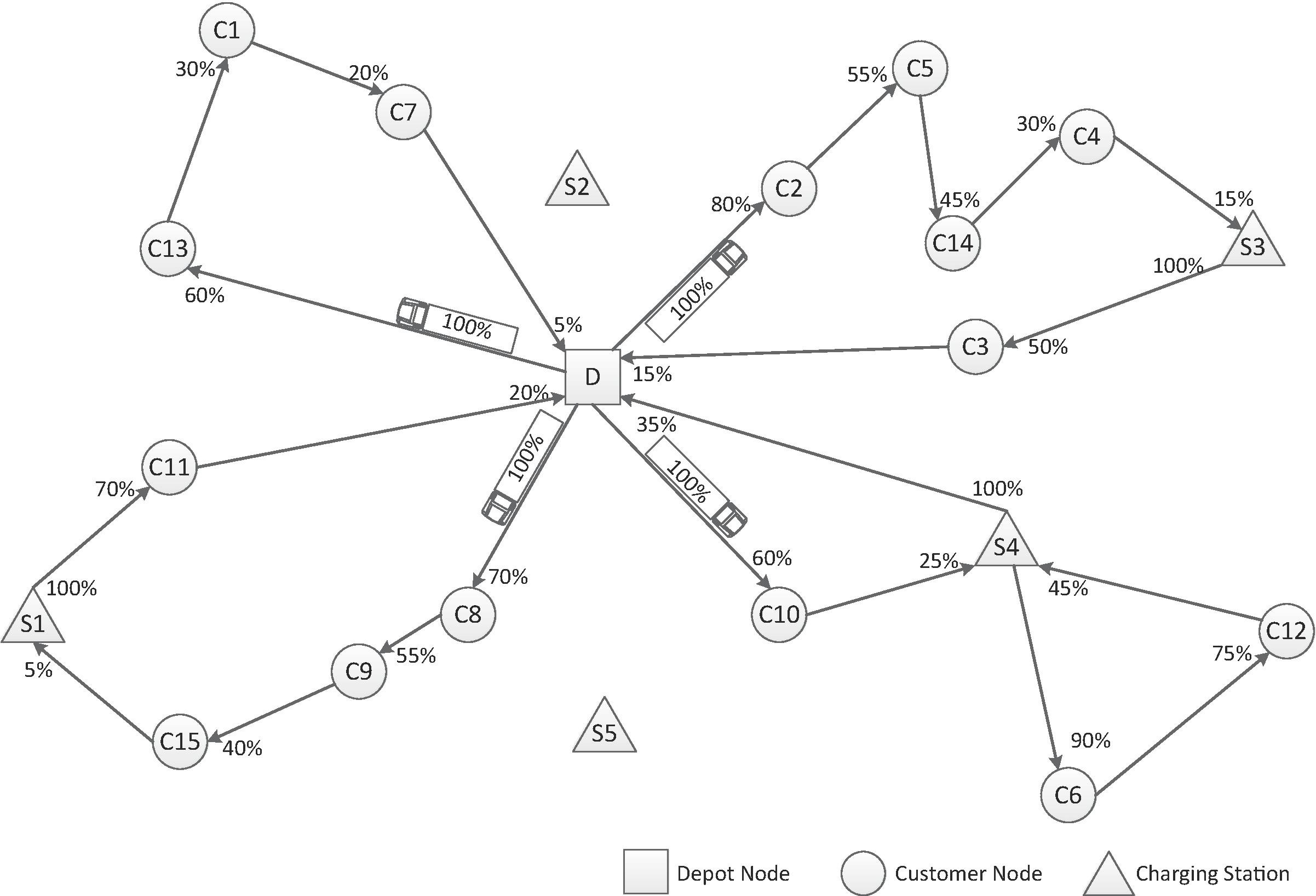}
\caption{An illustrative example for EVRP \cite{kucukoglu2021electric}}
\label{An illustrative example for EVRP}
\end{figure}

\begin{figure}[htp]
\centering
\includegraphics[width=0.45\textwidth]{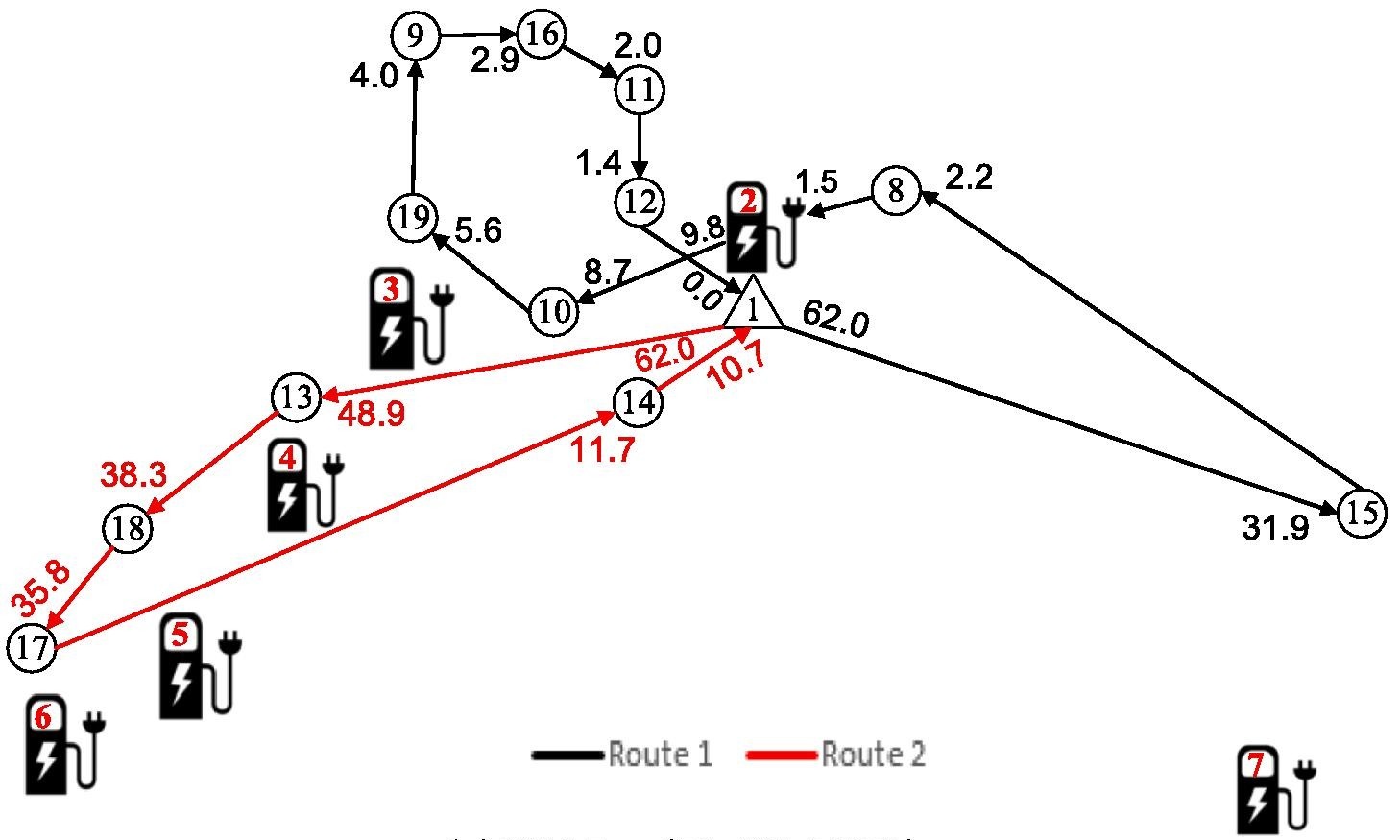}
\caption{An illustrative example for EVRP with time window \cite{rastani2019effects}}
\label{An illustrative example for EVRP with time window}
\end{figure}

The vehicle routing problem (VRP) involves a set of customers with diverse item requirements and a warehouse where the items are stored. The warehouse is responsible for dispatching vehicles to efficiently transport the items to the customers. The objective of optimization is to ensure that the vehicles deliver the items to the appropriate customers within a minimal transportation distance. VRP, similar to PSP, belongs to an extensively researched category of combinatorial optimization problems. Various types of extended problems modify the optimization objective or add constraints to the traditional VRP for specific problem scenarios. An illustrative example demonstrating electric vehicle routing problems (EVRP) and EVRP with time window (EVRPTW) can be seen in Fig.\ref{An illustrative example for EVRP} and Fig.\ref{An illustrative example for EVRP with time window}, respectively. An endless number of VRP variants motivate algorithmic approaches, and some studies employ RL-EA to solve specific problem scenarios.

The application of RL-EA in solving various types of VRPs with complex constraints and large problem sizes is demonstrated in Table \ref{other combinatorial optimization problem}. This prompts researchers to incorporate heuristic rules into the algorithm design process for obtaining high-quality solutions. Consequently, RL can be dynamically selected during the iterative search based on the performance of these heuristic rules. Accordingly, RL can make an adaptive selection during the iterative search based on the performance of these heuristic rules. Moreover, well-established RL methods for solving VRP can be directly integrated into the EA framework to expedite solution generation \cite{zhang2022multi}. The advantage of adopting such a combination of algorithmic ideas lies in its ability to facilitate rapid algorithm design while effectively ensuring solution effectiveness.

Table \ref{Results of the HHASA applied to large instances of the benchmark} gives the performance of the hyper-heuristic adaptive simulated annealing with reinforcement learning algorithm (HHASA) \cite{rodriguez2022new} and its variants tested on the benchmark of the capacitated electric vehicle routing problem (CEVRP). Among these ten large-scale examples, HHASA and its variants can find better planning results in seven instances. Even the most basic HHASA can achieve better results than the comparison algorithms in six instances.

\begin{table*}[htbp]
	\centering
	\scriptsize
        \caption{Results of the HHASA applied to large instances of the benchmark \cite{rodriguez2022new} }
	\label{Results of the HHASA applied to large instances of the benchmark}
	\begin{threeparttable}
	\begin{tabular}{llllll|lllll}
		\toprule[1.5pt]
		\textbf{Instance} & \textbf{Indicator} & \textbf{HHASA$_{TS}$} & \textbf{HHASA$ _{UCB1} $} & \textbf{HHASA$ _{\varepsilon-G} $} & \textbf{HHASA} & \textbf{BACO}  &       \textbf{VNS}    &     \textbf{SA}   &      \textbf{GA}   &    \textbf{GRASP} \\
		\midrule[0.75pt]
		\multirow{3}{*}{X143} & min & 15910.86 & 15912.77 & 15899.86 & 15921.68 & 15901.23 & 16028.05 & 16610.37 & 16488.60 & 16460.80 \\
		& mean & 16214.37 & 16231.33 & 16173.06 & 16271.78 & \textbf{16031.46*} & 16459.31 & 17188.90 & 16911.50 & 16823.00 \\
		& std & 215.77 & 173.73 & 198.91 & 250.09 & 262.47 & 242.59 & 170.44 & 282.30 & 157.00 \\
		\multirow{3}{*}{X214} & min & 11090.28 & 11097.63 & 11098.34 & 11120.28 & 11133.14 & 11323.56 & 11404.44 & 11762.07 & 11575.60 \\
		& mean & \textbf{11206.60*} & 11260.83 & 11247.30 & 11251.80 & 11219.70 & 11482.20 & 11680.35 & 12007.06 & 11740.70 \\
		& std & 84.58 & 88.73 & 99.53 & 73.03 & 46.25 & 76.14 & 116.47 & 156.69 & 80.41 \\
		\multirow{3}{*}{X352} & min & 26622.42 & 26549.88 & 26486.05 & 26606.06 & 26478.34 & 27064.88 & 27222.96 & 28008.09 & 27521.20 \\
		& mean & 26750.60 & 26760.35 & 26760.58 & 26812.89 & \textbf{26593.18*} & 27217.77 & 27498.03 & 28336.07 & 27775.30 \\
		& std & 102.55 & 116.44 & 135.77 & 90.44 & 72.86 & 86.20 & 155.62 & 205.29 & 111.99 \\
		\multirow{3}{*}{X459} & min & 24794.35 & 24769.67 & 24752.03 & 24815.37 & 24763.93 & 25370.80 & 27222.96 & 26048.21 & 25929.20 \\
		& mean & 25041.10 & 25036.67 & 24979.89 & 25060.02 & \textbf{24916.60*} & 25582.27 & 25809.47 & 26345.12 & 26263.30 \\
		& std & 237.58 & 114.68 & 151.54 & 121.58 & 94.08 & 106.89 & 157.97 & 185.14 & 134.66 \\
		\multirow{3}{*}{X573} & min & 51436.90 & 51436.00 & 51485.68 & 51545.10 & 53822.87 & 52181.51 & 51929.24 & 54189.62 & 52584.50 \\
		& mean & \textbf{51776.70} & \textbf{51764.24} & \textbf{51771.50} & \textbf{51748.42*} & 54567.15 & 52548.09 & 52793.66 & 55327.62 & 52990.90 \\
		& std & 166.86 & 152.69 & 158.01 & 119.57 & 231.05 & 278.85 & 577.24 & 548.05 & 246.79 \\
		\multirow{3}{*}{X685} & min & 69955.95 & 70348.53 & 70323.62 & 70413.81 & 70834.88 & 71345.40 & 72549.90 & 73925.56 & 72481.60 \\
		& mean & \textbf{70401.25*} & \textbf{70719.10} & \textbf{70684.34} & \textbf{70791.10} & 71440.57 & 71770.57 & 73124.98 & 74508.03 & 72792.70 \\
		& std & 218.98 & 291.53 & 174.26 & 222.85 & 281.78 & 197.08 & 320.07 & 409.43 & 189.53 \\
		\multirow{3}{*}{X749} & min & 79779.87 & 79829.23 & 79850.73 & 79732.99 & 80299.76 & 81002.01 & 81392.78 & 84034.73 & 82187.30 \\
		& mean & \textbf{80135.67*} & \textbf{80256.36} & \textbf{80318.42} & \textbf{80397.82} & 80694.54 & 81327.39 & 81848.13 & 84759.79 & 82733.40 \\
		& std & 219.50 & 303.30 & 399.47 & 432.11 & 223.91 & 176.19 & 275.26 & 376.10 & 213.21 \\
		\multirow{3}{*}{X819} & min & 161924.79 & 162350.42 & 162387.34 & 162523,88 & 164720.80 & 164289.95 & 165069.77 & 170965.68 & 166500.00 \\
		& mean & \textbf{162530.67*} & \textbf{162819.78} & \textbf{162883.17} & \textbf{163031.19} & 165565.79 & 164926.41 & 165895.78 & 172410.12 & 166970.00 \\
		& std & 289.41 & 258.35 & 300.11 & 389.12 & 401.02 & 318.62 & 403.70 & 568.58 & 211.84 \\
		\multirow{3}{*}{X916} & min & 336717.71 & 337200.96 & 337520.94 & 338007.56 & 342993.01 & 341649.91 & 342796.88 & 357391.57 & 345777.00 \\
		& mean & \textbf{337641.92*} & \textbf{338349.57} & \textbf{338639.53} & \textbf{338688.50} & 344999.95 & 342460.70 & 343533.85 & 360269.94 & 347269.00 \\
		& std & 461.47 & 454.28 & 544.69 & 328.64 & 905.72 & 510.66 & 556.98 & 229.19 & 654.93 \\
		\multirow{3}{*}{X1001} & min & 75469.29 & 75864.07 & 75782.95 & 75850.15 & 76297.09 & 77476.36 & 78053.86 & 78832.90 & 77636.20 \\
		& mean & \textbf{75931.28*} & \textbf{76131.56} & \textbf{76245.73} & \textbf{76234.51} & 77434.33 & 77920.52 & NA & 79163.34 & 78111.20 \\
		& std & 304.10 & 212.24 & 226.30 & 271.18 & 719.86 & 234.73 & 306.27 & NA & 315.31\\
		\bottomrule[1.5pt]
	\end{tabular}
	\begin{tablenotes}
	\item $ TS, UCB_1, \varepsilon-G $: Simple methods for solving the multi-armed bandit problem
	\item GRASP: Greedy Randomized Adaptive Search Procedure 
	\item BACO: Bilevel Ant Colony Optimization
\end{tablenotes}
	\end{threeparttable}
\end{table*}

\subsubsection{Other Combinatorial Optimization Problems}

\begin{table*}[htbp]
	\centering
	\caption{Application in other combinatorial optimization problems}
	\label{other combinatorial optimization problem}
	\begin{tabularx}{\textwidth}{m{1.5cm}m{3cm}m{5cm}m{3cm}m{3cm}}
		\toprule[1.5pt]
		Year & Authors & Problem & Type of EA & Type of RL \\
		\midrule[0.75pt]
		2012 & Buzdalova and Buzdalov \cite{buzdalova2012increasing} & Royal roads model problem, hierarchical-if-and-only-if function & (1 + 1) evolutionary strategy & Q-learning \\
		2014 & Buzdalova et al. \cite{buzdalova2014selecting} & Hierarchical-if-and-only-if function, TSP & EA & Q-learning \\
		2018 & Choong et al. \cite{choong2018automatic} & Combinatorial optimization problem used in the CHeSC competition & Hyper-heuristic & Q-learning \\
		2020 & Liu and Wu \cite{liu2020driver} & Driver behavior modeling & PSO & inverse RL \\
		2020 & Kaur and Kumar \cite{kaur2020reinforcement} & Spectrum allocation in cognitive radio networks & NSGA-II & Q-learning \\
		2021 & Hu et al. \cite{hu2021reinforcement} & Parameters extraction of photovoltaic models & DE & Q-learning \\
		2021 & Huynh et al. \cite{huynh2021q} & Truss structural optimization problem & Brief DE & Q-learning \\
		2021 & Radaideh and Shirvan \cite{radaideh2021rule} & Nuclear fuel assembly combinatorial optimization problem & GA, SA, PSO, DE, natural evolution strategies & PPO \\
		2022 & Liu et al. \cite{liu2022hybridization} & Multi-objective knapsack problem, TSP & MOEA & Dynamic Pointer Network \\
		2022 & Song et al. \cite{song2022reinforcement} & Large-scale earth observation satellite scheduling problem & Bees Algorithm & Q-learning \\
		2023 & Gao et al. \cite{gao2023ensemble} & Unmanned Surface Vessels Scheduling Problems & GA, ABC, JAYA, PSO, harmony search & Q-learning \\
		2023 & Li et al. \cite{li2023scheduling} & Multi-objective scheduling problem in a continuous annealing operation of the steel industry & Multi-objective DE & DDQN DDPG \\
		2023 & Peng et al. \cite{peng2023reinforcement} & Global optimization of interplanetary trajectory design & DE & Q-learning \\
		2023 & Zheng et al. \cite{zheng2023reinforced} & TSP & GA & Q-learning \\
		2023 & Zhou et al. \cite{zhou2023improved} & MOP-cloud storage optimization of blockchain & NSGA-III & DQN \\
		2023 & Panzer et al. \cite{panzer2023deep} & Modular production control & Hyper-heuristic & DQN \\
		2023 & Tu et al. \cite{tu2023deep} & Online bin packing problem & Hyper-heuristic & Dueling double deep Q network \\
		2023 & Song et al. \cite{song2023rl} & Electromagnetic detection satellite scheduling problem & GA & Q-learning \\
		2023 & Song et al. \cite{song2023laga} & Earth electromagnetic satellite scheduling problem & GA & Dueling deep Q network \\
		2023 & Guo et al. \cite{guo2023reinforcement} & Team formation problem considering person-job matching & GP & Q-learning \\
		2023 & Liu et al. \cite{liu2023NeuroCrossover} & Bin packing problem, TSP, capacitated VRP & GA & PPO\\
		2023 & Li et al.  \cite{li2023evolutionary} & Multitask optimization problem           & Multifactorial evolutionary algorithm                    & Q-learning \\
		2023 & Lin et al. \cite{lin2023scheduling} & Eight-phase urban traffic light problem & Jaya algorithm, HS, GA, PSO, water cycle algorithm (WCA) & Q-learning \\
		2023 & Wang et al. \cite{wang2023problem}& Urban traffic light scheduling problem  & HS, WCA, Jaya, ABC & Q-learning \\
		\bottomrule[1.5pt]
	\end{tabularx}
\end{table*}

\begin{figure*}[htp]
\centering
\includegraphics[width=0.8\textwidth]{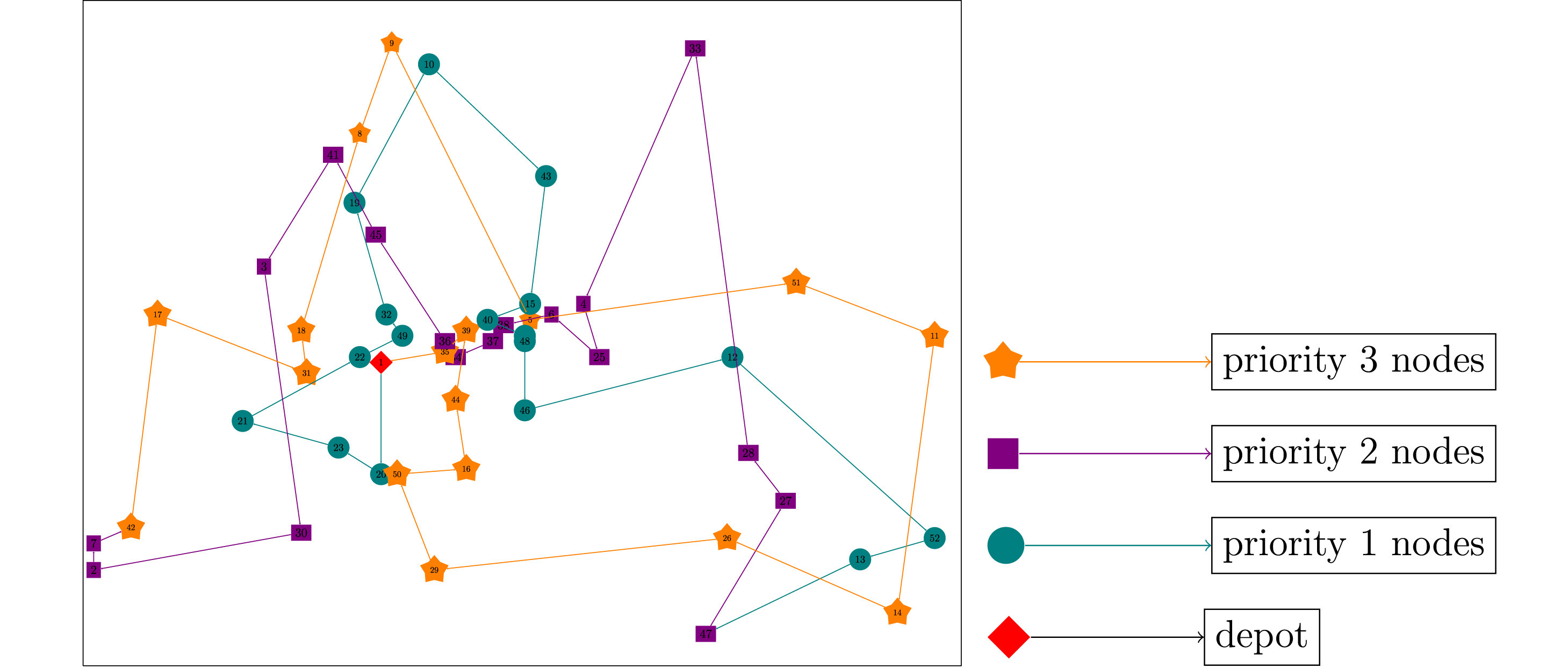}
\caption{An example of clustered traveling salesman problem with d-relaxed priority rule (CTSP-d) \cite{dasari2023two}}
\label{An example of clustered traveling salesman problem with d-relaxed priority rule (CTSP-d)}
\end{figure*}

In addition to the aforementioned two types of combinatorial optimization problems, RL-EA has been applied to solve various other combinatorial optimization problems, which is shown in Table \ref{other combinatorial optimization problem}. It is evident that RL-EA has demonstrated successful outcomes in solving multiple types of combinatorial optimization problems. For example, literature \cite{buzdalova2014selecting,zheng2023reinforced} proposed some RL-EA approaches to addressing different variations of the traveling salesman problem (TSP). TSP belongs to a class of classical optimization problems where a salesman needs to visit multiple cities and return to the starting city. Specifically, the optimization objective of TSP is to find the solution with the shortest distance for the salesman to visit all cities. Fig.\ref{An example of clustered traveling salesman problem with d-relaxed priority rule (CTSP-d)} illustrates an example \cite{dasari2023two} involving clustered TSP with d-relaxed priority rule, which introduces varying priorities among cities and adds complexity to TSPs. Additionally, there exist other TSP variants that consider specific scenarios. Among various solution algorithms, the successful application of RL-EA offers novel insights into tackling such classical combinatorial optimization problems. Table \ref{Comparison of RHGA and other algorithms on 16 hard instances.} shows the path optimization performance of a reinforced hybrid genetic algorithm (RHGA) \cite{zheng2023reinforced} on the TSP benchmark. The RHGA not only significantly outperforms the comparison algorithms but also approaches the known optimal solution.

\begin{sidewaystable}[htbp]
	\scriptsize
	\caption{Comparison of RHGA and other algorithms on 16 hard instances. \cite{zheng2023reinforced}}
	\centering
	\label{Comparison of RHGA and other algorithms on 16 hard instances.}
	\begin{threeparttable}
		\begin{tabular}{llllllllll}
			\toprule[1.5pt]
			\multirow{2}{*}{Instance} & \multirow{2}{*}{Best-known Solution} & \multicolumn{2}{l}{RHGA} & EAX-GA-300 \tnote{*} &  & EAX-GA-400 &  & \multicolumn{2}{l}{Lin–Kernighan–Helsgaun   local search} \\
			\cmidrule(l){3-4} \cmidrule(l){5-6} \cmidrule(l){7-8} \cmidrule(l){9-10}
			&  & Best(gap\%) & Average(gap\%) & Best(gap\%) & Average(gap\%) & Best(gap\%) & Average(gap\%) & Best(gap\%) & Average(gap\%) \\
			\midrule[0.75pt]
			u2319 & 234256 & \textbf{234256.0(0.0000)} & 234256.0(0.0000) & 234273(0.0073) & 234330.4(0.0318) & \textbf{234256(0.0000)} & 234318.8(0.0268) & \textbf{234256(0.0000)} & 234256.0(0.0000) \\
			fl3795 & 28772 & \textbf{28772(0.0000)} & 28777.6(0.0195) & 28815(0.1495) & 28824.0(0.1807) & 28779(0.0243) & 28821.5(0.1720) & 28819(0.1634) & 28906.5(0.4675) \\
			pla7397 & 23260728 & \textbf{23260728(0.0000)} & 23260805.4(0.0003) & 23260814(0.0004) & 23261302.6(0.0025) & 23260814(0.0004) & 23261052.6(0.0014) & \textbf{23260728(0.0000)} & 23260728.0(0.0000) \\
			vm22775 & 569288 & \textbf{569288(0.0000)} & 569291.6(0.0006) & 569289(0.0002) & 569293.6(0.0010) & \textbf{569288(0.0000)} & 569291.3(0.0006) & 569298(0.0018) & 569317.7(0.0052) \\
			icx28698 & 78087 & \textbf{78088(0.0013)} & 78089.1(0.0027) & 78089(0.0026) & 78090.1(0.0040) & 78089(0.0026) & 78089.1(0.0027) & 78098(0.0141) & 78106.1(0.0245) \\
			xib32892 & 96757 & \textbf{96757(0.0000)} & 96758.2(0.0012) & 96758(0.0010) & 96758.7(0.0018) & 96757(0.0000) & 96758.5(0.0016) & 96780(0.0238) & 96789.5(0.0336) \\
			bm33708 & 959289 & \textbf{959289(0.0000)} & 959291.4(0.0003) & 959297(0.0008) & 959301.1(0.0013) & 959291(0.0002) & 959297.2(0.0009) & 959300(0.0011) & 959328.9(0.0042)    16854.9 \\
			pla33810 & 66048945 & \textbf{66050069(0.0017)} & 66050888.1(0.0029) & 66051574(0.0040) & 66054829.0(0.0089) & 66053453(0.0068) & 66055157.3(0.0094) & 66061997(0.0198) & 66.071480.3(0.0341)  16938.4 \\
			pba38478 & 108318 & \textbf{108318(0.0000)} & 108319.8(0.0017) & 108319(0.0009) & 108321.1(0.0029) & \textbf{108318(0.0000)} & 108319.6(0.0015) & 108319(0.0009) & 108337.0(0.0175) \\
			ics39603 & 106819 & 106820(0.0009) & 106822.2(0.0030) & 106820(0.0009) & 106822.2(0.0030) & \textbf{106819(0.0000)} & 106820.3(0.0012) & 106823(0.0037) & 106834.2(0.0142) \\
			ht47608 & 125104 & \textbf{125105(0.0008)} & 125107.5(0.0028) & \textbf{125105(0.0008)} & 125108.5(0.0036) & \textbf{125105(0.0008)} & 125108.5(0.0036) & 125128(0.0192) & 125143.7(0.0317) \\
			fna52057 & 147789 & \textbf{147789(0.0000)} & 147790.7(0.0012) & 147790(0.0007) & 147792.9(0.0026) & 147790(0.0007) & 147792.9(0.0026) & 147800(0.0074) & 147816.8(0.0188) \\
			bna56769 & 158078 & \textbf{158078(0.0000)} & 158080.4(0.0015) & 158079(0.0006) & 158080.6(0.0016) & 158079(0.0006) & 158080.6(0.0016) & 158105(0.0171) & 158117.7(0.0251) \\
			dan59296 & 165371 & \textbf{165372(0.0006)} & 165373.3(0.0014) & 165373(0.0012) & 165375.0(0.0024) & 165373(0.0012) & 165375.0(0.0024) & 165397(0.0157) & 165419.4(0.0293) \\
			ch71009 & 4566506 & 4566508(0.0000) & 4566513.9(0.0002) & 4566508.0(0.0000) & 4566518.0(0.0003) & \textbf{4566507(0.0000)} & 4566508.7(0.0001) & 4566624(0.0026) & 4566887.5(0.0084)   71013.5 \\
			pla85900 & 142382641 & \textbf{142384855(0.0016)} & 142386.004.9(0.0024) & 142390195(0.0053) & 142398897.0(0.0114) & 142388810(0.0043) & 142393735.8(0.0078) & 142409640(0.0190) & 142415681.4(0.0232)\\
			\bottomrule[1.5pt]
		\end{tabular}
		\begin{tablenotes}
			\item[*] EAX-GA: Edge Assembly Crossover Genetic Algorithm
		\end{tablenotes}
	\end{threeparttable}
	
\end{sidewaystable}

\begin{figure}[htp]
\centering
\includegraphics[width=0.45\textwidth]{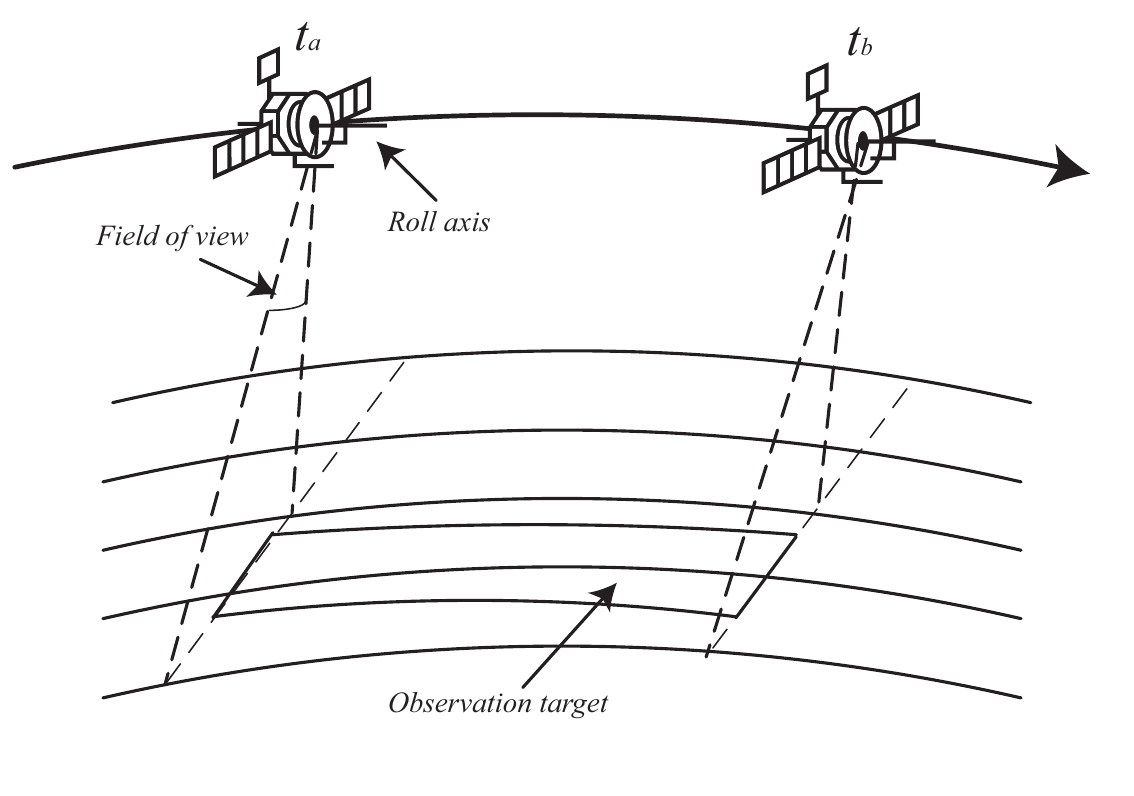}
\caption{A schematic of a region observed by conventional EOS (CEOS) \cite{wang2019robust}}
\label{A schematic of a region observed by conventional EOS (CEOS)}
\end{figure}

\begin{figure}[htp]
\centering
\includegraphics[width=0.45\textwidth]{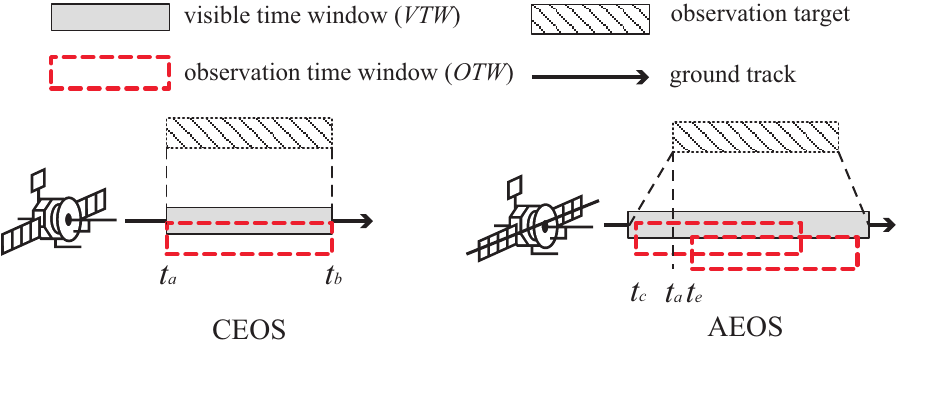}
\caption{Plot of VTW ranges for AEOS and CEOS \cite{wang2020agile}}
\label{Plot of VTW ranges for AEOS and CEOS}
\end{figure}

The RL-EA algorithm has also been used successfully on several occasions to solve the earth observation satellite scheduling problem (EOSSP) \cite{song2023rl,song2023laga}. In this context, we provide a concise introduction to EOSSP. Typically, the objective of an EOSSP is to allocate missions and their specific execution times to a series of satellites. Figure \ref{A schematic of a region observed by conventional EOS (CEOS)} illustrates the observation process of a conventional EOS (CEOS), which can only detect areas directly below and near its flight path. The period during which the target area can be observed is called the visible time window (VTW), and there are strict restrictions on the start and end time of the CEOS observation missions. Additionally, there exists another category of highly maneuverable EOS known as agile EOS (AEOS). As depicted in Figure \ref{Plot of VTW ranges for AEOS and CEOS}, AEOS exhibits longer VTW duration compared to CEOS. The fact that many types of EOSSPs have been proven to be NP-hard poses significant challenges for solution algorithms. Despite these difficulties, RL-EA consistently outperforms state-of-the-art algorithms in terms of scheduling performance.
 
The emergence and utilization of numerous RL-EAs to tackle diverse combinatorial optimization problems can be observed from Table \ref{other combinatorial optimization problem}. These highly complex problems are associated with transportation, manufacturing, energy, and other domains. It is evident that the industry has devoted significant attention to incorporating RL into EA frameworks and has made substantial innovative endeavors. The selection of RL and EA methods constituting RL-EA is flexible and can be determined based on experience or experimental comparisons. We can anticipate a wealth of future research outcomes relevant to real-world applications.

\subsection{Open Datasets in Main Application Domains}
The commonly used datasets are presented in Table \ref{dataset used} to facilitate readers in quickly locating the datasets utilized in related studies. As depicted in Table \ref{dataset used}, the IEEE Congress on Evolutionary Computation (CEC) dataset is extensively employed for continuous optimization problems, encompassing SOP, MOP, MMOP, etc. Numerous test functions within the CEC dataset serve as crucial evaluation criteria for assessing algorithmic solution performance. A Matlab platform for evolutionary multi-objective optimization named PlatEMO \footnote{\url{https://github.com/BIMK/PlatEMO}} makes it easy to use various benchmarks. Additionally, there are also some open datasets for various combinatorial optimization problems. These open datasets enable easy assessment of a new algorithm's performance concerning optimization objective value, convergence rate, solution speed, and more.

\begin{table*}[htbp]
        \footnotesize
	\centering
	\caption{Public datasets for main application domains}
	\label{dataset used}
	\begin{tabularx}{\textwidth}{|m{2cm}|m{3.5cm}|m{8cm}|m{2cm}|}
	\toprule[1.5pt]
	Domain & Dataset & Link & Reference \\
	\hline
	& ZDT test functions & \url{https://jbuisine.github.io/macop/\_build/html/zdt\_example.html} & \cite{huang2020fitness} \\
	& DTLZ test functions & \url{https://github.com/fillipe-gsm/dtlz-test-functions} & \cite{huang2020fitness} \\
	& SMOP test suite & \url{https://github.com/victorlei/smop-testsuite} & \cite{gao2023efficient} \\
        & SOCO 2011 test suite & \url{https://sci2s.ugr.es/EAMHCO} & \cite{tatsis2020reinforced,tatsis2023reinforcement}\\
        & CEC 2005 test suite & \url{https://github.com/P-N-Suganthan/CEC2005} & \cite{rakshit2013realization} \\
	& CEC 2010 test suite & \url{https://github.com/P-N-Suganthan/CEC-2010-LSO-Large-Scale-Opt-} & \cite{wang2022reinforcement} \\
	& CEC 2011 test suite & \url{https://github.com/P-N-Suganthan/CEC-2011--Real\_World\_Problems} & \cite{karafotias2014generic} \\
	& CEC 2013 test suite & \url{https://github.com/P-N-Suganthan/CEC2013} & \cite{fister2022reinforcement, sun2021learning, shiyuan2022reinforcement, liu2023learning,tatsis2023reinforcement} \\
	& CEC2013 MMOP benchmarks & \url{https://github.com/mikeagn/CEC2013} & \cite{xia2021reinforcement} \\
	& CEC 2014 test suite &  \url{https://github.com/P-N-Suganthan/CEC2014} & \cite{fister2022reinforcement} \\
        & CEC 2015 test suite & \url{https://github.com/P-N-Suganthan/CEC2015} & \cite{sadhu2018synergism} \\
	& CEC 2016 test suite & \url{https://github.com/P-N-Suganthan/CEC2016} & \cite{li2019differential} \\
	& CEC 2017 test suite & \url{https://github.com/P-N-Suganthan/CEC2017-BoundContrained} & \cite{zhao2023inverse,fister2022reinforcement, sun2021learning, li2023reinforcement} \\
	& CEC 2018 test suite & \url{https://github.com/P-N-Suganthan/CEC2018} & \cite{zhang2022variational} \\
	\multirow{-15}{*}{\makecell[l]{Continuous \\optimization\\ problem}} & CEC 2022 test suite &  \url{https://github.com/P-N-Suganthan/2022-SO-BO} & \cite{li2023reinforcement} \\
        \hline
	& Taillard benchmarks &  \url{http://mistic.heig-vd.ch/taillard/problemes.dir/ordonnancement.dir/ordonnancement.html} & \cite{zhao2022hyperheuristic, cheng2022multi, karimi2023learning, zhao2022reinforcementb} \\
	& IEEE WCCI 2020 competition for CEVRP &  \url{https://www.researchgate.net/publication/342068571\_Benchmark\_Set\_for\_the\_IEEE\_WCCI-2020\_Competition\_on\_Evolutionary\_Computation\_for\_the\_Electric\_Vehicle\_Routing\_Problem} & \cite{rodriguez2022new} \\
	& FFJSP instances & \url{https://cuglirui.github.io/downloads.htm} & \cite{li2022learning} \\
	& Pickup and Delivery Problem with Time Windows instances & \url{http://iot.ntnu.no/users/larsmahv/benchmarks/} & \cite{kallestad2023general} \\
	& Solomon instances &  \url{https://www.sintef.no/projectweb/top/vrptw/solomon-benchmark/} & \cite{qi2022qmoea} \\
	& DIMACS benchmarks & \url{https://www.cs.ubc.ca/~hoos/SATLIB/benchm.html} & \cite{sun2022reinforcement} \\
	& FJSP-LS instances & \url{https://github.com/CinjaLiao/InstanceDataOfFJSPLS} & \cite{li2023reinforcementb} \\
	& FJSP instances & \url{https://people.idsia.ch/~monaldo/fjsp.html} & \cite{cheng2022multi,li2022reinforcement} \\
	& ESA-ACT team's global trajectory optimization problem database & \url{http://www.esa.int/gsp/ACT/projects/gtop/} & \cite{peng2023reinforcement} \\
	& CHeSC competition instances & \url{http://www.asap.cs.nott.ac.uk/chesc2011/} & \cite{choong2018automatic} \\
	& TSPLIB & \url{http://comopt.ifi.uni-heidelberg.de/software/TSPLIB95/} & \cite{zheng2023reinforced,liu2023NeuroCrossover} \\
	& CVRPLIB & \url{http://vrp.atd-lab.inf.puc-rio.br/index.php/en/} & \cite{liu2023NeuroCrossover} \\
	& FSP instances &  \url{http://people.brunel.ac.uk/~mastjjb/jeb/orlib/files/flowshop1.txt} & \cite{li2022improved} \\
	& OR Library & \url{http://people.brunel.ac.uk/~mastjjb/jeb/info.html} & \cite{li2022improved} \\
	& National TSP benchmarks & \url{https://www.math.uwaterloo.ca/tsp/world/countries.html} & \cite{zheng2023reinforced} \\
	& VLSI TSP benchmarks & \url{https://www.math.uwaterloo.ca/tsp/vlsi/index.html} & \cite{zheng2023reinforced} \\
	\multirow{-17}{*}{\makecell[l]{Combinatorial\\ optimization\\ problem}} & PFSP instances &  \url{http://soa.iti.es} & \cite{zhao2023knowledge}\\
		\bottomrule[1.5pt]                            
	\end{tabularx}
\end{table*}

\section{Future Research Directions}
\label{Future Research Directions}
In recent years, RL-EA has been developed and successfully applied in various fields, providing valuable experiences that can guide related algorithmic research. However, due to the limited number of in-depth algorithmic studies conducted within a short time frame, there are still numerous open problems that require resolution. Here, we unfold some promising research directions in the area of RL-EA for optimization problems.

\textbf{1. Integration of advanced RL models}: The effectiveness of improved RL-assisted strategies in RL-EA heavily relies on the integration of state-of-the-art RL models. Previous studies have predominantly utilized classical RL models, which are not at the forefront of research. Incorporating state-of-the-art RL models into the algorithmic framework is expected to enhance exploration and exploitation capabilities significantly. Moreover, advanced RL models typically exhibit strong generalization abilities, thereby contributing to enhancing performance across diverse problem scenarios. It should be noted that advanced RL models often entail complex network structures and a large number of model parameters. Considering the functions fulfilled by RL within the EA framework, a certain degree of simplification should be carried out in practice.

\textbf{2. Assistance strategy based on multi-agent RL}: RL methods can be categorized into single-agent RL and multi-agent RL based on the number of agents involved. Most of the RL-EAs in the previous work use single-agent RL methods. If these optimization tasks are accomplished by multi-agent RL, it facilitates cooperation among agents, enabling EA to leverage diverse strategies via a unique information synchronization mechanism \cite{bucsoniu2010multi}. Additionally, RL-EA can also employ multiple agents to assist sub-populations or individuals in performing evolutionary operations. This approach allows for a more comprehensive population search that encompasses a wider range of search strategies, facilitating effective ensemble strategies.

\textbf{3. More complex algorithmic mechanism design}: The majority of employed mechanisms in existing RL-EA studies bear striking resemblance to ensemble strategies and traditional adaptive mechanisms. In other words, many newly proposed algorithms simply replace the traditional adaptive mechanisms with RL techniques. The algorithmic mechanism design can be improved from two aspects. Firstly, powerful RL can accomplish more complex tasks compared to the existing assistance strategies, which certainly requires more complex mechanisms to be designed in the algorithms, e.g., selecting the problem subspace and determining the optimization direction of MOP. Moreover, RL can also be utilized to determine whether or not offspring from evolved populations should be retained. Secondly, more advanced ensemble strategies also deserve to be thoroughly investigated. It is important for algorithms to maintain optimal stability during optimization while preserving the diversity of search strategies. Therefore, effective interaction mechanisms need to be designed to facilitate collaboration among different components.

\textbf{4. Digging deeper into the problem characteristics}: The solution space significantly impacts search efficiency and algorithm performance. Even with a stochastic search method, EA can enhance convergence through appropriate guidance. This requires that the problem features can be fully exploited to make the algorithm smarter. Prior research suggests that inverse RL is well-suited for optimizing problems \cite{lindner2022active}. Based on this, researchers can further design the EA framework accordingly. It is necessary to judge whether the information explored by the algorithm itself is valuable for problem evaluation. Only when the information is valuable can it be used as knowledge to guide the algorithm for further searches.

\textbf{5. RL-assisted local search and population search information interaction}: Currently, numerous studies have employed RL-assisted LS, yielding commendable optimization performance. However, the strategies employed in these studies solely concentrate on LS without considering the impact of invalid or erroneous searches on population searches. Particularly when LS is executed in the wrong direction, it not only fails to obtain improved outputs but may also lead to deteriorating search performance. Hence, it is imperative to design an information interaction mechanism between LS and population search to avoid such situations. This mechanism should ensure that the algorithm can proceed in a more optimized direction.

\textbf{6. Combining multiple RL methods}: Most existing RL-EAs use only one RL method, which can only improve the algorithmic performance from a single aspect. Such improvement is often insufficient for solving a complex optimization problem, and therefore many improved versions of EA are designed in detail for multiple aspects of the algorithm. These algorithmic improvement ideas can be practically all realized by RL methods. Since RL is categorized into value-based and strategy-based, two types of decisions can be made for both discrete and continuous classes. By utilizing multiple RL methods simultaneously, the algorithm can be improved in several ways to obtain better performance.

\textcolor[rgb]{0,0,0}{\textbf{7. Low-overhead RL training methods}: The model training method employed in the existing RL-EA is identical to that used for the standalone RL model. While this classical training approach incurs high costs when applied to RL-assisted EA, a low-overhead alternative would significantly enhance the practical value of the algorithm. For example, the evolutionary strategy method proposed by OpenAI can train RL at a much lower cost than the traditional backpropagation training method \cite{salimans2017evolution}. Besides, some pre-training methods and transfer learning methods can also be used to reduce the training workload of RL models.}

\textbf{8. Theoretical analysis of RL-EA}: Despite the existing literature on RL-EA, researchers have predominantly focused on devising novel algorithmic mechanisms to enhance optimization performance and validating them through extensive simulation experiments. Theoretical analysis of each component and the entire RL-EA is imperative, as it can provide valuable guidance for subsequent researchers in designing their algorithms to address optimization problems and enhance algorithmic design objectives. These studies do not fundamentally advance the widespread use of RL-EA because they only show that RL-EA is effective for specific problems. It is necessary to analyze each component and the whole RL-EA theoretically. This work can guide subsequent researchers to make effective improvements based on the optimization problem to be solved and the design goals of the algorithm.

\textbf{9. Efficient RL hyper-parameter setting methods}: The combination of RL and EA makes the new algorithm smart and efficient. However, the integration of these new methods leads to an increasing number of hyper-parameters within the algorithm. Consequently, setting hyper-parameters in RL-EA becomes a time-consuming task. Specifically, determining parameters such as learning rate and discount factor, which significantly impact the algorithm's performance, poses a previously unmet challenge within this integrated framework. In the face of this new challenge, it is imperative to propose a novel approach to setting hyper-parameters in RL that offers flexibility to adapt to optimization objectives or actual usage requirements, thereby enabling effective responses to diverse optimization scenarios. In addition, tailored tuning methods for RL-EA can be designed to further improve the performance of the algorithm.

\textbf{10. Benchmark for RL-EA}: In the literature on optimization problems solved by RL-EA, the benchmarks utilized are either identical or randomly generated as those employed in traditional EA. Existing EA benchmarks are not sufficient to test the newly proposed RL-EA. Therefore, it is imperative to develop tailored RL-EA benchmarks. Although designing RL-EA benchmarks poses challenges, one expedient approach is to derive them from existing EA benchmarks with appropriate modifications. 

\section{Conclusion}
\label{Conclusion}

This paper has presented a comprehensive survey of RL-EA for optimization, mainly focusing on its strategies and applications. Firstly, we analyze the ways of combining RL and EA, including direct and indirect integration. By determining the appropriate integration approach, researchers can quickly establish algorithm structures and concentrate on strategy design. Next, we present the RL-assisted strategies and other attribute settings in RL employed in previous studies. With the assistance of RL, evolutionary operations within EA, such as solution generation, fitness evaluation, operator selection, and parameter control, can be effectively executed. This assisted approach in the algorithm can enhance the RL-EA's ability to conduct efficient search operations based on the acquired environmental information. We then statistically analyze the application of RL-EA in various domains and find that while some research results have been achieved, there is still significant potential for further development whether in general or in specific domains. The use of RL-EA to solve optimization problems across diverse fields can enhance the research of related algorithmic and the emergence of more valuable algorithms. Finally, we summarize existing challenges and suggest some potential future directions for integrating RL into the EA framework.

\section*{Acknowledgements}

This work is supported by the Science and Technology Innovation Team of Shaanxi Province (2023-CX-TD-07), the Special Project in Major Fields of Guangdong Universities (2021ZDZX1019).

We thank all the authors for their efforts and the anonymous reviewers for their valuable comments. In particular, we would like to thank Prof. Ye Tian for his guidance and support.

\section*{Declaration of Competing Interest}
The authors declare that they have no known competing financial interests or personal relationships that could have appeared to influence the work reported in this paper.


\bibliographystyle{unsrt}

\bibliography{mybib}

\begin{thebibliography}{100}

\bibitem{singh2012overview}
Ajay Singh.
\newblock An overview of the optimization modelling applications.
\newblock {\em Journal of Hydrology}, 466:167--182, 2012.

\bibitem{monaci2013exact}
Michele Monaci, Ulrich Pferschy, and Paolo Serafini.
\newblock Exact solution of the robust knapsack problem.
\newblock {\em Computers \& operations research}, 40(11):2625--2631, 2013.

\bibitem{babaei2013general}
M~Babaei.
\newblock A general approach to approximate solutions of nonlinear differential
  equations using particle swarm optimization.
\newblock {\em Applied Soft Computing}, 13(7):3354--3365, 2013.

\bibitem{de2017evolutionary}
Kenneth De~Jong.
\newblock Evolutionary computation: a unified approach.
\newblock In {\em Proceedings of the Genetic and Evolutionary Computation
  Conference Companion}, pages 373--388, 2017.

\bibitem{mahesh2020machine}
Batta Mahesh.
\newblock Machine learning algorithms-a review.
\newblock {\em International Journal of Science and Research
  (IJSR).[Internet]}, 9(1):381--386, 2020.

\bibitem{talbi2021machine}
El-Ghazali Talbi.
\newblock Machine learning into metaheuristics: A survey and taxonomy.
\newblock {\em ACM Computing Surveys (CSUR)}, 54(6):1--32, 2021.

\bibitem{jordan2015machine}
Michael~I Jordan and Tom~M Mitchell.
\newblock Machine learning: Trends, perspectives, and prospects.
\newblock {\em Science}, 349(6245):255--260, 2015.

\bibitem{mnih2013playing}
Volodymyr Mnih, Koray Kavukcuoglu, David Silver, Alex Graves, Ioannis
  Antonoglou, Daan Wierstra, and Martin Riedmiller.
\newblock Playing atari with deep reinforcement learning.
\newblock {\em arXiv preprint arXiv:1312.5602}, 2013.

\bibitem{franccois2018introduction}
Vincent Fran{\c{c}}ois-Lavet, Peter Henderson, Riashat Islam, Marc~G Bellemare,
  Joelle Pineau, et~al.
\newblock An introduction to deep reinforcement learning.
\newblock {\em Foundations and Trends{\textregistered} in Machine Learning},
  11(3-4):219--354, 2018.

\bibitem{drugan2019reinforcement}
Madalina~M Drugan.
\newblock Reinforcement learning versus evolutionary computation: A survey on
  hybrid algorithms.
\newblock {\em Swarm and evolutionary computation}, 44:228--246, 2019.

\bibitem{bai2023evolutionary}
Hui Bai, Ran Cheng, and Yaochu Jin.
\newblock Evolutionary reinforcement learning: A survey.
\newblock {\em arXiv preprint arXiv:2303.04150}, 2023.

\bibitem{mazyavkina2021reinforcement}
Nina Mazyavkina, Sergey Sviridov, Sergei Ivanov, and Evgeny Burnaev.
\newblock Reinforcement learning for combinatorial optimization: A survey.
\newblock {\em Computers \& Operations Research}, 134:105400, 2021.

\bibitem{yang2020survey}
Yunhao Yang and Andrew Whinston.
\newblock A survey on reinforcement learning for combinatorial optimization.
\newblock {\em arXiv preprint arXiv:2008.12248}, 2020.

\bibitem{song2023ensemble}
Yanjie Song, PN~Suganthan, Witold Pedrycz, Junwei Ou, Yongming He, and Yingwu
  Chen.
\newblock Ensemble reinforcement learning: A survey.
\newblock {\em arXiv preprint arXiv:2303.02618}, 2023.

\bibitem{mirjalili2019genetic}
Seyedali Mirjalili and Seyedali Mirjalili.
\newblock Genetic algorithm.
\newblock {\em Evolutionary Algorithms and Neural Networks: Theory and
  Applications}, pages 43--55, 2019.

\bibitem{price2013differential}
Kenneth~V Price.
\newblock Differential evolution.
\newblock In {\em Handbook of optimization: From classical to modern approach},
  pages 187--214. Springer, 2013.

\bibitem{kennedy1995particle}
James Kennedy and Russell Eberhart.
\newblock Particle swarm optimization.
\newblock In {\em Proceedings of ICNN'95-international conference on neural
  networks}, volume~4, pages 1942--1948. IEEE, 1995.

\bibitem{wolpert1997no}
David~H Wolpert and William~G Macready.
\newblock No free lunch theorems for optimization.
\newblock {\em IEEE transactions on evolutionary computation}, 1(1):67--82,
  1997.

\bibitem{neri2012memetic}
Ferrante Neri and Carlos Cotta.
\newblock Memetic algorithms and memetic computing optimization: A literature
  review.
\newblock {\em Swarm and Evolutionary Computation}, 2:1--14, 2012.

\bibitem{deb2002fast}
Kalyanmoy Deb, Amrit Pratap, Sameer Agarwal, and TAMT Meyarivan.
\newblock A fast and elitist multiobjective genetic algorithm: Nsga-ii.
\newblock {\em IEEE transactions on evolutionary computation}, 6(2):182--197,
  2002.

\bibitem{deb2013evolutionary}
Kalyanmoy Deb and Himanshu Jain.
\newblock An evolutionary many-objective optimization algorithm using
  reference-point-based nondominated sorting approach, part i: solving problems
  with box constraints.
\newblock {\em IEEE transactions on evolutionary computation}, 18(4):577--601,
  2013.

\bibitem{powell2007approximate}
Warren~B Powell.
\newblock {\em Approximate Dynamic Programming: Solving the curses of
  dimensionality}, volume 703.
\newblock John Wiley \& Sons, 2007.

\bibitem{sutton2018reinforcement}
Richard~S Sutton and Andrew~G Barto.
\newblock {\em Reinforcement learning: An introduction}.
\newblock MIT press, 2018.

\bibitem{sutton1999policy}
Richard~S Sutton, David McAllester, Satinder Singh, and Yishay Mansour.
\newblock Policy gradient methods for reinforcement learning with function
  approximation.
\newblock {\em Advances in neural information processing systems}, 12, 1999.

\bibitem{mnih2016asynchronous}
Volodymyr Mnih, Adria~Puigdomenech Badia, Mehdi Mirza, Alex Graves, Timothy
  Lillicrap, Tim Harley, David Silver, and Koray Kavukcuoglu.
\newblock Asynchronous methods for deep reinforcement learning.
\newblock In {\em International conference on machine learning}, pages
  1928--1937. PMLR, 2016.

\bibitem{schulman2017proximal}
John Schulman, Filip Wolski, Prafulla Dhariwal, Alec Radford, and Oleg Klimov.
\newblock Proximal policy optimization algorithms.
\newblock {\em arXiv preprint arXiv:1707.06347}, 2017.

\bibitem{lillicrap2015continuous}
Timothy~P Lillicrap, Jonathan~J Hunt, Alexander Pritzel, Nicolas Heess, Tom
  Erez, Yuval Tassa, David Silver, and Daan Wierstra.
\newblock Continuous control with deep reinforcement learning.
\newblock {\em arXiv preprint arXiv:1509.02971}, 2015.

\bibitem{watkins1992q}
Christopher~JCH Watkins and Peter Dayan.
\newblock Q-learning.
\newblock {\em Machine learning}, 8:279--292, 1992.

\bibitem{mnih2015human}
Volodymyr Mnih, Koray Kavukcuoglu, David Silver, Andrei~A Rusu, Joel Veness,
  Marc~G Bellemare, Alex Graves, Martin Riedmiller, Andreas~K Fidjeland, Georg
  Ostrovski, et~al.
\newblock Human-level control through deep reinforcement learning.
\newblock {\em nature}, 518(7540):529--533, 2015.

\bibitem{schaul2015prioritized}
Tom Schaul, John Quan, Ioannis Antonoglou, and David Silver.
\newblock Prioritized experience replay.
\newblock {\em arXiv preprint arXiv:1511.05952}, 2015.

\bibitem{seymour2004temporal}
Ben Seymour, John~P O'Doherty, Peter Dayan, Martin Koltzenburg, Anthony~K
  Jones, Raymond~J Dolan, Karl~J Friston, and Richard~S Frackowiak.
\newblock Temporal difference models describe higher-order learning in humans.
\newblock {\em Nature}, 429(6992):664--667, 2004.

\bibitem{wang2016dueling}
Ziyu Wang, Tom Schaul, Matteo Hessel, Hado Hasselt, Marc Lanctot, and Nando
  Freitas.
\newblock Dueling network architectures for deep reinforcement learning.
\newblock In {\em International conference on machine learning}, pages
  1995--2003. PMLR, 2016.

\bibitem{van2016deep}
Hado Van~Hasselt, Arthur Guez, and David Silver.
\newblock Deep reinforcement learning with double q-learning.
\newblock In {\em Proceedings of the AAAI conference on artificial
  intelligence}, volume~30, 2016.

\bibitem{li2023reinforcementb}
Yibing Li, Cheng Liao, Lei Wang, Yu~Xiao, Yan Cao, and Shunsheng Guo.
\newblock A reinforcement learning-artificial bee colony algorithm for flexible
  job-shop scheduling problem with lot streaming.
\newblock {\em Applied Soft Computing}, page 110658, 2023.

\bibitem{hu2021reinforcement}
Zhenzhen Hu, Wenyin Gong, and Shuijia Li.
\newblock Reinforcement learning-based differential evolution for parameters
  extraction of photovoltaic models.
\newblock {\em Energy Reports}, 7:916--928, 2021.

\bibitem{wu2019ensemble}
Guohua Wu, Rammohan Mallipeddi, and Ponnuthurai~Nagaratnam Suganthan.
\newblock Ensemble strategies for population-based optimization algorithms--a
  survey.
\newblock {\em Swarm and evolutionary computation}, 44:695--711, 2019.

\bibitem{zhang2022multi}
Xiangyin Zhang, Shuang Xia, Xiuzhi Li, and Tian Zhang.
\newblock Multi-objective particle swarm optimization with multi-mode
  collaboration based on reinforcement learning for path planning of unmanned
  air vehicles.
\newblock {\em Knowledge-Based Systems}, 250:109075, 2022.

\bibitem{sun2022reinforcement}
Zhe Sun, Una Benlic, Mingjie Li, and Qinghua Wu.
\newblock Reinforcement learning based tabu search for the minimum load
  coloring problem.
\newblock {\em Computers \& Operations Research}, 143:105745, 2022.

\bibitem{liu2022hybridization}
Wei Liu, Rui Wang, Tao Zhang, Kaiwen Li, Wenhua Li, and Hisao Ishibuchi.
\newblock Hybridization of evolutionary algorithm and deep reinforcement
  learning for multi-objective orienteering optimization.
\newblock {\em IEEE Transactions on Evolutionary Computation}, 2022.

\bibitem{lin2022pareto}
Xi~Lin, Zhiyuan Yang, Xiaoyuan Zhang, and Qingfu Zhang.
\newblock Pareto set learning for expensive multi-objective optimization.
\newblock {\em Advances in Neural Information Processing Systems},
  35:19231--19247, 2022.

\bibitem{khadilkar2018scalable}
Harshad Khadilkar.
\newblock A scalable reinforcement learning algorithm for scheduling railway
  lines.
\newblock {\em IEEE Transactions on Intelligent Transportation Systems},
  20(2):727--736, 2018.

\bibitem{mao2016resource}
Hongzi Mao, Mohammad Alizadeh, Ishai Menache, and Srikanth Kandula.
\newblock Resource management with deep reinforcement learning.
\newblock In {\em Proceedings of the 15th ACM workshop on hot topics in
  networks}, pages 50--56, 2016.

\bibitem{waschneck2018optimization}
Bernd Waschneck, Andr{\'e} Reichstaller, Lenz Belzner, Thomas Altenm{\"u}ller,
  Thomas Bauernhansl, Alexander Knapp, and Andreas Kyek.
\newblock Optimization of global production scheduling with deep reinforcement
  learning.
\newblock {\em Procedia Cirp}, 72:1264--1269, 2018.

\bibitem{su2023evolution}
Chupeng Su, Cong Zhang, Dan Xia, Baoan Han, Chuang Wang, Gang Chen, and Longhan
  Xie.
\newblock Evolution strategies-based optimized graph reinforcement learning for
  solving dynamic job shop scheduling problem.
\newblock {\em Applied Soft Computing}, page 110596, 2023.

\bibitem{budhraja2017neuroevolution}
Karan~K Budhraja and Tim Oates.
\newblock Neuroevolution-based inverse reinforcement learning.
\newblock In {\em 2017 IEEE Congress on Evolutionary Computation (CEC)}, pages
  67--76. IEEE, 2017.

\bibitem{zhao2023inverse}
Fuqing Zhao, Qiaoyun Wang, and Ling Wang.
\newblock An inverse reinforcement learning framework with the q-learning
  mechanism for the metaheuristic algorithm.
\newblock {\em Knowledge-Based Systems}, 265:110368, 2023.

\bibitem{liu2020driver}
Zeng-Jie Liu and Huai-Ning Wu.
\newblock Driver behavior modeling via inverse reinforcement learning based on
  particle swarm optimization.
\newblock In {\em 2020 Chinese Automation Congress (CAC)}, pages 7232--7237.
  IEEE, 2020.

\bibitem{choong2018automatic}
Shin~Siang Choong, Li-Pei Wong, and Chee~Peng Lim.
\newblock Automatic design of hyper-heuristic based on reinforcement learning.
\newblock {\em Information Sciences}, 436:89--107, 2018.

\bibitem{zhao2022hyperheuristic}
Fuqing Zhao, Shilu Di, and Ling Wang.
\newblock A hyperheuristic with q-learning for the multiobjective
  energy-efficient distributed blocking flow shop scheduling problem.
\newblock {\em IEEE Transactions on Cybernetics}, 2022.

\bibitem{wu2023aq}
Fang-Chun Wu, Bin Qian, Rong Hu, Zi-Qi Zhang, and Bin Wang.
\newblock A q-learning-based hyper-heuristic evolutionary algorithm for the
  distributed flexible job-shop scheduling problem.
\newblock In {\em International Conference on Intelligent Computing}, pages
  251--261. Springer, 2023.

\bibitem{zhang2023q}
Zi-Qi Zhang, Fang-Chun Wu, Bin Qian, Rong Hu, Ling Wang, and Huai-Ping Jin.
\newblock A q-learning-based hyper-heuristic evolutionary algorithm for the
  distributed flexible job-shop scheduling problem with crane transportation.
\newblock {\em Expert Systems with Applications}, page 121050, 2023.

\bibitem{zhu2023hyper}
Jin-Han Zhu, Rong Hu, Zuo-Cheng Li, Bin Qian, and Zi-Qi Zhang.
\newblock Hyper-heuristic q-learning algorithm for flow-shop scheduling problem
  with fuzzy processing times.
\newblock In {\em International Conference on Intelligent Computing}, pages
  194--205. Springer, 2023.

\bibitem{shang2023green}
Chunjian Shang, Liang Ma, and Yong Liu.
\newblock Green location routing problem with flexible multi-compartment for
  source-separated waste: A q-learning and multi-strategy-based hyper-heuristic
  algorithm.
\newblock {\em Engineering Applications of Artificial Intelligence},
  121:105954, 2023.

\bibitem{cheng2022multi}
Lixin Cheng, Qiuhua Tang, Liping Zhang, and Zikai Zhang.
\newblock Multi-objective q-learning-based hyper-heuristic with bi-criteria
  selection for energy-aware mixed shop scheduling.
\newblock {\em Swarm and Evolutionary Computation}, 69:100985, 2022.

\bibitem{qin2021novel}
Wei Qin, Zilong Zhuang, Zizhao Huang, and Haozhe Huang.
\newblock A novel reinforcement learning-based hyper-heuristic for
  heterogeneous vehicle routing problem.
\newblock {\em Computers \& Industrial Engineering}, 156:107252, 2021.

\bibitem{zhang2022deep}
Yuchang Zhang, Ruibin Bai, Rong Qu, Chaofan Tu, and Jiahuan Jin.
\newblock A deep reinforcement learning based hyper-heuristic for combinatorial
  optimisation with uncertainties.
\newblock {\em European Journal of Operational Research}, 300(2):418--427,
  2022.

\bibitem{tu2023deep}
Chaofan Tu, Ruibin Bai, Uwe Aickelin, Yuchang Zhang, and Heshan Du.
\newblock A deep reinforcement learning hyper-heuristic with feature fusion for
  online packing problems.
\newblock {\em Expert Systems with Applications}, page 120568, 2023.

\bibitem{yang2010engineering}
Xin-She Yang and Suash Deb.
\newblock Engineering optimisation by cuckoo search.
\newblock {\em International Journal of Mathematical Modelling and Numerical
  Optimisation}, 1(4):330--343, 2010.

\bibitem{bertsimas1993simulated}
Dimitris Bertsimas and John Tsitsiklis.
\newblock Simulated annealing.
\newblock {\em Statistical science}, 8(1):10--15, 1993.

\bibitem{buzdalova2014selecting}
Arina Buzdalova, Vladislav Kononov, and Maxim Buzdalov.
\newblock Selecting evolutionary operators using reinforcement learning:
  Initial explorations.
\newblock In {\em Proceedings of the companion publication of the 2014 annual
  conference on genetic and evolutionary computation}, pages 1033--1036, 2014.

\bibitem{li2019differential}
Zhihui Li, Li~Shi, Caitong Yue, Zhigang Shang, and Boyang Qu.
\newblock Differential evolution based on reinforcement learning with fitness
  ranking for solving multimodal multiobjective problems.
\newblock {\em Swarm and Evolutionary Computation}, 49:234--244, 2019.

\bibitem{fister2022reinforcement}
Iztok Fister, Du{\v{s}}an Fister, and Iztok Fister~Jr.
\newblock Reinforcement learning-based differential evolution for global
  optimization.
\newblock In {\em Differential Evolution: From Theory to Practice}, pages
  43--75. Springer, 2022.

\bibitem{li2023scheduling}
Tianyang Li, Ying Meng, and Lixin Tang.
\newblock Scheduling of continuous annealing with a multi-objective
  differential evolution algorithm based on deep reinforcement learning.
\newblock {\em IEEE Transactions on Automation Science and Engineering}, 2023.

\bibitem{zhang2023reinforcement}
Zikai Zhang, Qiuhua Tang, Manuel Chica, and Zixiang Li.
\newblock Reinforcement learning-based multiobjective evolutionary algorithm
  for mixed-model multimanned assembly line balancing under uncertain demand.
\newblock {\em IEEE Transactions on Cybernetics}, 2023.

\bibitem{song2023rl}
Yanjie Song, Luona Wei, Qing Yang, Jian Wu, Lining Xing, and Yingwu Chen.
\newblock Rl-ga: A reinforcement learning-based genetic algorithm for
  electromagnetic detection satellite scheduling problem.
\newblock {\em Swarm and Evolutionary Computation}, 77:101236, 2023.

\bibitem{zhang2007moea}
Qingfu Zhang and Hui Li.
\newblock Moea/d: A multiobjective evolutionary algorithm based on
  decomposition.
\newblock {\em IEEE Transactions on evolutionary computation}, 11(6):712--731,
  2007.

\bibitem{tian2022deep}
Ye~Tian, Xiaopeng Li, Haiping Ma, Xingyi Zhang, Kay~Chen Tan, and Yaochu Jin.
\newblock Deep reinforcement learning based adaptive operator selection for
  evolutionary multi-objective optimization.
\newblock {\em IEEE Transactions on Emerging Topics in Computational
  Intelligence}, 2022.

\bibitem{karimi2023learning}
Maryam Karimi-Mamaghan, Mehrdad Mohammadi, Bastien Pasdeloup, and Patrick
  Meyer.
\newblock Learning to select operators in meta-heuristics: An integration of
  q-learning into the iterated greedy algorithm for the permutation flowshop
  scheduling problem.
\newblock {\em European Journal of Operational Research}, 304(3):1296--1330,
  2023.

\bibitem{ren2023novel}
Yaxian Ren, Kaizhou Gao, Yaping Fu, Hongyan Sang, Dachao Li, and Zile Luo.
\newblock A novel q-learning based variable neighborhood iterative search
  algorithm for solving disassembly line scheduling problems.
\newblock {\em Swarm and Evolutionary Computation}, 80:101338, 2023.

\bibitem{wang2022adaptive}
Jing Wang, Deming Lei, and Jingcao Cai.
\newblock An adaptive artificial bee colony with reinforcement learning for
  distributed three-stage assembly scheduling with maintenance.
\newblock {\em Applied Soft Computing}, 117:108371, 2022.

\bibitem{li2022improved}
Hanxiao Li, Kaizhou Gao, Pei-Yong Duan, Jun-Qing Li, and Le~Zhang.
\newblock An improved artificial bee colony algorithm with q-learning for
  solving permutation flow-shop scheduling problems.
\newblock {\em IEEE Transactions on Systems, Man, and Cybernetics: Systems},
  53(5):2684--2693, 2022.

\bibitem{zhao2022reinforcementa}
Fuqing Zhao, Zhenyu Wang, and Ling Wang.
\newblock A reinforcement learning driven artificial bee colony algorithm for
  distributed heterogeneous no-wait flowshop scheduling problem with
  sequence-dependent setup times.
\newblock {\em IEEE Transactions on Automation Science and Engineering}, 2022.

\bibitem{zhou2023adaptive}
Binghai Zhou and Zhe Zhao.
\newblock An adaptive artificial bee colony algorithm enhanced by deep
  q-learning for milk-run vehicle scheduling problem based on supply hub.
\newblock {\em Knowledge-Based Systems}, 264:110367, 2023.

\bibitem{zheng2023reinforced}
Jiongzhi Zheng, Jialun Zhong, Menglei Chen, and Kun He.
\newblock A reinforced hybrid genetic algorithm for the traveling salesman
  problem.
\newblock {\em Computers \& Operations Research}, 157:106249, 2023.

\bibitem{qi2022qmoea}
Rui Qi, Jun-qing Li, Juan Wang, Hui Jin, and Yu-yan Han.
\newblock Qmoea: A q-learning-based multiobjective evolutionary algorithm for
  solving time-dependent green vehicle routing problems with time windows.
\newblock {\em Information Sciences}, 608:178--201, 2022.

\bibitem{li2023muti}
Peize Li, Qiang Xue, Ziteng Zhang, Jian Chen, and Dequn Zhou.
\newblock Muti-objective energy-efficient hybrid flow shop scheduling using
  q-learning and gvns driven nsga-ii.
\newblock {\em Computers \& Operations Research}, page 106360, 2023.

\bibitem{du2022knowledge}
Yu~Du, Jun-qing Li, Xiao-long Chen, Pei-yong Duan, and Quan-ke Pan.
\newblock Knowledge-based reinforcement learning and estimation of distribution
  algorithm for flexible job shop scheduling problem.
\newblock {\em IEEE Transactions on Emerging Topics in Computational
  Intelligence}, 2022.

\bibitem{yan2023novel}
Zheping Yan, Jinyu Yan, Yifan Wu, Sijia Cai, and Hongxing Wang.
\newblock A novel reinforcement learning based tuna swarm optimization
  algorithm for autonomous underwater vehicle path planning.
\newblock {\em Mathematics and Computers in Simulation}, 209:55--86, 2023.

\bibitem{zhao2022reinforcementb}
Fuqing Zhao, Tao Jiang, and Ling Wang.
\newblock A reinforcement learning driven cooperative meta-heuristic algorithm
  for energy-efficient distributed no-wait flow-shop scheduling with
  sequence-dependent setup time.
\newblock {\em IEEE Transactions on Industrial Informatics}, 2022.

\bibitem{gao2023ensemble}
Minglong Gao, Kaizhou Gao, Zhenfang Ma, and Weiyu Tang.
\newblock Ensemble meta-heuristics and q-learning for solving unmanned surface
  vessels scheduling problems.
\newblock {\em Swarm and Evolutionary Computation}, page 101358, 2023.

\bibitem{zhao2023knowledge}
Fuqing Zhao, Gang Zhou, Tianpeng Xu, Ningning Zhu, et~al.
\newblock A knowledge-driven cooperative scatter search algorithm with
  reinforcement learning for the distributed blocking flow shop scheduling
  problem.
\newblock {\em Expert Systems with Applications}, page 120571, 2023.

\bibitem{jia2023q}
Yanhe Jia, Qi~Yan, and Hongfeng Wang.
\newblock Q-learning driven multi-population memetic algorithm for distributed
  three-stage assembly hybrid flow shop scheduling with flexible preventive
  maintenance.
\newblock {\em Expert Systems with Applications}, page 120837, 2023.

\bibitem{guo2023reinforcement}
Yangyang Guo, Hao Wang, Lei He, Witold Pedrycz, PN~Suganthan, and Yanjie Song.
\newblock A reinforcement learning-assisted genetic programming algorithm for
  team formation problem considering person-job matching.
\newblock {\em arXiv preprint arXiv:2304.04022}, 2023.

\bibitem{eiben2012evolutionary}
Agoston~Endre Eiben and Selmar~K Smit.
\newblock Evolutionary algorithm parameters and methods to tune them.
\newblock In {\em Autonomous search}, pages 15--36. Springer, 2012.

\bibitem{shahrabi2017reinforcement}
Jamal Shahrabi, Mohammad~Amin Adibi, and Masoud Mahootchi.
\newblock A reinforcement learning approach to parameter estimation in dynamic
  job shop scheduling.
\newblock {\em Computers \& Industrial Engineering}, 110:75--82, 2017.

\bibitem{rakshit2013realization}
Pratyusha Rakshit, Amit Konar, Pavel Bhowmik, Indrani Goswami, Swagatam Das,
  Lakhmi~C Jain, and Atulya~K Nagar.
\newblock Realization of an adaptive memetic algorithm using differential
  evolution and q-learning: A case study in multirobot path planning.
\newblock {\em IEEE Transactions on Systems, Man, and Cybernetics: Systems},
  43(4):814--831, 2013.

\bibitem{karafotias2014generic}
Giorgos Karafotias, Agoston~Endre Eiben, and Mark Hoogendoorn.
\newblock Generic parameter control with reinforcement learning.
\newblock In {\em Proceedings of the 2014 annual conference on genetic and
  evolutionary computation}, pages 1319--1326, 2014.

\bibitem{sadhu2018synergism}
Arup~Kumar Sadhu, Amit Konar, Tanuka Bhattacharjee, and Swagatam Das.
\newblock Synergism of firefly algorithm and q-learning for robot arm path
  planning.
\newblock {\em Swarm and Evolutionary Computation}, 43:50--68, 2018.

\bibitem{kaur2020reinforcement}
Amandeep Kaur and Krishan Kumar.
\newblock A reinforcement learning based evolutionary multi-objective
  optimization algorithm for spectrum allocation in cognitive radio networks.
\newblock {\em Physical Communication}, 43:101196, 2020.

\bibitem{huynh2021q}
Thanh~N Huynh, Dieu~TT Do, and Jaehong Lee.
\newblock Q-learning-based parameter control in differential evolution for
  structural optimization.
\newblock {\em Applied Soft Computing}, 107:107464, 2021.

\bibitem{sun2021learning}
Jianyong Sun, Xin Liu, Thomas B{\"a}ck, and Zongben Xu.
\newblock Learning adaptive differential evolution algorithm from optimization
  experiences by policy gradient.
\newblock {\em IEEE Transactions on Evolutionary Computation}, 25(4):666--680,
  2021.

\bibitem{tessari2022reinforcement}
Michele Tessari and Giovanni Iacca.
\newblock Reinforcement learning based adaptive metaheuristics.
\newblock In {\em Proceedings of the Genetic and Evolutionary Computation
  Conference Companion}, pages 1854--1861, 2022.

\bibitem{hansen2003reducing}
Nikolaus Hansen, Sibylle~D M{\"u}ller, and Petros Koumoutsakos.
\newblock Reducing the time complexity of the derandomized evolution strategy
  with covariance matrix adaptation (cma-es).
\newblock {\em Evolutionary computation}, 11(1):1--18, 2003.

\bibitem{zhang2022variational}
Haotian Zhang, Jianyong Sun, Yuhao Wang, Jialong Shi, and Zongben Xu.
\newblock Variational reinforcement learning for hyper-parameter tuning of
  adaptive evolutionary algorithm.
\newblock {\em IEEE Transactions on Emerging Topics in Computational
  Intelligence}, 2022.

\bibitem{cheng2022scheduling}
Lixin Cheng, Qiuhua Tang, Liping Zhang, and Chunlong Yu.
\newblock Scheduling flexible manufacturing cell with no-idle flow-lines and
  job-shop via q-learning-based genetic algorithm.
\newblock {\em Computers \& Industrial Engineering}, 169:108293, 2022.

\bibitem{li2022learning}
Rui Li, Wenyin Gong, Chao Lu, and Ling Wang.
\newblock A learning-based memetic algorithm for energy-efficient flexible job
  shop scheduling with type-2 fuzzy processing time.
\newblock {\em IEEE Transactions on Evolutionary Computation}, 2022.

\bibitem{li2022reinforcement}
Rui Li, Wenyin Gong, and Chao Lu.
\newblock A reinforcement learning based rmoea/d for bi-objective fuzzy
  flexible job shop scheduling.
\newblock {\em Expert Systems with Applications}, 203:117380, 2022.

\bibitem{shiyuan2022reinforcement}
Shiyuan Yin.
\newblock Reinforcement learning based parameters adaption method for particle
  swarm optimization.
\newblock {\em arXiv preprint arXiv:2206.00835}, 2022.

\bibitem{peng2023reinforcement}
Lei Peng, Zhuoming Yuan, Guangming Dai, Maocai Wang, and Zhe Tang.
\newblock Reinforcement learning-based hybrid differential evolution for global
  optimization of interplanetary trajectory design.
\newblock {\em Swarm and Evolutionary Computation}, page 101351, 2023.

\bibitem{liu2023learning}
Xin Liu, Jianyong Sun, Qingfu Zhang, Zhenkun Wang, and Zongben Xu.
\newblock Learning to learn evolutionary algorithm: A learnable differential
  evolution.
\newblock {\em IEEE Transactions on Emerging Topics in Computational
  Intelligence}, 2023.

\bibitem{song2023laga}
Yanjie Song, Jie Chun, Qinwen Yang, Junwei Ou, Lining Xing, and Yingwu Chen.
\newblock Laga: A learning adaptive genetic algorithm for earth electromagnetic
  satellite scheduling problem.
\newblock {\em arXiv preprint arXiv:2301.02764}, 2023.

\bibitem{li2023reinforcement}
Wei Li, Peng Liang, Bo~Sun, Yafeng Sun, and Ying Huang.
\newblock Reinforcement learning-based particle swarm optimization with
  neighborhood differential mutation strategy.
\newblock {\em Swarm and Evolutionary Computation}, 78:101274, 2023.

\bibitem{tatsis2020reinforced}
Vasileios~A Tatsis and Konstantinos~E Parsopoulos.
\newblock Reinforced online parameter adaptation method for population-based
  metaheuristics.
\newblock In {\em 2020 IEEE Symposium Series on Computational Intelligence
  (SSCI)}, pages 360--367. IEEE, 2020.

\bibitem{tatsis2023reinforcement}
Vasileios~A Tatsis and Konstantinos~E Parsopoulos.
\newblock Reinforcement learning for enhanced online gradient-based parameter
  adaptation in metaheuristics.
\newblock {\em Swarm and Evolutionary Computation}, page 101371, 2023.

\bibitem{gao2023improved}
Yi-Jie Gao, Qing-Xia Shang, Yuan-Yuan Yang, Rong Hu, and Bin Qian.
\newblock Improved particle swarm optimization algorithm combined with
  reinforcement learning for solving flexible job shop scheduling problem.
\newblock In {\em International Conference on Intelligent Computing}, pages
  288--298. Springer, 2023.

\bibitem{buzdalova2012increasing}
Arina Buzdalova and Maxim Buzdalov.
\newblock Increasing efficiency of evolutionary algorithms by choosing between
  auxiliary fitness functions with reinforcement learning.
\newblock In {\em 2012 11th International Conference on Machine Learning and
  Applications}, volume~1, pages 150--155. IEEE, 2012.

\bibitem{huang2020fitness}
Ying Huang, Wei Li, Furong Tian, and Xiang Meng.
\newblock A fitness landscape ruggedness multiobjective differential evolution
  algorithm with a reinforcement learning strategy.
\newblock {\em Applied Soft Computing}, 96:106693, 2020.

\bibitem{xia2021reinforcement}
Hai Xia, Changhe Li, Sanyou Zeng, Qingshan Tan, Junchen Wang, and Shengxiang
  Yang.
\newblock A reinforcement-learning-based evolutionary algorithm using solution
  space clustering for multimodal optimization problems.
\newblock In {\em 2021 IEEE Congress on Evolutionary Computation (CEC)}, pages
  1938--1945. IEEE, 2021.

\bibitem{radaideh2021rule}
Majdi~I Radaideh and Koroush Shirvan.
\newblock Rule-based reinforcement learning methodology to inform evolutionary
  algorithms for constrained optimization of engineering applications.
\newblock {\em Knowledge-Based Systems}, 217:106836, 2021.

\bibitem{wang2022reinforcement}
Feng Wang, Xujie Wang, and Shilei Sun.
\newblock A reinforcement learning level-based particle swarm optimization
  algorithm for large-scale optimization.
\newblock {\em Information Sciences}, 602:298--312, 2022.

\bibitem{gao2023efficient}
Mengqi Gao, Xiang Feng, Huiqun Yu, and Xiuquan Li.
\newblock An efficient evolutionary algorithm based on deep reinforcement
  learning for large-scale sparse multiobjective optimization.
\newblock {\em Applied Intelligence}, pages 1--24, 2023.

\bibitem{zhou2023improved}
Yu~Zhou, Yanli Ren, Mengtian Xu, and Guorui Feng.
\newblock An improved nsga-iii algorithm based on deep q-networks for cloud
  storage optimization of blockchain.
\newblock {\em IEEE Transactions on Parallel and Distributed Systems},
  34(5):1406--1419, 2023.

\bibitem{liu2023NeuroCrossover}
Haoqiang Liu, Zefang Zong, and Deping Jin.
\newblock Neurocrossover: An intelligent genetic locus selection scheme for
  genetic algorithm using reinforcement learning.
\newblock {\em Applied Soft Computing}, page 110680, 2023.

\bibitem{qiu2023q}
Haiyun Qiu, Bowen Xue, Qinge Xiao, and Ben Niu.
\newblock Q-learning based particle swarm optimization with multi-exemplar and
  elite learning.
\newblock In {\em International Conference on Intelligent Computing}, pages
  310--321. Springer, 2023.

\bibitem{zhao2023multi}
Fuqing Zhao, Zhenyu Wang, Ling Wang, Tianpeng Xu, Ningning Zhu, et~al.
\newblock A multi-agent reinforcement learning driven artificial bee colony
  algorithm with the central controller.
\newblock {\em Expert Systems with Applications}, 219:119672, 2023.

\bibitem{wang2014mommop}
Yong Wang, Han-Xiong Li, Gary~G Yen, and Wu~Song.
\newblock Mommop: Multiobjective optimization for locating multiple optimal
  solutions of multimodal optimization problems.
\newblock {\em IEEE transactions on cybernetics}, 45(4):830--843, 2014.

\bibitem{zhao2023reinforcement}
Fuqing Zhao, Xiaotong Hu, Ling Wang, Tianpeng Xu, Ningning Zhu, and Jonrinaldi.
\newblock A reinforcement learning-driven brain storm optimisation algorithm
  for multi-objective energy-efficient distributed assembly no-wait flow shop
  scheduling problem.
\newblock {\em International Journal of Production Research}, 61(9):2854--2872,
  2023.

\bibitem{zhao2023cooperative}
Fuqing Zhao, Gang Zhou, and Ling Wang.
\newblock A cooperative scatter search with reinforcement learning mechanism
  for the distributed permutation flowshop scheduling problem with
  sequence-dependent setup times.
\newblock {\em IEEE Transactions on Systems, Man, and Cybernetics: Systems},
  2023.

\bibitem{yu2023improved}
Hui Yu, Kai-Zhou Gao, Zhen-Fang Ma, and Yu-Xia Pan.
\newblock Improved meta-heuristics with q-learning for solving distributed
  assembly permutation flowshop scheduling problems.
\newblock {\em Swarm and Evolutionary Computation}, 80:101335, 2023.

\bibitem{rodriguez2022new}
Erick Rodr{\'\i}guez-Esparza, Antonio~D Masegosa, Diego Oliva, and Enrique
  Onieva.
\newblock A new hyper-heuristic based on adaptive simulated annealing and
  reinforcement learning for the capacitated electric vehicle routing problem.
\newblock {\em arXiv preprint arXiv:2206.03185}, 2022.

\bibitem{kucukoglu2021electric}
Ilker Kucukoglu, Reginald Dewil, and Dirk Cattrysse.
\newblock The electric vehicle routing problem and its variations: A literature
  review.
\newblock {\em Computers \& Industrial Engineering}, 161:107650, 2021.

\bibitem{rastani2019effects}
Sina Rastani, Tu{\u{g}}{\c{c}}e Y{\"u}ksel, and B{\"u}lent {\c{C}}atay.
\newblock Effects of ambient temperature on the route planning of electric
  freight vehicles.
\newblock {\em Transportation Research Part D: Transport and Environment},
  74:124--141, 2019.

\bibitem{song2022reinforcement}
Yan-jie Song, Jun-wei Ou, DT~Pham, Ji-ting Li, Jing-bo Huang, and Li-ning Xing.
\newblock A reinforcement-learning-driven bees algorithm for large-scale earth
  observation satellite scheduling.
\newblock In {\em International Conference on Bio-Inspired Computing: Theories
  and Applications}, pages 81--91. Springer, 2022.

\bibitem{panzer2023deep}
Marcel Panzer, Benedict Bender, and Norbert Gronau.
\newblock A deep reinforcement learning based hyper-heuristic for modular
  production control.
\newblock {\em International Journal of Production Research}, pages 1--22,
  2023.

\bibitem{li2023evolutionary}
Shuijia Li, Wenyin Gong, Ling Wang, and Qiong Gu.
\newblock Evolutionary multitasking via reinforcement learning.
\newblock {\em IEEE Transactions on Emerging Topics in Computational
  Intelligence}, 2023.

\bibitem{lin2023scheduling}
Zhongjie Lin, Kaizhou Gao, Naiqi Wu, and Ponnuthurai~Nagaratnam Suganthan.
\newblock Scheduling eight-phase urban traffic light problems via ensemble
  meta-heuristics and q-learning based local search.
\newblock {\em IEEE Transactions on Intelligent Transportation Systems}, 2023.

\bibitem{wang2023problem}
Liang Wang, Kaizhou Gao, Zhongjie Lin, Wuze Huang, and Ponnuthurai~Nagaratnam
  Suganthan.
\newblock Problem feature based meta-heuristics with q-learning for solving
  urban traffic light scheduling problems.
\newblock {\em Applied Soft Computing}, 147:110714, 2023.

\bibitem{dasari2023two}
Kasi~Viswanath Dasari and Alok Singh.
\newblock Two heuristic approaches for clustered traveling salesman problem
  with d-relaxed priority rule.
\newblock {\em Expert Systems with Applications}, 224:120003, 2023.

\bibitem{wang2019robust}
Xinwei Wang, Guopeng Song, Roel Leus, and Chao Han.
\newblock Robust earth observation satellite scheduling with uncertainty of
  cloud coverage.
\newblock {\em IEEE Transactions on Aerospace and Electronic Systems},
  56(3):2450--2461, 2019.

\bibitem{wang2020agile}
Xinwei Wang, Guohua Wu, Lining Xing, and Witold Pedrycz.
\newblock Agile earth observation satellite scheduling over 20 years:
  Formulations, methods, and future directions.
\newblock {\em IEEE Systems Journal}, 15(3):3881--3892, 2020.

\bibitem{kallestad2023general}
Jakob Kallestad, Ramin Hasibi, Ahmad Hemmati, and Kenneth S{\"o}rensen.
\newblock A general deep reinforcement learning hyperheuristic framework for
  solving combinatorial optimization problems.
\newblock {\em European Journal of Operational Research}, 309(1):446--468,
  2023.

\bibitem{bucsoniu2010multi}
Lucian Bu{\c{s}}oniu, Robert Babu{\v{s}}ka, and Bart De~Schutter.
\newblock Multi-agent reinforcement learning: An overview.
\newblock {\em Innovations in multi-agent systems and applications-1}, pages
  183--221, 2010.

\bibitem{lindner2022active}
David Lindner, Andreas Krause, and Giorgia Ramponi.
\newblock Active exploration for inverse reinforcement learning.
\newblock {\em Advances in Neural Information Processing Systems},
  35:5843--5853, 2022.

\bibitem{salimans2017evolution}
Tim Salimans, Jonathan Ho, Xi~Chen, Szymon Sidor, and Ilya Sutskever.
\newblock Evolution strategies as a scalable alternative to reinforcement
  learning.
\newblock {\em arXiv preprint arXiv:1703.03864}, 2017.

\end{thebibliography}



\end{document}